\documentclass[letterpaper, 10 pt, conference]{ieeeconf}  

\IEEEoverridecommandlockouts                              

\makeatletter
\let\NAT@parse\undefined
\makeatother

\usepackage[colorlinks=true]{hyperref}
\usepackage{url}

\usepackage{amsmath}
\usepackage{amssymb}

\usepackage{flushend}

\usepackage[capitalize]{cleveref}

\usepackage[acronym]{glossaries}
\newacronym{llm}{LLM}{Large Language Model}
\newacronym{lora}{LoRA}{Low-Rank Adaptation}
\newacronym{ap}{AP}{Average Precision}
\newacronym{kp-mse}{KP-MSE}{Keypoint Mean Squared Error}
\newacronym{iou}{IoU}{Intersection Over Union}
\newacronym{vr}{VR}{Virtual Reality}

\usepackage{booktabs}
\usepackage[table,dvipsnames]{xcolor}

\usepackage{dirtree}

\usepackage{graphicx}

\usepackage{subcaption}

\usepackage{algorithm}
\usepackage{algorithmic}
\renewcommand{\algorithmiccomment}[1]{\hfill$\rhd$#1}

\usepackage[acronym]{glossaries}
\newacronym{nocs}{NOCS}{Normalized Object Canonical Space}

\usepackage{xspace}
\makeatletter
\DeclareRobustCommand\onedot{\futurelet\@let@token\@onedot}
\def\@onedot{\ifx\@let@token.\else.\null\fi\xspace}

\def\eg{\emph{e.g}\onedot} 
\def\ie{\emph{i.e}\onedot}

\def\etal{\emph{et al}\onedot}
\makeatother

\providecommand{\norm}[1]{\ensuremath{\left\lVert#1\right\rVert}}

\renewcommand{\aa}{\mathbf{a}}

\providecommand{\oo}{\mathbf{o}}

\providecommand{\tt}{\mathbf{t}}

\providecommand{\vv}{\mathbf{v}}

\title{\LARGE \bf
    BiFold: Bimanual Cloth Folding with Language Guidance
}

\author{
Oriol Barbany$^{1}$
\hspace{4ex} Adrià Colomé$^{1}$
\hspace{4ex} Carme Torras$^{1}$\\
\normalsize{${}^{1}$Institut de Robòtica i Informàtica Industrial, CSIC-UPC} \\ {\tt\small \{obarbany,acolome,torras\}@iri.upc.edu} \\[.5em]
\url{https://barbany.github.io/bifold}
}

\begin{document}

\thispagestyle{empty}
\pagestyle{empty}

\twocolumn[{%
\renewcommand\twocolumn[1][]{#1} %
\maketitle
\thispagestyle{empty}
\begin{center}
    \centering
    \vspace{-2em}
    \captionsetup{type=figure}
    \includegraphics[width=.8\textwidth]{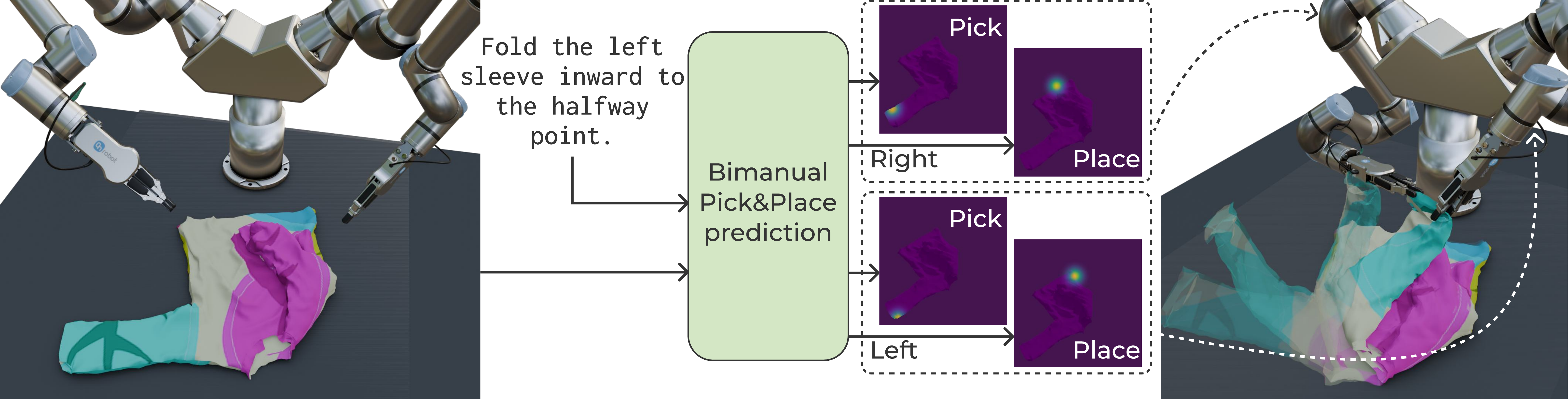} 
    \captionof{figure}{\textbf{Overview}: This paper introduces a language-conditioned model that predicts actions for bimanual cloth folding. Leveraging an observation of the cloth state and a text instruction, the model generates probability distributions for the pick and place positions of the left and right arms. To address the lack of datasets for this task, we present an annotation pipeline that augments an existing dataset of human demonstrations in a simulated environment with aligned language instructions.}
    \label{fig:teaser}
    \vspace{-0.5em}
\end{center}
}]

\begin{abstract}
    Cloth folding is a complex task due to the inevitable self-occlusions of clothes, their complicated dynamics, and the disparate materials, geometries, and textures that garments can have.
    In this work, we learn folding actions conditioned on text commands. Translating high-level, abstract instructions into precise robotic actions requires sophisticated language understanding and manipulation capabilities.
    To do that, we leverage a pre-trained vision-language model and repurpose it to predict manipulation actions. Our model, BiFold, can take context into account and achieves state-of-the-art performance on an existing language-conditioned folding benchmark.
    To address the lack of annotated bimanual folding data, we introduce a novel dataset with automatically parsed actions and language-aligned instructions, enabling better learning of text-conditioned manipulation. BiFold attains the best performance on our dataset and demonstrates strong generalization to new instructions, garments, and environments.
\end{abstract}

\vspace{-1em}

\section{INTRODUCTION}

Manipulation of garments remains a challenging problem due to the high flexibility of textiles and their nearly infinite number of degrees of freedom. We focus on cloth folding, a fundamental task in daily activities that presents significant challenges due to the variability of cloth materials, the need for precise manipulation, and the intricate sequences of actions involved.
Robotic cloth folding has broad applications in industrial automation, domestic assistance, and healthcare. Automating this task can enhance productivity and free humans from repetitive and labor-intensive activities.

 When learning folding actions from demonstrations, an additional difficulty stems from the variability of preferred folding steps, which can all achieve the desired output with a different series of actions. For example, even a simple rectangular cloth can be folded in various ways, \eg, in halves or thirds or joining the adjacent corners of the shorter or the longer edge \cite{household_dataset}. Grasping points for a rectangular towel are usually located at the corner, but for more complex garments such as T-shirts, these strictly depend on the manipulation strategy \cite{benchmarking_bimanual}. On top of that, folding garments with more complex geometries and topologies leads to a repertoire of even more folding strategies that depend on personal preferences.
 \looseness=-1


Cloth folding involves handling a highly deformable object with complex dynamics, which makes predicting and controlling its state challenging. Doing so requires accounting for the variability in cloth properties and the complexity of human-like folding techniques. For dealing with the inherent ambiguities of cloth folding actions, previous methods rely on conditioning signals such as goal images \cite{mo2022foldsformer} or language instructions \cite{shridhar2021cliport,language_deformable_manipulation}. However, these works focus on unimanual manipulations, limiting the efficiency and accuracy of the manipulation and being unable to be applied to human demonstrations, typically using both hands for folding clothes. One of the reasons the latest works focus on unimanual goal-conditioned folding are hardware constraints and the lack of adequate datasets with detailed language annotations, which also limits the ability to train and evaluate language-based models. This work directly tackles the latter.

This paper proposes BiFold, a novel method for bimanual cloth folding using language-specified tasks. We depict the overview of the model in \cref{fig:teaser}. Our approach uses a transformer-based model to fuse information from different modalities and leverages a frozen language component to enable the robot to handle the variability of human language and understand text instructions. The model outputs pick and place positions, which then can be provided to robot-specific motion primitives and used to operate two robotic arms in tandem. To train this model, we contribute by augmenting an existing dataset of human cloth folding demonstrations with language instructions. Instead of relying on hand-crafted annotations as in previous works, we develop an automatic process to parse folding demonstrations and annotate them with precise language descriptions.

The main contributions of this paper are:
\begin{itemize}
    \item A new model that leverages a foundational vision-language backend to learn cloth folding manipulations taking previous actions into account (\cref{sec:model}).
    \item A novel dataset of language-aligned bimanual cloth folding actions obtained with an automatic pipeline that could be applied to other datasets (\cref{sec:dataset_annotation}).
    \item We demonstrate the superiority of our model on an existing unimanual benchmark, the newly introduced bimanual dataset, and on a real-world setup (\cref{sec:experiments}).
\end{itemize}

\section{RELATED WORK}
\label{sec:related_work}

\subsection{Unconditioned cloth folding}

A typical approach to predict folding actions is to resort to action primitives and infer their parameters, \eg, the starting and end locations of a pick-and-place primitive. SpeedFolding \cite{speedfolding} uses a U-Net \cite{unet} to predict value maps for different gripper orientations from an RGB image and either uses the T-shirt 2-second fold method\footnote{\scriptsize \url{https://www.wikihow.com/Fold-a-T-Shirt-in-Two-Seconds}} or a method combining flings and folds needing prior knowledge about the dimensions of the manipulated object.

Cloth Funnels \cite{canberk2022clothfunnels} applies a policy for canonical alignment to leave the garment in a known configuration and predicts value maps from a predefined set of scales and rotations to augment the input RGB image. While it can work with different clothes, it needs an independent model for each category.
\looseness=-1

UniFolding \cite{xue2023unifolding} obtains a segmented point cloud, downsamples it to a fixed size of points, and uses it to predict a primitive action and its pick and place positions. The gripper's position is determined as a point in the sampled point cloud, hence not allowing the pick and place position to be on a non-sampled point. Additionally, the predicted positions cannot fall outside the object mask, meaning that the model cannot replicate some valid place positions. 

\subsection{Goal-conditioned cloth folding}

CLIPort \cite{shridhar2021cliport} leverages CLIP \cite{clip} to provide a broad semantic understanding of images conditioned on text, and Transporter \cite{zeng2020transporter}, which uses input image augmentations similar to Cloth Funnels \cite{canberk2022clothfunnels}, to achieve spatial precision in predicting pick and place positions. Foldsformer \cite{mo2022foldsformer} uses sub-goal observations for conditioning the predictions and fuses temporal and spatial information using attention \cite{transformers}.
FoldsFormer learns from a dataset of rectangular clothes and folds that are either random actions biased towards picking the corners or expert demonstrations. Conditioning with goal images can result in over-specified actions with irrelevant factors, such as the final position of the garment and its visual appearance. These approaches are incompatible with unseen clothes and cannot generalize to new tasks \cite{language_deformable_manipulation}.

Deng \etal \cite{language_deformable_manipulation} learn pick-and-place actions using a transformer encoder that takes depth images and the embeddings from a frozen CLIP text encoder \cite{clip}. Similarly to UniFolding \cite{xue2023unifolding}, this model predicts pick positions from a down-sampled point cloud but samples the place positions on the pixel space.
Deng \etal trains on perfect depth and segmentation maps, hence discarding color information that can provide useful cues for state estimation and experimenting a large reality gap when the inputs come from real noisy sensors. Moreover, this method works with only 200 points, which effectively fails to cover relevant parts of the manipulated garment, and uses fixed spatial thresholds to create a visual connectivity graph \cite{linLearningVisibleConnectivity2021} that makes the approach highly dependent on scale.

There are other options to specify folding actions other than images and text. For example, SpeedFolding \cite{speedfolding} can take enumerated user-specified folding lines drawn on top of an image of the initial configuration of the garment and used along a segmentation mask to define pick and place positions.\looseness=-1

\subsection{VR-Folding dataset}

VR-Garment \cite{garment_tracking} is a pipeline for simulated data collection using a \gls*{vr} headset and gloves that can track finger positions. Similarly to other works leveraging \gls*{vr} settings for data capture \cite{vr_dataset_julia}, VR-Garment uses Unity and the Obi particle physics simulator for a plausible cloth simulation model. The public VR-Folding dataset \cite{garment_tracking}, a garment manipulation dataset with flattening and folding actions performed by humans, and the private data collection of Unifolding \cite{xue2023unifolding} use VR-Garment.

The folding task starts with a flattened T-pose garment, and a volunteer performs pick-and-place actions with both hands until the garment is folded. The authors of VR-Folding ask the volunteers to follow a predefined set of instructions: For long-sleeved shirts, the first two actions fold the sleeves, and the last action folds the trunk. For all the other clothes, the garment is folded in half along the left and right direction and then in half along the up and down direction. 

\section{METHODOLOGY}
\label{sec:method}

\subsection{Learning language-conditioned pick and place positions}
\label{sec:model}

\begin{figure*}
    \centering
    \includegraphics[width=.9\linewidth]{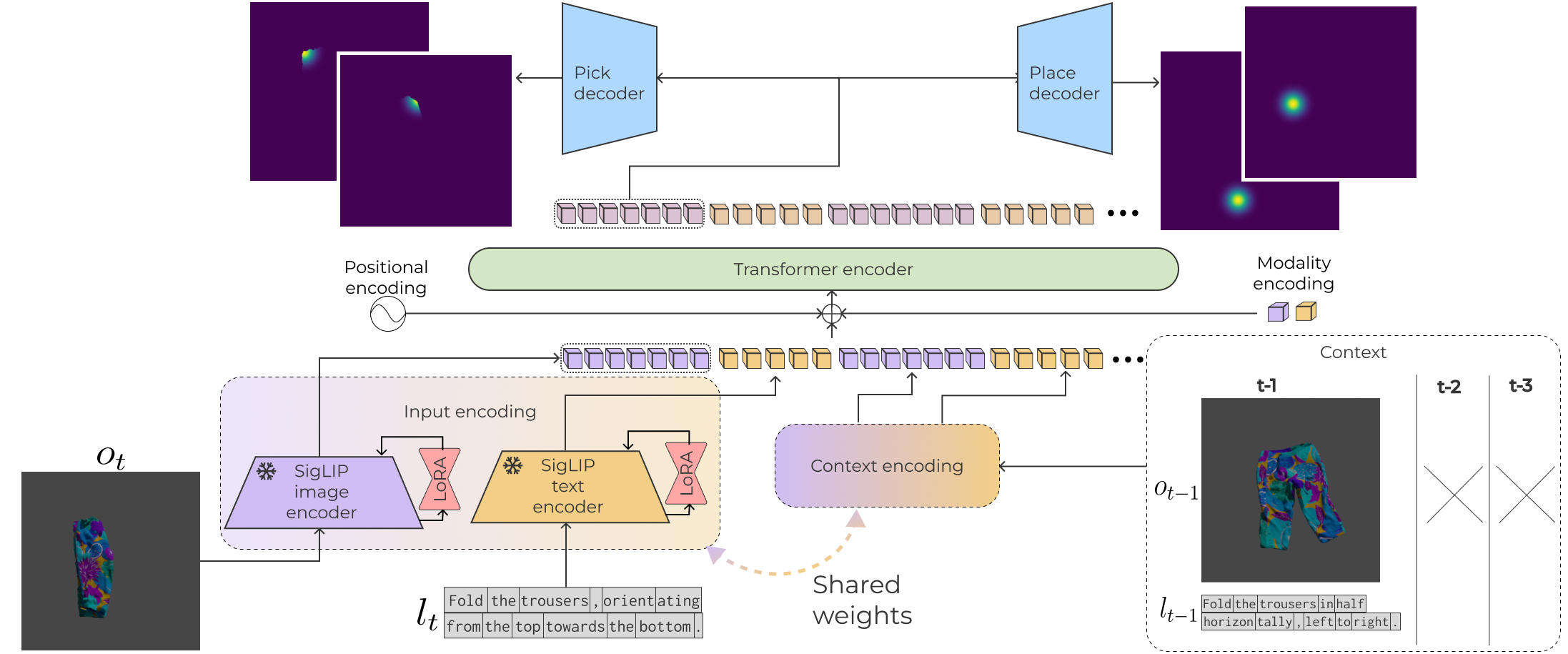}    \caption{\textbf{BiFold model architecture:} We use a frozen SigLIP \cite{siglip} model and adapt it using LoRA \cite{hu2022lora} to obtain tokens from an RGB image and an input text. The same encoders are used to incorporate past observations that provide context to the model. The domain of each token is indicated using a modality encoding and the sequence order using positional encodings. The concatenated sequence is processed using a transformer encoder \cite{transformers}, which fuses the different modalities. The output of the transformer is used to determine the pick and place positions for the left and right arms using convolutional decoders.}
    \label{fig:architecture}
    \vspace{-1.5em}
\end{figure*}

Given an RGB image observation $\oo_t$ and a language instruction $\ell_t$, we want to design a policy $\pi_\theta(\aa_t | \ell_t, \oo_t)$ parametrized by $\theta$ to obtain manipulation actions $\aa_t$ at step $t$. In this work, we constrain $\aa_t$ to a quasi-static pick-an-place manipulation (see the \href{https://barbany.github.io/bifold/}{project page} for details of the manipulation primitive).
While some prior works incorporate additional actions such as flings and drags \cite{speedfolding}, our approach focuses solely on pick-and-place operations, as these are sufficient for garment folding  and are the only actions present in the unimanual and bimanual datasets used to train the policy. Moreover, quasi-static actions simplify cloth manipulations since rapid movements are more unpredictable, making it hard to obtain optimal folding results. Instead, the robot can better control the cloth's shape and prevent unwanted wrinkles by applying slow and controlled movements.

We leverage the SigLIP model \cite{siglip} to obtain high-level representations of $\oo_t$ and $\ell_t$. SigLIP is a pre-trained contrastive vision-language model similar to CLIP \cite{clip} which consists of image and text transformer-based encoders that provide aligned features. Compared to other vision backbones  \cite{Open_Clip_2023_CVPR,sun2023evaclipimprovedtrainingtechniques,oquab2024dinov,mae_2022} it offers better visual representations \cite{tong2024cambrian}, presents better 3D awareness \cite{Probing_3D_2024_CVPR}, offers superior visually-situated text understanding capabilities \cite{beyer2024paligemma,chen2023pali3visionlanguagemodels}, and suffers a lower sim-to-real gap in robotics setups \cite{SPOC_2024_CVPR}.

We fuse the image and language features by concatenating them and applying attention layers \cite{transformers}. This is an increasingly popular approach applied to multimodal models \cite{4m,4m21} and robotics \cite{octo_2023,rt2}. By using SigLIP, the features are aligned before the fusion step similarly to ALBEF \cite{albef}.

We then feed the processed observation tokens as input to convolutional-based decoders. While the decoder can also be based on transformers \cite{mae_2022}, we found that these yield patch artifacts in low data regimes. The decoders yield probability distributions over the observation space used to sample pick and place locations. Given that the pick position has to fall on the garment, we use a segmentation mask to constrain the support of the predicted distributions. Given the output probabilities, we greedily sample the most probable position to obtain the pick-and-place action $\aa_t$.

To adapt the SigLIP encoders to this task while retaining its knowledge, we apply \gls*{lora} \cite{hu2022lora} to the query and value matrices of the attention layers, keeping all other parameters frozen. We fit the parameters $\theta$ by minimizing the binary cross-entropy between the predicted probability distribution and a Gaussian distribution centered at the ground truth position and with variance $\Sigma=5\cdot I$ as in previous works \cite{mo2022foldsformer,language_deformable_manipulation}. In case there are multiple pick and place positions, which happens for the VR-Folding dataset as can be seen in \cref{fig:dataset_samples_multipoints}, we define a Gaussian mixture model with equal weights centered in all the correct positions and normalize the result so that the maximum is 1.

Folding a cloth requires at least two steps. While the text instructions indicate the pick and place positions in natural language, reducing the inherent ambiguity due to the usual garment symmetry, some late manipulation steps may be ill-defined given the current observation. For example, observe how it is impossible to tell top and bottom apart in \cref{fig:example_need_context}, motivating the use of a policy $\pi_\theta(\aa_t | \ell_t, \oo_t, \oo_{t-1}, \dots, \oo_{t-H})$ that uses a context of $H$ previous observations.
To incorporate context, we encode the previous observations using the same encoders and add a learned positional encoding element-wise for the model to differentiate the time steps. We set a fixed context size $H=3$, as over 95\% of the bimanual dataset samples consist of three or fewer actions. This choice balances capturing sufficient historical context while maintaining efficiency and being enough for longer manipulations.  We use attention masking to filter out empty tokens if the context does not fill the maximum allowed observation horizon. The pick and place decoders still take the tokens corresponding to $\oo_t$ as input, and the output tokens of $\oo_{t-1}\dots, \oo_1$ and $\ell_t$ are discarded. We depict the architecture of our model in \cref{fig:architecture}, which we dub BiFold.

\subsection{Creating aligned language instructions}
\label{sec:dataset_annotation}

\begin{figure}[t]
    \centering
    \begin{subfigure}[t]{0.32\linewidth}
        \centering
        \includegraphics[width=\linewidth]{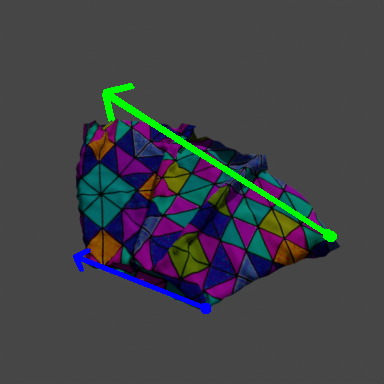}
        \caption{\scriptsize Fold the skirt in half horizontally, right to left.}
    \end{subfigure}
    \begin{subfigure}[t]{0.32\linewidth}
        \centering
        \includegraphics[width=\linewidth]{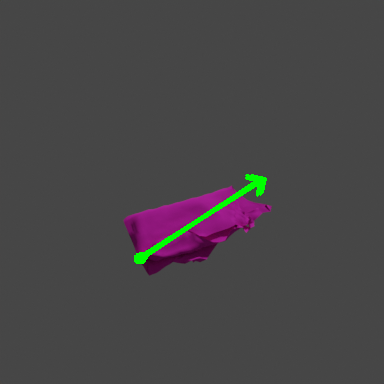}
        \caption{\scriptsize  Fold the top, making sure the bottom right side touches the top right side only using the right arm.}
        \label{fig:example_need_context}
    \end{subfigure}
    \begin{subfigure}[t]{0.32\linewidth}
        \centering
        \includegraphics[width=\linewidth]{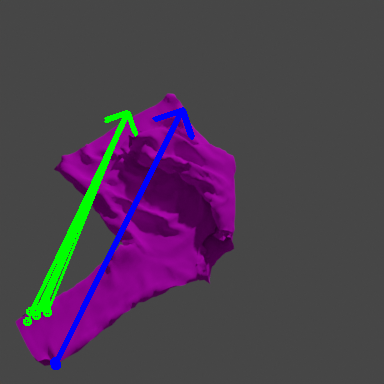}
        \caption{\scriptsize Fold the left sleeve inward to the halfway point.}
        \label{fig:dataset_samples_multipoints}
    \end{subfigure}
    \caption{\textbf{Dataset samples:} Examples of language-aligned bimanual cloth folding instructions obtained using our proposed annotation pipeline. Pick and place positions for {\color{green}\textbf{right}} and {\color{blue}\textbf{left}} actions are represented as the origin and endpoints of an arrow. Each action uses eight vertices, which might be distinct and fall into different pixels.}
    \label{fig:dataset_samples}
    \vspace{-2em}
\end{figure}

The VR-Folding dataset contains almost 4$k$ bimanual folding demonstrations from humans. We segment around 7$k$ actions and obtain aligned text instructions. Our annotation method generates diverse set of language instructions like \cref{fig:dataset_samples}, totaling more than 1$k$ unique prompts.

\textbf{Preprocessing: }All garments in VR-Folding have a green pattern for the exterior of the mesh and an orange color for the interior,
hindering generalization to other textures. For this reason, we re-render RGB-D images using the simulation meshes and textures of the CLOTH3D dataset \cite{cloth3d}. The VR-Folding dataset contains some instances in which the simulator became unstable and yielded unrealistic clothes, resulting in sharp and spiked meshes. We filter these out by thresholding the ratio between the maximum z-score of the edge length distributions of the simulation mesh and the \gls*{nocs} mesh \cite{nocs}.\looseness=-1

\textbf{Parsing actions: }We remove spurious actions, \ie, actions performed in the span of 5 frames or less or whose distance is less than 0.1 m. Given that the same action may have slightly different start and end times, we align the left and right actions that overlap in time, considering that some actions only use one hand. This event is not considered in the predefined actions provided to the volunteers but occurs in the dataset. Several volunteers obviate these instructions and hence assigning language instructions is non trivial.

\textbf{Semantic grasp locations: }We identify garment parts in the canonical space and applying predefined thresholds, similar to PoseScript \cite{posescript}. We infer the semantic place location by finding the nearest neighbors of the simulation mesh on the \gls*{nocs} \cite{nocs} at the start of the action and the final position of the picker in the observation space. We fuse the information from both pickers following a heuristics that provides the type of action and the variables of the language template.
\looseness=-1

\textbf{Grasp in natural language: }We define different instruction templates for folding sleeves (\eg, \texttt{"Fold the right sleeve towards the inside."}), performing generic folds from a semantic location to another (\eg, \texttt{"Fold the Skirt in half, from left to right."}), and refining the position of a garment (\eg, \texttt{"Ensure the bottom part of the Top is well-positioned."}). Unlike Deng \etal \cite{language_deformable_manipulation}, we parametrize the prompts with different garments and locations obtained automatically. If the action only uses one arm, we append \texttt{"only using the \{left/right\} arm"}.

\section{EXPERIMENTS}
\label{sec:experiments}

\begin{table*}[t]
    \begin{minipage}{.6\linewidth}
        \centering
        \small
        \caption{\textbf{Unimanual simulation results: } This table reports the average success rates (\%) on testing tasks of the dataset by Deng \etal \cite{language_deformable_manipulation}. All models have two variants trained with either 100 or 1000 demonstrations per task. The best performance is in \textbf{bold} and second-best is \underline{underlined}.}
        \resizebox{%
          \ifdim\width>\linewidth
            \linewidth
          \else
            \width
          \fi
        }{!}{
        \begin{tabular}{lcccccccccc}
            \toprule
            \# demonstrations $\rightarrow$ & 100 & 1000& 100 & 1000& 100 & 1000& 100 & 1000& 100 & 1000 \\
            Method $\downarrow$ / Task $\rightarrow$ & \multicolumn{2}{c}{Corner} & \multicolumn{2}{c}{Triangle}  & \multicolumn{2}{c}{Half}  & \multicolumn{2}{c}{T-shirt} & \multicolumn{2}{c}{Trousers}  \\
            \hline
            \rowcolor[HTML]{E3E3E3} \multicolumn{11}{c}{\textbf{Seen instructions}} \\
            \hline
            Foldsformer~\cite{mo2022foldsformer} & {80.0} & {90.0} & 52.0 & 68.0 & 16.0 & 26.0 & 18.0 & 24.0 & 8.0 & 20.0 \\
            CLIPORT~\cite{shridhar2021cliport} & 78.0 & 86.0 & 76.0 & 78.0 & 50.0 & 56.0 & 54.0 & 74.0 & 38.0 & 46.0 \\
            Deng \etal \cite{language_deformable_manipulation} & \textbf{100.0} & \textbf{100.0} & 72.0 & \underline{92.0} & 52.0 & \textbf{74.0} & \underline{86.0} & 84.0 & 74.0 & 86.0 \\
            \midrule
            BiFold w/o context & \textbf{100.0} & \textbf{100.0} & \underline{80.0} & \textbf{93.3} & \textbf{64.7} & \underline{70.0} & 82.5 & \textbf{94.0} & \underline{79.3} & \textbf{91.3} \\
            BiFold & \textbf{100.0} & \textbf{100.0} & \textbf{84.0} & 91.0 &  \underline{55.3} & 69.3 &  \textbf{88.5} & \underline{93.5} & \textbf{87.3} & \underline{90.7} \\
            \hline
            \rowcolor[HTML]{E3E3E3} \multicolumn{11}{c}{\textbf{Unseen instructions}} \\
            \hline
            CLIPORT~\cite{shridhar2021cliport} & 80.0 & \underline{90.0} & 76.0 & 74.0 & 38.0 & 50.0 & 56.0 & 70.0 & 32.0 & 38.0 \\
            Deng \etal \cite{language_deformable_manipulation} & \textbf{96.0} & \textbf{100.0} & 80.0 & \textbf{92.0} & \underline{40.0} & \underline{68.0} & 80.0 & 80.0 & 74.0 & \textbf{92.0} \\
            \midrule
            BiFold w/o context & \textbf{96.0} & 73.5 & \textbf{85.0} & \underline{91.0} & 36.0 & 66.7 & \underline{91.0} & \textbf{93.5} & \underline{78.0} & \underline{89.3} \\
            BiFold & 77.0 & 62.5 & \underline{82.0} & \underline{91.0} &  \textbf{51.3} & \textbf{69.3} &  \textbf{93.0} & \underline{91.5} & \textbf{81.3} & 88.7 \\
            \hline
            \rowcolor[HTML]{E3E3E3} \multicolumn{11}{c}{\textbf{Unseen tasks}} \\
            \hline
            CLIPORT~\cite{shridhar2021cliport} & 70.0 & 76.0 & 70.0 & 74.0 & 0.0 & 0.0 & 22.0 & 46.0 & 0.0 & 8.0 \\
            Deng \etal \cite{language_deformable_manipulation} & 36.0 & {78.0} & 76.0 & 88.0 & 0.0 & 0.0 & \textbf{42.0} & \textbf{80.0} & 0.0 & \underline{32.0} \\
            \midrule
            BiFold w/o context & \underline{94.0} & \textbf{100.0} & \underline{78.0} & \underline{90.0} & \underline{27.3} & \textbf{34.7} & 20.5 & \underline{26.5} & \textbf{6.0} & 26.7 \\
            BiFold & \textbf{100.0} & \textbf{100.0} & \textbf{88.0} & \textbf{93.0} &  \textbf{28.0} & \underline{30.7} &  \underline{33.0} & 18.0 & \underline{4.0} & \textbf{76.7} \\
            \bottomrule
        \end{tabular}
        }
        \label{tab:experiments_single}
    \end{minipage}
    \begin{minipage}{.39\linewidth}
        \centering
        \small
        \caption{\textbf{Bimanual results on test images.}}
        \vspace{-0.5em}
        \resizebox{%
          \ifdim\width>\linewidth
            \linewidth
          \else
            \width
          \fi
        }{!}{
        \begin{tabular}{lccc}
            \toprule
            & $\text{AP}_{5}$ ($\uparrow$)& $\text{AP}_{10}$ ($\uparrow$)& $\text{AP}_{20}$ ($\uparrow$) \\
            \midrule
            Deng \etal \cite{language_deformable_manipulation} & 3.8 & 8.6 & 18.6 \\
            BiFold w/o context & {40.3} & {55.7} & {72.7} \\
            BiFold & \textbf{75.1} & \textbf{92.8} & \textbf{97.3}\\
            \midrule
            & $\text{AP}_{50}$ ($\uparrow$) & KP-MSE ($\downarrow$) & Quantile (\%) ($\uparrow$) \\
            \midrule
            Deng \etal \cite{language_deformable_manipulation} & 40.3 & 5.5 & 86.0 \\
            BiFold w/o context & {88.9} & \textbf{1.1} & {97.1} \\
            BiFold & \textbf{98.9} & {2.5} & \textbf{98.8} \\
            \bottomrule
        \end{tabular}
        }
        \label{tab:experiments_bimanual_vr_folding}
        \caption{\textbf{Bimanual results on SoftGym.}}
        \vspace{-0.5em}
        \resizebox{%
          \ifdim\width>\linewidth
            \linewidth
          \else
            \width
          \fi
        }{!}{
        \begin{tabular}{lcccc}
            \toprule
            Method $\downarrow$ / Task $\rightarrow$ & Skirt & Top & Trousers & T-shirt \\
            \hline
            \rowcolor[HTML]{E3E3E3} \multicolumn{5}{c}{\textbf{Error (mm) ($\downarrow$)}} \\
            \hline
            Deng \etal \cite{language_deformable_manipulation} & 63.4 & 68.5 & 94.0 & 65.3 \\
            BiFold w/o context & {41.3} & {38.7} & {44.2} & {33.6} \\
            BiFold & \textbf{32.9} & \textbf{35.3} & \textbf{30.4} & \textbf{26.3} \\
            \hline
            \rowcolor[HTML]{E3E3E3} \multicolumn{5}{c}{\textbf{mIoU ($\uparrow$)}} \\
            \hline
            Deng \etal \cite{language_deformable_manipulation} & 49.1 & 32.7 & 42.7 & 51.8 \\
            BiFold w/o context & {64.1} & {54.8} & {70.0} & {74.2} \\
            BiFold & \textbf{71.9} & \textbf{57.3} & \textbf{77.6} & \textbf{80.3} \\
            \hline
            \rowcolor[HTML]{E3E3E3} \multicolumn{5}{c}{\textbf{$\text{Success}_{\text{IoU}\geq 80}$ (\%) ($\uparrow$)}} \\
            \hline
            Deng \etal \cite{language_deformable_manipulation} & 9.3 & 0.0 & 2.9 & 6.9 \\
            BiFold w/o context & \textbf{34.9} & {10.0} & {45.9} & {50.6} \\
            BiFold & 31.7 & \textbf{10.5} & \textbf{48.5} & \textbf{60.5} \\
            \bottomrule
        \end{tabular}
        }
        \label{tab:experiments_bimanual_softgym}
    \end{minipage}
    \vspace{-1em}
\end{table*}

\subsection{Experimental setup}

We train our models with a batch size of 24 using Adam \cite{adam} with a learning rate $10^{-4}$ for 100 epochs. We use \gls*{lora} \cite{hu2022lora} with rank 8, $\alpha=32$ and dropout 0.01. The transformer encoder has 8 blocks and 16 heads. We use the SigLIP \cite{siglip} model with patch size 16. We perform SE(3) augmentations to the input observations during training and standard image normalization. For our dataset, we use 90\% of samples for training and the rest for testing. The model is trained on a single NVIDIA GeForce RTX 3090 GPU.

\textbf{Simulation evaluation: }We load the cloth meshes to the SoftGym simulator \cite{lin_2021_softgym} and execute a pick and place primitive with one or two grippers. The unimanual dataset \cite{language_deformable_manipulation} is obtained with SoftGym and can be simulated straightforwardly. To simulate the bimanual dataset, we transform the workspace of the VR data collection setup to the one used for the unimanual dataset and make sure that meshes in the initial configuration do not diverge due to unsatisfied constraints.\looseness=-1

\textbf{Real world evaluation: }We obtain zenithal view RGB-D images of different clothes with an Azure Kinect camera. We use SAM \cite{sam_Kirillov_2023_ICCV} for obtaining segmentation masks of the cloth from RGB images. See more details in the \href{https://barbany.github.io/bifold/}{project page}.  Similarly to previous work \cite{language_deformable_manipulation,mo2022foldsformer,GPTFabric2024,weng2021fabricflownet}, we assume that the cloth starts from a flattened state as this is the initial state for all the samples of the training dataset. Note that the dataset contains some refinement actions on states similar to an initial crumpled configuration. However, the language instruction in this case specifies which part to correct and is never the first folding action. If the initial cloth is crumpled, we can use a flattening model \cite{seita_fabrics_2020,ha2021flingbot} to achieve the desired configuration before starting the folding task.\looseness=-1

\subsection{Unimanual cloth folding}
\label{sec:unimanual_experiments}

Given the lack of language-guided bimanual cloth folding benchmarks, we first compare the performance of BiFold on single-arm manipulation. Following Deng \etal, we report the average success rate, defined based on the vertex-to-vertex distance between the obtained and ground truth cloth meshes. Concretely, the state of the manipulated garment after applying a pick-and-place action is compared to that obtained when generating the dataset with an oracle manipulation. Then, we consider an action successful if the mean Euclidean distance between the vertices of these meshes is less than the diameter of a particle in SoftGym \cite{lin_2021_softgym}, \ie, 0.0125 m. \cref{tab:experiments_single} presents the results of this experiment, where we take the baseline results from Deng \etal \cite{language_deformable_manipulation}.

In \cref{tab:experiments_single}, unseen instructions are paraphrases, \ie, language instructions obtained from templates that are not in the training dataset, and unseen tasks add complexity on top of that by introducing new semantic locations of the manipulated object. For example, if the training set contains top-left, top-right, and bottom-left folds, a bottom-right fold is considered an unseen task. Foldsformer \cite{mo2022foldsformer} is only evaluated on the first setting as it uses subgoal images as conditioning, not text. As we can see, BiFold without context achieves the best overall performance, surpassing Deng \etal \cite{language_deformable_manipulation} by up to 64 percentage points. When adding context to the input, we can observe a general performance increase but at the expense of a higher complexity due to the quadratic scaling of transformers with the input size.

\subsection{Bimanual cloth folding}
\label{sec:bimanual_experiments}

\begin{figure*}[t]
    \centering
    \begin{subfigure}{\linewidth}
        \renewcommand\thesubfigure{\roman{subfigure}}
        \setcounter{subfigure}{0}
        \begin{subfigure}[t]{0.19\linewidth}    
            \centering
            \includegraphics[width=\linewidth]{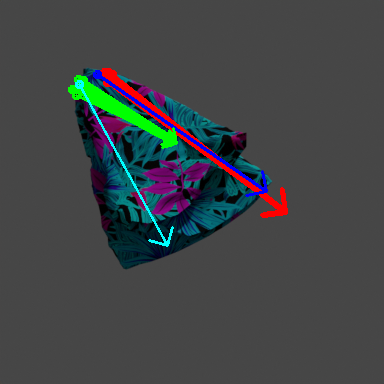}
            \caption{\scriptsize Fold the skirt, making a crease from the top to the bottom.}
        \end{subfigure}
        \begin{subfigure}[t]{0.19\linewidth}
            \centering
            \includegraphics[width=\linewidth]{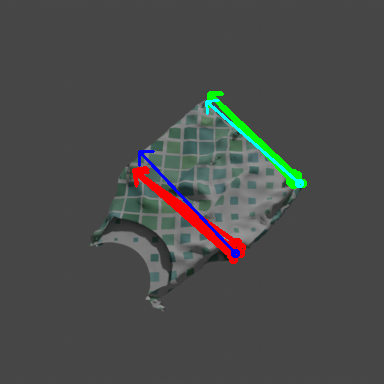}
            \caption{\scriptsize Bend the top in half, from bottom right to bottom left.}
        \end{subfigure}
        \begin{subfigure}[t]{0.19\linewidth}
            \centering
            \includegraphics[width=\linewidth]{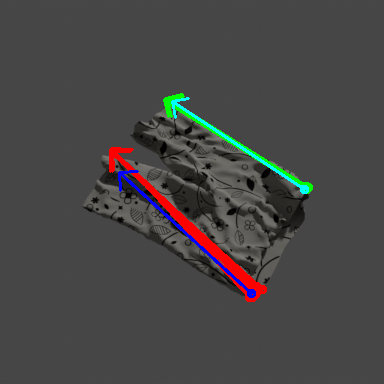}
            \caption{\scriptsize Fold the trousers from the top side towards the bottom side.}
        \end{subfigure}
        \begin{subfigure}[t]{0.19\linewidth}
            \centering
            \includegraphics[width=\linewidth]{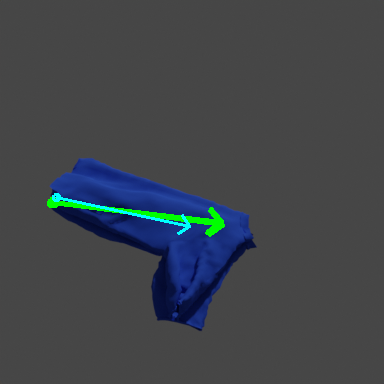}
            \caption{\scriptsize Fold the left sleeve to the centerline of the shirt only using the left arm.}
        \end{subfigure}
        \begin{subfigure}[t]{0.19\linewidth}
            \centering
            \includegraphics[width=\linewidth]{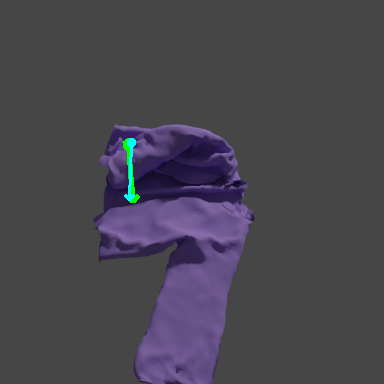}
            \caption{\scriptsize Fold the right sleeve towards the body only using the right arm.}
        \end{subfigure}
        \setcounter{subfigure}{0}
        \renewcommand\thesubfigure{\alph{subfigure}}
        \vspace{-0.5em}
        \caption{\textbf{Simulation dataset: }BiFold can replicate unseen actions on new clothes almost perfectly (ii, iii), perform more optimal actions than human volunteers (i), and understand when a single hand should be used (iv, v).}
        \label{fig:qualitative_sim}
    \end{subfigure}
    \begin{subfigure}{\linewidth}
        \renewcommand\thesubfigure{\roman{subfigure}}
        \setcounter{subfigure}{0}
        \begin{subfigure}{.19\linewidth}
            \centering
            \includegraphics[width=\linewidth]{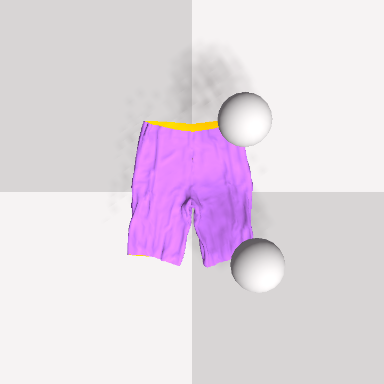}
        \end{subfigure}
        \begin{subfigure}{.19\linewidth}
            \centering  
            \includegraphics[width=\linewidth]{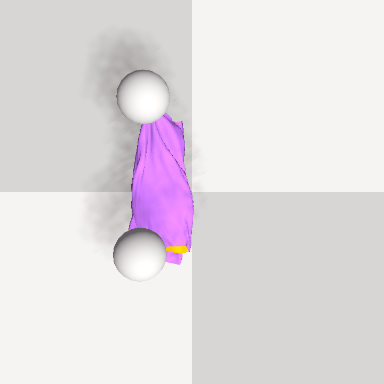}
        \end{subfigure}
        \begin{subfigure}{.19\linewidth}
            \centering  
            \includegraphics[width=\linewidth]{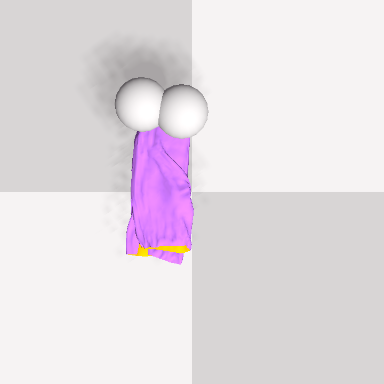}
        \end{subfigure}
        \begin{subfigure}{.19\linewidth}
            \centering  
            \includegraphics[width=\linewidth]{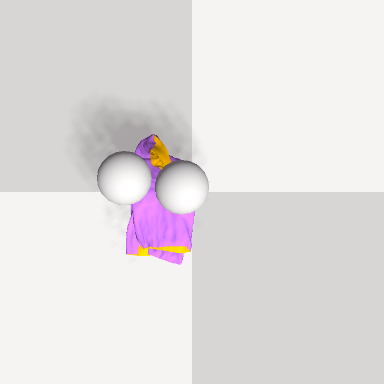}
        \end{subfigure}
        \begin{subfigure}{.19\linewidth}
            \centering  
            \includegraphics[width=\linewidth]{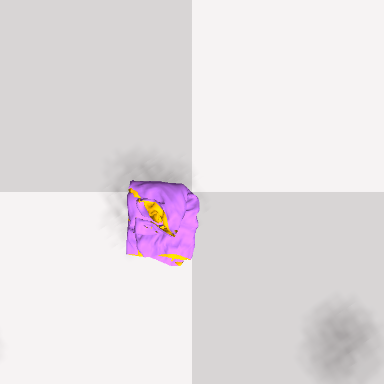}
        \end{subfigure}
        \setcounter{subfigure}{1}
        \renewcommand\thesubfigure{\alph{subfigure}}
        \caption{\textbf{End-to-end bimanual folding: }Example of an end-to-end folding rollout using BiFold with context trained on the bimanual dataset.}
        \label{fig:qualitative_bimanual_rollout}
    \end{subfigure}
    \begin{subfigure}{\linewidth}
        \centering
        \renewcommand\thesubfigure{\roman{subfigure}}
        \setcounter{subfigure}{0}
        \begin{subfigure}[t]{0.19\linewidth}
            \centering
            \includegraphics[width=\linewidth]{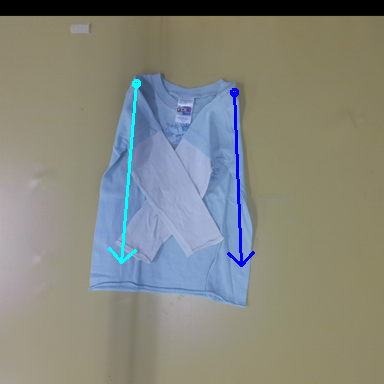}
            \caption{\scriptsize Fold the tshirt, top side over bottom side.}
        \end{subfigure}
        \begin{subfigure}[t]{0.19\linewidth}
            \centering
            \includegraphics[width=\linewidth]{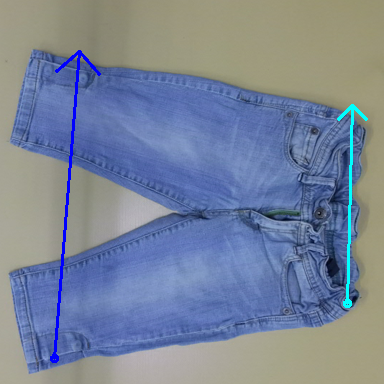}
            \caption{\scriptsize Fold the trousers, left side over right side.}
        \end{subfigure}
        \begin{subfigure}[t]{0.19\linewidth}
            \centering
            \includegraphics[width=\linewidth]{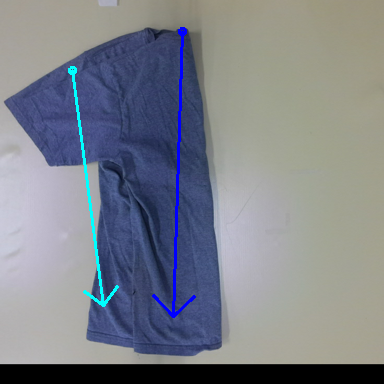}
            \caption{\scriptsize Fold the tshirt in half, with the top side overlapping the bottom.}
        \end{subfigure}
        \begin{subfigure}[t]{0.19\linewidth}
            \centering
            \includegraphics[width=\linewidth]{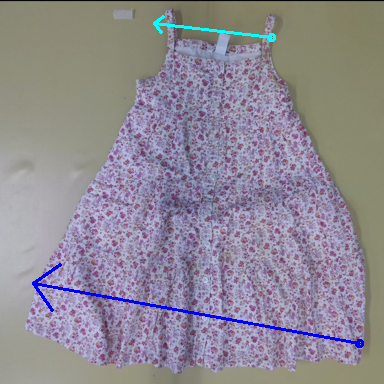}
            \caption{\scriptsize Fold the waistband of the dress in half, from left to right.}
        \end{subfigure}
        \begin{subfigure}[t]{0.19\linewidth}    
            \centering
            \includegraphics[width=\linewidth]{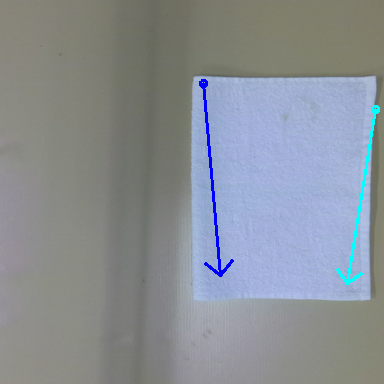}
            \caption{\scriptsize Create a fold in the towel, going from top to bottom.}
        \end{subfigure}
        \setcounter{subfigure}{2}
        \renewcommand\thesubfigure{\alph{subfigure}}
        \caption{\textbf{Real dataset: }BiFold can transfer well to seen (i, ii, iii) and unseen (iv, v) garment categories.}
        \label{fig:qualitative_real}
    \end{subfigure}
    \caption{\textbf{Qualitative examples:} Action predictions obtained with our model. Pick and place actions for right and left are represented as the origin and endpoints of arrows: {\color{red}\textbf{red}} and {\color{green}\textbf{green}} for ground truth, and {\color{blue}\textbf{blue}} and {\color{Aquamarine}\textbf{light blue}} for our model.}
    \label{fig:qualitative}
    \vspace{-1em}
\end{figure*}

We adapt the model from Deng \etal \cite{language_deformable_manipulation} to have pick and place predictions for two grippers instead of one. In \cref{tab:experiments_bimanual_vr_folding}, we report image-based metrics on the test partition of our dataset. Concretely, we use the common keypoint detection metrics of \gls*{ap} at different pixel thresholds and the \gls*{kp-mse}. We also report the quantile of the ground truth in the model output, \ie, how likely is the real manipulation in the predicted pick and place position distributions. Our model consistently achieves the best performance in all metrics and obtains qualitatively good results, as shown in \cref{fig:qualitative_sim}. Looking at the quantile measure, we can see that the average for Deng \etal is 86.0\%, being split into 94.3\% for the pixel-based place point detection and 77.7\% point cloud-based point selection, showing a better transferability of value maps and supporting our design choice. BiFold without context can achieve a slightly better \gls*{kp-mse} value than the version with context, which attains a mean error of only 2.5 pixels on a 384$\times$384 image. However, adding context yields better probability calibration and more localized predictions, almost doubling the average precision at 10 pixels.

We present the manipulation results on simulation in \cref{tab:experiments_bimanual_softgym} and a qualitative folding rollout that concatenates several actions in \cref{fig:qualitative_bimanual_rollout}. We train the models with data from a different renderer, so the sim-to-sim gap affects the performance. Moreover, our dataset contains many more meshes than the unimanual one, and some of them have more than twice the number of vertices of any in the dataset by Deng \etal \cite{language_deformable_manipulation}. For this reason, we report the error, mean \gls*{iou}, and the success defined by those instances for which the \gls*{iou} exceeds 80\%. Similarly to the image metrics, we can see that BiFold consistently outperforms the baseline. One of the reasons for the drop in performance regarding \cref{tab:experiments_single} is the use of human folding demonstrations, which introduces considerably more variability in the demonstrations than using scripted policies on simulation and in some cases are suboptimal (\cref{fig:qualitative_sim} (i)). Another reason is that Deng \etal \cite{language_deformable_manipulation} obtain pick positions by manually defining some vertices in the simulation mesh, greatly restricting the possible starting positions of the manipulation action. However, human demonstrators can grab any part of the garments. To showcase how BiFold transfers to real-world observations, we perform an offline qualitative evaluation on test images, which is presented in \cref{fig:qualitative_real}. Despite the difference in image appearances, lighting, and unseen garments, BiFold can predict coherent actions.

\subsection{Ablations}

We include some model ablations in \cref{tab:ablation}. We replace self-attention with cross-attention to tame the quadratic increase in time and memory with the growing token count, \eg, when using larger image resolutions, longer text, or more context. Attempting to provide multi-scale information and avoid losing relevant information of the original image, we also test a U-Net architecture \cite{unet} where the text information was introduced employing FiLM layers \cite{perez2018film}. Finally, we replace the SigLIP text encoder with the \texttt{t5-base} model \cite{t5_2020} used by the generalist robotics model Octo \cite{octo_2023}. Overall, BiFold attains the best average performance and offers a more stable success rate than its variants.

\begin{table}[th]
    \centering
    \small
    \caption{\textbf{Ablations: } Success rates (\%) on testing tasks of the dataset by Deng \etal \cite{language_deformable_manipulation} using 1000 demonstrations per task.}
    \resizebox{%
      \ifdim\width>\linewidth
        \linewidth
      \else
        \width
      \fi
    }{!}{
    \begin{tabular}{lccccc}
        \toprule
        Method $\downarrow$ / Task $\rightarrow$ & Corner & Triangle  & Half  & T-shirt & Trousers  \\
        \hline
        \rowcolor[HTML]{E3E3E3} \multicolumn{6}{c}{\textbf{Seen instructions}} \\
        \hline
        BiFold  & \textbf{100.0}  & \textbf{91.0} &  \textbf{69.3} & {93.5} & \textbf{90.7} \\
        Cross-attention & \textbf{100.0} & 88.0 & 66.7 & \textbf{98.0} & 88.0 \\
        Text-conditioned U-Net & \textbf{100.0} & 75.0 & 54.7 & 92.5 & 83.3 \\
        T5 text encoder & \textbf{100.0} & 90.0 & 67.3 & 93.5 & 88.0 \\
        \hline
        \rowcolor[HTML]{E3E3E3} \multicolumn{6}{c}{\textbf{Unseen instructions}} \\
        \hline
        BiFold & 62.5 & \textbf{91.0} & {69.3}& {91.5} & \textbf{88.7} \\
        Cross-attention & \textbf{98.0} & 89.0 & 64.0 & {91.5} & 87.3  \\
        Text-conditioned U-Net & 92.0 & 70.0 & 46.7 & 70.5 & 85.3 \\
        T5 text encoder & 78.0 & 86.0 & \textbf{73.3} & \textbf{92.5} & 88.0 \\
        \hline
        \rowcolor[HTML]{E3E3E3} \multicolumn{6}{c}{\textbf{Unseen tasks}} \\
        \hline
        BiFold  & \textbf{100.0} & \textbf{93.0}  & \textbf{30.7} & 18.0 & \textbf{76.7} \\
        Cross-attention & 53.5 & \textbf{93.0} & 26.7 & 13.0 & 3.3 \\
        Text-conditioned U-Net & 86.5 & 75.0 & {27.3} & \textbf{32.5} & 0.7 \\
        T5 text encoder & 99.5 & 90.0 & 26.0 & 25.5 & 20.7 \\
        \bottomrule
    \end{tabular}
    }
    \label{tab:ablation}
    \end{table}

\section{CONCLUSIONS}
\label{sec:conclusions}

In this paper, we propose BiFold, a model that uses a pre-trained vision-language model for bimanual cloth folding. By incorporating natural language processing, we bridge the gap between high-level human instructions and robotic execution, facilitating more nuanced and adaptive manipulation strategies. We achieve enhanced robustness to text modifications and visual changes by adopting a model pre-trained on large-scale data. While methods like RT-1 \cite{rt1} already incorporated context, they use consecutive observations that are generally highly redundant and provide little information for high enough frame rates. Instead, BiFold conditions the manipulation on \textbf{keyframes}, motivating the use of models that keep a memory of only the relevant previous observations. This work also presents a scalable pipeline for language annotation and provides a new dataset that constitutes a challenging benchmark.

BiFold learns folding subactions and we perform complete folds as in \cref{fig:qualitative_bimanual_rollout} by following a predefined list of ordered instructions, but future work could use a \gls*{llm}-based planner to break down the task into steps \cite{huang2023voxposer}. Even if BiFold predicts correct actions, the folding action can fail due to poor simulator physics \cite{sim2real_gap} or incorrect grasps. A natural next step could be to not rely on primitives and directly predict robot actions, \eg, use a diffusion policy \cite{chi2023diffusionpolicy} and condition it with the predicted pick and place actions \cite{ma2024hierarchical}. While the \gls*{nocs} approach is efficacious for CLOTH3D cloth assets \cite{cloth3d}, it may fail for clothes where topologies are too diverse in a given category (\eg, designer clothes with significant deviations from standard shapes). In this case, we could rely on distilled language features \cite{shen2023distilled}. Further experiments, as well as discussions on limitations and future work, can be found on the \href{https://barbany.github.io/bifold/}{project page}.

\section*{ACKNOWLEDGMENT}

This work was funded by project SGR 00514 (Departament de Recerca i Universitats de la Generalitat de Catalunya) and  CSIC project 202350E080
(ClothIRI). O.B. acknowledges travel support from ELISE (GA no 951847).

\bibliographystyle{IEEEtran}
\bibliography{IEEEabrv,bibliography}

\begin{thebibliography}{10}
\providecommand{\url}[1]{#1}
\csname url@samestyle\endcsname
\providecommand{\newblock}{\relax}
\providecommand{\bibinfo}[2]{#2}
\providecommand{\BIBentrySTDinterwordspacing}{\spaceskip=0pt\relax}
\providecommand{\BIBentryALTinterwordstretchfactor}{4}
\providecommand{\BIBentryALTinterwordspacing}{\spaceskip=\fontdimen2\font plus
\BIBentryALTinterwordstretchfactor\fontdimen3\font minus \fontdimen4\font\relax}
\providecommand{\BIBforeignlanguage}[2]{{%
\expandafter\ifx\csname l@#1\endcsname\relax
\typeout{** WARNING: IEEEtran.bst: No hyphenation pattern has been}%
\typeout{** loaded for the language `#1'. Using the pattern for}%
\typeout{** the default language instead.}%
\else
\language=\csname l@#1\endcsname
\fi
#2}}
\providecommand{\BIBdecl}{\relax}
\BIBdecl

\bibitem{household_dataset}
I.~Garcia-Camacho, J.~Borràs, B.~Calli, A.~Norton, and G.~Alenyà, ``Household cloth object set: Fostering benchmarking in deformable object manipulation,'' \emph{IEEE Robotics and Automation Letters}, vol.~7, no.~3, pp. 5866--5873, 2022.

\bibitem{benchmarking_bimanual}
I.~Garcia-Camacho, M.~Lippi, M.~C. Welle, H.~Yin, R.~Antonova, A.~Varava, J.~Borras, C.~Torras, A.~Marino, G.~Alenyà, and D.~Kragic, ``Benchmarking bimanual cloth manipulation,'' \emph{IEEE Robotics and Automation Letters}, vol.~5, no.~2, pp. 1111--1118, 2020.

\bibitem{mo2022foldsformer}
K.~Mo, C.~Xia, X.~Wang, Y.~Deng, X.~Gao, and B.~Liang, ``{Foldsformer: Learning Sequential Multi-Step Cloth Manipulation With Space-Time Attention},'' \emph{IEEE Robotics and Automation Letters}, vol.~8, no.~2, pp. 760--767, 2023.

\bibitem{shridhar2021cliport}
M.~Shridhar, L.~Manuelli, and D.~Fox, ``{CLIPort: What and Where Pathways for Robotic Manipulation},'' in \emph{CoRL}, 2021.

\bibitem{language_deformable_manipulation}
Y.~Deng, K.~Mo, C.~Xia, and X.~Wang, ``{Learning Language-Conditioned Deformable Object Manipulation with Graph Dynamics},'' in \emph{ICRA}, 2024.

\bibitem{speedfolding}
Y.~Avigal, L.~Berscheid, T.~Asfour, T.~Kröger, and K.~Goldberg, ``{SpeedFolding: Learning Efficient Bimanual Folding of Garments},'' in \emph{IROS}, 2022, pp. 1--8.

\bibitem{unet}
O.~Ronneberger, P.~Fischer, and T.~Brox, ``U-net: Convolutional networks for biomedical image segmentation,'' in \emph{Medical Image Computing and Computer-Assisted Intervention}, N.~Navab, J.~Hornegger, W.~M. Wells, and A.~F. Frangi, Eds., 2015, pp. 234--241.

\bibitem{canberk2022clothfunnels}
A.~Canberk, C.~Chi, H.~Ha, B.~Burchfiel, E.~Cousineau, S.~Feng, and S.~Song, ``{Cloth Funnels: Canonicalized-Alignment for Multi-Purpose Garment Manipulation},'' in \emph{ICRA}, 2022.

\bibitem{xue2023unifolding}
H.~Xue, Y.~Li, W.~Xu, H.~Li, D.~Zheng, and C.~Lu, ``Unifolding: Towards sample-efficient, scalable, and generalizable robotic garment folding,'' in \emph{CoRL}, 2023.

\bibitem{clip}
A.~Radford, J.~W. Kim, C.~Hallacy, A.~Ramesh, G.~Goh, S.~Agarwal, G.~Sastry, A.~Askell, P.~Mishkin, J.~Clark, G.~Krueger, and I.~Sutskever, ``Learning {Transferable} {Visual} {Models} {From} {Natural} {Language} {Supervision},'' in \emph{ICML}, 2021.

\bibitem{zeng2020transporter}
A.~Zeng, P.~Florence, J.~Tompson, S.~Welker, J.~Chien, M.~Attarian, T.~Armstrong, I.~Krasin, D.~Duong, V.~Sindhwani, and J.~Lee, ``Transporter networks: Rearranging the visual world for robotic manipulation,'' in \emph{CoRL}, 2020.

\bibitem{transformers}
A.~Vaswani, N.~Shazeer, N.~Parmar, J.~Uszkoreit, L.~Jones, A.~N. Gomez, L.~u. Kaiser, and I.~Polosukhin, ``{Attention is All you Need},'' in \emph{NeurIPS}, vol.~30, 2017.

\bibitem{linLearningVisibleConnectivity2021}
X.~Lin, Y.~Wang, Z.~Huang, and D.~Held, ``Learning {{Visible Connectivity Dynamics}} for {{Cloth Smoothing}},'' in \emph{CoRL}, 2021.

\bibitem{garment_tracking}
H.~Xue, W.~Xu, J.~Zhang, T.~Tang, Y.~Li, W.~Du, R.~Ye, and C.~Lu, ``{GarmentTracking: Category-Level Garment Pose Tracking},'' in \emph{CVPR}, June 2023, pp. 21\,233--21\,242.

\bibitem{vr_dataset_julia}
J.~Borràs, A.~Boix-Granell, S.~Foix, and C.~Torras, ``A virtual reality framework for fast dataset creation applied to cloth manipulation with automatic semantic labelling,'' in \emph{ICRA}, 2023, pp. 11\,605--11\,611.

\bibitem{siglip}
X.~Zhai, B.~Mustafa, A.~Kolesnikov, and L.~Beyer, ``Sigmoid loss for language image pre-training,'' in \emph{ICCV}, 2023, pp. 11\,975--11\,986.

\bibitem{hu2022lora}
\BIBentryALTinterwordspacing
E.~J. Hu, Y.~Shen, P.~Wallis, Z.~Allen-Zhu, Y.~Li, S.~Wang, L.~Wang, and W.~Chen, ``Lo{RA}: Low-rank adaptation of large language models,'' in \emph{International Conference on Learning Representations}, 2022. [Online]. Available: \url{https://openreview.net/forum?id=nZeVKeeFYf9}
\BIBentrySTDinterwordspacing

\bibitem{Open_Clip_2023_CVPR}
M.~Cherti, R.~Beaumont, R.~Wightman, M.~Wortsman, G.~Ilharco, C.~Gordon, C.~Schuhmann, L.~Schmidt, and J.~Jitsev, ``Reproducible scaling laws for contrastive language-image learning,'' in \emph{CVPR}, 2023, pp. 2818--2829.

\bibitem{sun2023evaclipimprovedtrainingtechniques}
Q.~Sun, Y.~Fang, L.~Wu, X.~Wang, and Y.~Cao, ``Eva-clip: Improved training techniques for clip at scale,'' arXiv:2303.15389, 2023.

\bibitem{oquab2024dinov}
M.~Oquab, T.~Darcet, T.~Moutakanni, H.~V. Vo, M.~Szafraniec, V.~Khalidov, P.~Fernandez, D.~HAZIZA, F.~Massa, A.~El-Nouby, M.~Assran, N.~Ballas, W.~Galuba, R.~Howes, P.-Y. Huang, S.-W. Li, I.~Misra, M.~Rabbat, V.~Sharma, G.~Synnaeve, H.~Xu, H.~Jegou, J.~Mairal, P.~Labatut, A.~Joulin, and P.~Bojanowski, ``{DINO}v2: Learning robust visual features without supervision,'' \emph{Transactions on Machine Learning Research}, 2024.

\bibitem{mae_2022}
K.~He, X.~Chen, S.~Xie, Y.~Li, P.~Dollár, and R.~Girshick, ``Masked autoencoders are scalable vision learners,'' in \emph{CVPR}, 2022, pp. 15\,979--15\,988.

\bibitem{tong2024cambrian}
S.~Tong, E.~Brown, P.~Wu, S.~Woo, M.~Middepogu, S.~C. Akula, J.~Yang, S.~Yang, A.~Iyer, X.~Pan, A.~Wang, R.~Fergus, Y.~LeCun, and S.~Xie, ``{Cambrian-1: A Fully Open, Vision-Centric Exploration of Multimodal LLMs},'' arXiv:2406.16860, 2024.

\bibitem{Probing_3D_2024_CVPR}
M.~El~Banani, A.~Raj, K.-K. Maninis, A.~Kar, Y.~Li, M.~Rubinstein, D.~Sun, L.~Guibas, J.~Johnson, and V.~Jampani, ``Probing the 3d awareness of visual foundation models,'' in \emph{CVPR}, 2024, pp. 21\,795--21\,806.

\bibitem{beyer2024paligemma}
L.~Beyer, A.~Steiner, A.~S. Pinto, A.~Kolesnikov, X.~Wang, D.~Salz, M.~Neumann, I.~Alabdulmohsin, M.~Tschannen, E.~Bugliarello, T.~Unterthiner, D.~Keysers, S.~Koppula, F.~Liu, A.~Grycner, A.~Gritsenko, N.~Houlsby, M.~Kumar, K.~Rong, J.~Eisenschlos, R.~Kabra, M.~Bauer, M.~Bošnjak, X.~Chen, M.~Minderer, P.~Voigtlaender, I.~Bica, I.~Balazevic, J.~Puigcerver, P.~Papalampidi, O.~Henaff, X.~Xiong, R.~Soricut, J.~Harmsen, and X.~Zhai, ``{PaliGemma: A versatile 3B VLM for transfer},'' arXiv:2407.07726, 2024.

\bibitem{chen2023pali3visionlanguagemodels}
X.~Chen, X.~Wang, L.~Beyer, A.~Kolesnikov, J.~Wu, P.~Voigtlaender, B.~Mustafa, S.~Goodman, I.~Alabdulmohsin, P.~Padlewski, D.~Salz, X.~Xiong, D.~Vlasic, F.~Pavetic, K.~Rong, T.~Yu, D.~Keysers, X.~Zhai, and R.~Soricut, ``Pali-3 vision language models: Smaller, faster, stronger,'' arXiv:2310.09199, 2023.

\bibitem{SPOC_2024_CVPR}
K.~Ehsani, T.~Gupta, R.~Hendrix, J.~Salvador, L.~Weihs, K.-H. Zeng, K.~P. Singh, Y.~Kim, W.~Han, A.~Herrasti, R.~Krishna, D.~Schwenk, E.~VanderBilt, and A.~Kembhavi, ``Spoc: Imitating shortest paths in simulation enables effective navigation and manipulation in the real world,'' in \emph{CVPR}, 2024, pp. 16\,238--16\,250.

\bibitem{4m}
D.~Mizrahi, R.~Bachmann, O.~F. Kar, T.~Yeo, M.~Gao, A.~Dehghan, and A.~Zamir, ``{4M}: Massively multimodal masked modeling,'' in \emph{Advances in Neural Information Processing Systems}, 2023.

\bibitem{4m21}
R.~Bachmann, O.~F. Kar, D.~Mizrahi, A.~Garjani, M.~Gao, D.~Griffiths, J.~Hu, A.~Dehghan, and A.~Zamir, ``{4M-21}: An any-to-any vision model for tens of tasks and modalities,'' arXiv:2406.09406, 2024.

\bibitem{octo_2023}
{Octo Model Team}, D.~Ghosh, H.~Walke, K.~Pertsch, K.~Black, O.~Mees, S.~Dasari, J.~Hejna, C.~Xu, J.~Luo, T.~Kreiman, Y.~Tan, L.~Y. Chen, P.~Sanketi, Q.~Vuong, T.~Xiao, D.~Sadigh, C.~Finn, and S.~Levine, ``Octo: An open-source generalist robot policy,'' in \emph{Proceedings of Robotics: Science and Systems}, Delft, Netherlands, 2024.

\bibitem{rt2}
A.~Brohan, N.~Brown, J.~Carbajal, Y.~Chebotar, X.~Chen, K.~Choromanski, T.~Ding, D.~Driess, A.~Dubey, C.~Finn, P.~Florence, C.~Fu, M.~G. Arenas, K.~Gopalakrishnan, K.~Han, K.~Hausman, A.~Herzog, J.~Hsu, B.~Ichter, A.~Irpan, N.~Joshi, R.~Julian, D.~Kalashnikov, Y.~Kuang, I.~Leal, L.~Lee, T.-W.~E. Lee, S.~Levine, Y.~Lu, H.~Michalewski, I.~Mordatch, K.~Pertsch, K.~Rao, K.~Reymann, M.~Ryoo, G.~Salazar, P.~Sanketi, P.~Sermanet, J.~Singh, A.~Singh, R.~Soricut, H.~Tran, V.~Vanhoucke, Q.~Vuong, A.~Wahid, S.~Welker, P.~Wohlhart, J.~Wu, F.~Xia, T.~Xiao, P.~Xu, S.~Xu, T.~Yu, and B.~Zitkovich, ``{RT-2: Vision-Language-Action Models Transfer Web Knowledge to Robotic Control},'' arXiv:2307.15818, 2023.

\bibitem{albef}
J.~Li, R.~R. Selvaraju, A.~D. Gotmare, S.~Joty, C.~Xiong, and S.~Hoi, ``Align before fuse: Vision and language representation learning with momentum distillation,'' in \emph{NeurIPS}, 2021.

\bibitem{cloth3d}
H.~Bertiche, M.~Madadi, and S.~Escalera, ``{CLOTH3D: Clothed 3D Humans},'' in \emph{ECCV}, 2020, pp. 344--359.

\bibitem{nocs}
H.~Wang, S.~Sridhar, J.~Huang, J.~Valentin, S.~Song, and L.~J. Guibas, ``{Normalized Object Coordinate Space for Category-Level 6D Object Pose and Size Estimation},'' in \emph{CVPR}, June 2019.

\bibitem{posescript}
G.~Delmas, P.~Weinzaepfel, T.~Lucas, F.~Moreno-Noguer, and G.~Rogez, ``Posescript: 3d human poses from natural language,'' in \emph{ECCV}, 2022, p. 346–362.

\bibitem{adam}
D.~P. Kingma and J.~Ba, ``{Adam: A Method for Stochastic Optimization},'' in \emph{ICLR}, 2015, pp. 1--15.

\bibitem{lin_2021_softgym}
X.~Lin, Y.~Wang, J.~Olkin, and D.~Held, ``{SoftGym: Benchmarking Deep Reinforcement Learning for Deformable Object Manipulation},'' in \emph{CoRL}, 2021.

\bibitem{sam_Kirillov_2023_ICCV}
A.~Kirillov, E.~Mintun, N.~Ravi, H.~Mao, C.~Rolland, L.~Gustafson, T.~Xiao, S.~Whitehead, A.~C. Berg, W.-Y. Lo, P.~Dollar, and R.~Girshick, ``Segment anything,'' in \emph{ICCV}, 2023, pp. 4015--4026.

\bibitem{GPTFabric2024}
V.~Raval, E.~Zhao, H.~Zhang, S.~Nikolaidis, and D.~Seita, ``Gpt-fabric: Folding and smoothing fabric by leveraging pre-trained foundation models,'' \emph{arXiv preprint arXiv:2406.09640}, 2024.

\bibitem{weng2021fabricflownet}
T.~Weng, S.~Bajracharya, Y.~Wang, K.~Agrawal, and D.~Held, ``Fabricflownet: Bimanual cloth manipulation with a flow-based policy,'' in \emph{CoRL}, 2021.

\bibitem{seita_fabrics_2020}
D.~Seita, A.~Ganapathi, R.~Hoque, M.~Hwang, E.~Cen, A.~K. Tanwani, A.~Balakrishna, B.~Thananjeyan, J.~Ichnowski, N.~Jamali, K.~Yamane, S.~Iba, J.~Canny, and K.~Goldberg, ``{Deep Imitation Learning of Sequential Fabric Smoothing From an Algorithmic Supervisor},'' in \emph{IEEE/RSJ International Conference on Intelligent Robots and Systems (IROS)}, 2020.

\bibitem{ha2021flingbot}
H.~Ha and S.~Song, ``Flingbot: The unreasonable effectiveness of dynamic manipulation for cloth unfolding,'' in \emph{Conference on Robotic Learning (CoRL)}, 2021.

\bibitem{perez2018film}
E.~Perez, F.~Strub, H.~de~Vries, V.~Dumoulin, and A.~C. Courville, ``Film: Visual reasoning with a general conditioning layer,'' in \emph{AAAI}, 2018.

\bibitem{t5_2020}
C.~Raffel, N.~Shazeer, A.~Roberts, K.~Lee, S.~Narang, M.~Matena, Y.~Zhou, W.~Li, and P.~J. Liu, ``Exploring the limits of transfer learning with a unified text-to-text transformer,'' \emph{Journal of Machine Learning Research}, vol.~21, no. 140, pp. 1--67, 2020.

\bibitem{rt1}
A.~Brohan, N.~Brown, J.~Carbajal, Y.~Chebotar, J.~Dabis, C.~Finn, K.~Gopalakrishnan, K.~Hausman, A.~Herzog, J.~Hsu, J.~Ibarz, B.~Ichter, A.~Irpan, T.~Jackson, S.~Jesmonth, N.~Joshi, R.~Julian, D.~Kalashnikov, Y.~Kuang, I.~Leal, K.-H. Lee, S.~Levine, Y.~Lu, U.~Malla, D.~Manjunath, I.~Mordatch, O.~Nachum, C.~Parada, J.~Peralta, E.~Perez, K.~Pertsch, J.~Quiambao, K.~Rao, M.~Ryoo, G.~Salazar, P.~Sanketi, K.~Sayed, J.~Singh, S.~Sontakke, A.~Stone, C.~Tan, H.~Tran, V.~Vanhoucke, S.~Vega, Q.~Vuong, F.~Xia, T.~Xiao, P.~Xu, S.~Xu, T.~Yu, and B.~Zitkovich, ``{RT-1: Robotics Transformer for Real-World Control at Scale},'' arXiv:2212.06817, 2022.

\bibitem{huang2023voxposer}
\BIBentryALTinterwordspacing
W.~Huang, C.~Wang, R.~Zhang, Y.~Li, J.~Wu, and L.~Fei-Fei, ``Voxposer: Composable 3d value maps for robotic manipulation with language models,'' in \emph{CoRL}, 2023. [Online]. Available: \url{https://openreview.net/forum?id=9_8LF30mOC}
\BIBentrySTDinterwordspacing

\bibitem{sim2real_gap}
D.~Blanco-Mulero, O.~Barbany, G.~Alcan, A.~Colomé, C.~Torras, and V.~Kyrki, ``{Benchmarking the Sim-to-Real Gap in Cloth Manipulation},'' \emph{IEEE Robotics and Automation Letters}, 2024.

\bibitem{chi2023diffusionpolicy}
C.~Chi, S.~Feng, Y.~Du, Z.~Xu, E.~Cousineau, B.~Burchfiel, and S.~Song, ``Diffusion policy: Visuomotor policy learning via action diffusion,'' in \emph{Proceedings of Robotics: Science and Systems (RSS)}, 2023.

\bibitem{ma2024hierarchical}
X.~Ma, S.~Patidar, I.~Haughton, and S.~James, ``Hierarchical diffusion policy for kinematics-aware multi-task robotic manipulation,'' \emph{CVPR}, 2024.

\bibitem{shen2023distilled}
W.~Shen, G.~Yang, A.~Yu, J.~Wong, L.~P. Kaelbling, and P.~Isola, ``{Distilled Feature Fields Enable Few-Shot Language-Guided Manipulation},'' in \emph{CoRL}, 2023.

\bibitem{zhou2023clothesnet}
B.~Zhou, H.~Zhou, T.~Liang, Q.~Yu, S.~Zhao, Y.~Zeng, J.~Lv, S.~Luo, Q.~Wang, X.~Yu, H.~Chen, C.~Lu, and L.~Shao, ``{ClothesNet: An Information-Rich 3D Garment Model Repository with Simulated Clothes Environment},'' in \emph{ICCV}, 2023.

\bibitem{blenderproc}
M.~Denninger, D.~Winkelbauer, M.~Sundermeyer, W.~Boerdijk, M.~Knauer, K.~H. Strobl, M.~Humt, and R.~Triebel, ``{BlenderProc2: A Procedural Pipeline for Photorealistic Rendering},'' \emph{Journal of Open Source Software}, vol.~8, no.~82, p. 4901, 2023.

\bibitem{dosovitskiy2021an}
\BIBentryALTinterwordspacing
A.~Dosovitskiy, L.~Beyer, A.~Kolesnikov, D.~Weissenborn, X.~Zhai, T.~Unterthiner, M.~Dehghani, M.~Minderer, G.~Heigold, S.~Gelly, J.~Uszkoreit, and N.~Houlsby, ``An image is worth 16x16 words: Transformers for image recognition at scale,'' in \emph{ICLR}, 2021. [Online]. Available: \url{https://openreview.net/forum?id=YicbFdNTTy}
\BIBentrySTDinterwordspacing

\bibitem{focal_loss}
N.~Carion, F.~Massa, G.~Synnaeve, N.~Usunier, A.~Kirillov, and S.~Zagoruyko, ``End-to-end object detection with transformers,'' in \emph{ECCV}, 2020, p. 213–229.

\bibitem{coumans_2021_bullet}
E.~Coumans and Y.~Bai, ``{PyBullet, a Python module for physics simulation for games, robotics and machine learning},'' \url{http://pybullet.org}, 2016--2021.

\bibitem{todorov_2012_mujoco}
E.~Todorov, T.~Erez, and Y.~Tassa, ``{MuJoCo: A physics engine for model-based control},'' in \emph{IEEE/RSJ International Conference on Intelligent Robots and Systems}, 2012.

\end{thebibliography}

\appendices

\onecolumn

\begin{center}
    {\LARGE\bf BiFold: Bimanual Cloth Folding with Language Guidance} \\[1em]
    {\centering\LARGE\bf Supplementary Material}
\end{center}

\tableofcontents

\clearpage

\section{Dataset generation}

An instance of the VR-Folding dataset \cite{garment_tracking} has the following structure:
\dirtree{%
.1 00001\_Skirt\_000000\_000000. 
.2 grip\_vertex\_id. 
.3 left\_grip\_vertex\_id (1,) int32. 
.3 right\_grip\_vertex\_id (1,) int32. 
.2 marching\_cube\_mesh. 
.3 is\_vertex\_on\_surface (3889,) bool. 
.3 marching\_cube\_faces (7770, 3) int32. 
.3 marching\_cube\_verts (3889, 3) float32. 
.2 mesh. 
.3 cloth\_faces\_tri (5168, 3) int32. 
.3 cloth\_nocs\_verts (2695, 3) float32. 
.3 cloth\_verts (2695, 3) float32. 
.2 point\_cloud. 
.3 cls (30000,) uint8. 
.3 nocs (30000, 3) float16. 
.3 point (30000, 3) float16. 
.3 rgb (30000, 3) uint8. 
.3 sizes (4,) int64. 
}
which does not include annotations for the language-conditioned manipulation task we tackle.

This dataset does not contain action-level segmentation of the actions or natural language annotations. For each instance, if one of the hands of the demonstrator is grasping a point, the dataset provides the indices of the closest vertices of the simulation mesh. Therefore, there is no information on how the hand approaches the garment. Additionally, the only perceptual input is a point cloud of fixed size. No RGB or depth images are available.

The following sections describe how we process the dataset to obtain the required model inputs and ground truth labels. The process we propose does not require human intervention, contrary to the one used by Deng \etal for collecting the unimanual dataset \cite{language_deformable_manipulation}, and therefore can be easily scaled.

\subsection{Rendering}

\begin{figure}[ht]
    \centering
    \begin{subfigure}{0.24\linewidth}
        \centering
        \includegraphics[width=\linewidth]{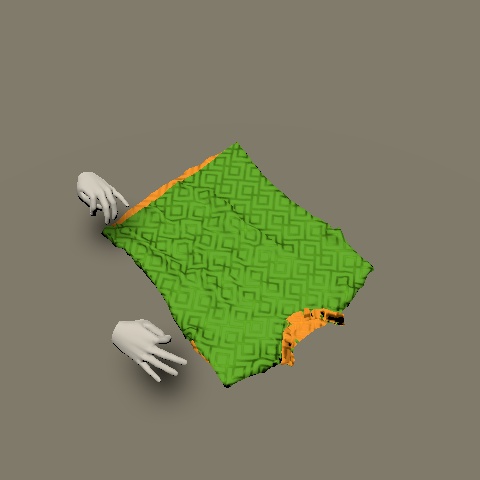}
        \caption{$t=45$}
    \end{subfigure}
    \begin{subfigure}{0.24\linewidth}
        \centering
        \includegraphics[width=\linewidth]{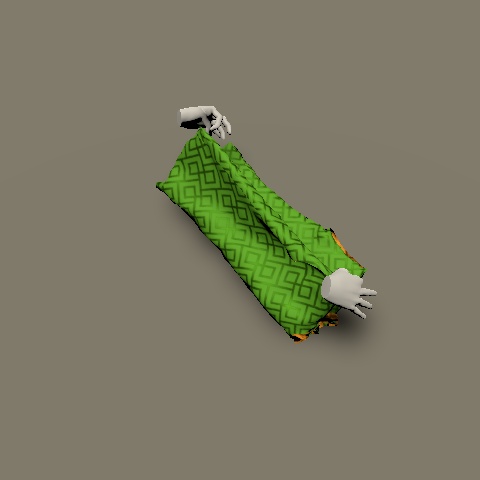}
        \caption{$t=105$}
    \end{subfigure}
    \begin{subfigure}{0.24\linewidth}
        \centering
        \includegraphics[width=\linewidth]{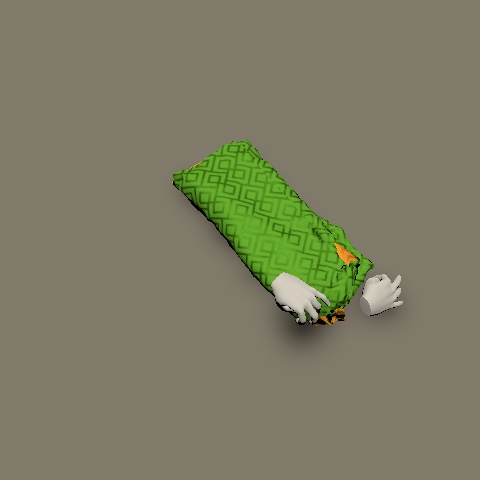}
        \caption{$t=155$}
    \end{subfigure}
    \begin{subfigure}{0.24\linewidth}
        \centering
        \includegraphics[width=\linewidth]{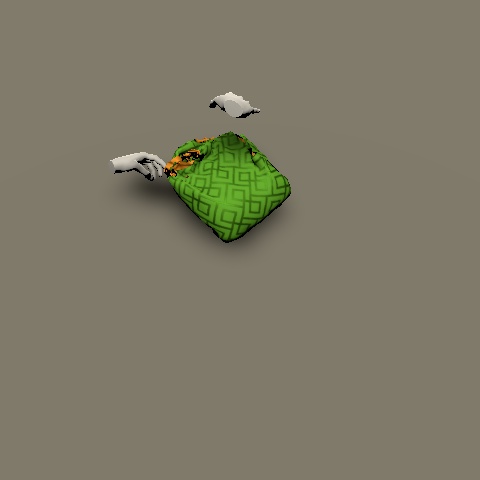}
        \caption{$t=210$}
    \end{subfigure}
    \caption{\textbf{VR-Folding sequence: }Sample sequence from the VR-Folding dataset in which hands are included using the sensed positions from the VR gloves. In this image we can see the tiling green pattern and constant orange interior.}
    \label{fig:vr_sample}
\end{figure}

The VR-Folding dataset does not contain RGB-D inputs, and the provided simulation meshes are untextured. Moreover, the garments used in the simulator have a constant color for the interior and a repetitive texture with a differentiated color that yields colored point clouds, as shown in \cref{fig:vr_sample}.

This design choice can lead to inputs that can be more easily registered. For example, one can identify the interior and exterior, self-intersections, and the scale of the cloth. However, this texturing could limit the generalization capabilities of our RGB-based models for the same reason, causing the learning process to focus only on such patterns and differentiating the few colors seen during training.

While VR-Folding obtains the simulation meshes by manipulating assets extracted from CLOTH3D \cite{cloth3d}, they are re-meshed to obtain triangular faces from the original quadratic faces. For this reason, applying face textures cannot be done straightforwardly and requires a first step of texture baking. After assigning each vertex and triangular face to the original CLOTH3D assets, we can transfer the cloth texture to the simulation meshes. When assigning a material to the mesh, we use a material definition from ClothesNet \cite{zhou2023clothesnet}, which effectively achieves a realistic cloth effect:\\

\noindent\texttt{Ns 28.763235\\
Ka 1.000000 1.000000 1.000000\\
Ks 0.075000 0.075000 0.075000\\
Ke 0.000000 0.000000 0.000000\\
Ni 1.450000\\
d 1.000000\\
illum 2\\
}

\begin{figure}[ht]
    \centering
    \includegraphics[width=.8\textwidth]{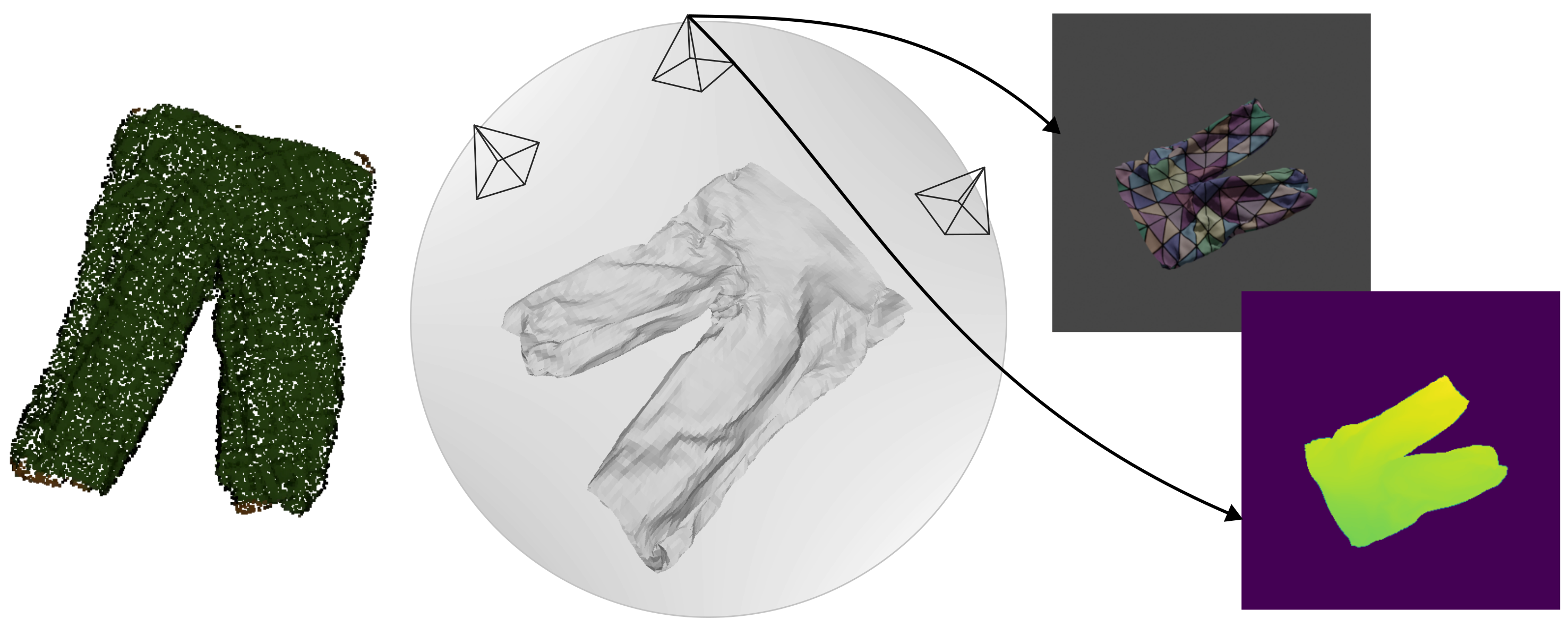}
    \caption{\textbf{Re-rendering step: }The only visual input in the original dataset is given as a colored point cloud with uniform colors and patterns (left). We take the simulation mesh, and randomly choose camera position (center). Finally, we apply a texture to the mesh and render RGB-D images (right).}
    \label{fig:render_draw}
\end{figure}

We render RGB-D images using BlenderProc2 \cite{blenderproc} with cameras pointing at the object whose position we randomly sample from the volume delimited by two spherical caps of different radii with centers on the manipulated object. Concretely, we define the volume using elevations $[45^{\circ}, 90^{\circ}]$ and radii $[1.8, 2.2]$. We use the same camera position for all the steps of the same sequence. Finally, we render RGB-D images with a resolution of 384x384 pixels. We include an example of the original input and our newly introduced samples in \cref{fig:render_draw}. The decision to use random cameras rather than fixed cameras, as in Deng \etal \cite{language_deformable_manipulation}, is one of the reasons why our dataset is challenging. When using random camera positions, the input clothes have different sizes, and their shape is more affected by perspective as the camera deviates from the zenithal position. We ensure that all the pick and place positions fall inside the image and resample a new camera to generate the complete sequence again otherwise.

\subsection{Language annotations}

\begin{figure}[ht]
    \centering
    \begin{subfigure}[t]{0.32\textwidth}
        \centering
        \includegraphics[width=\textwidth]{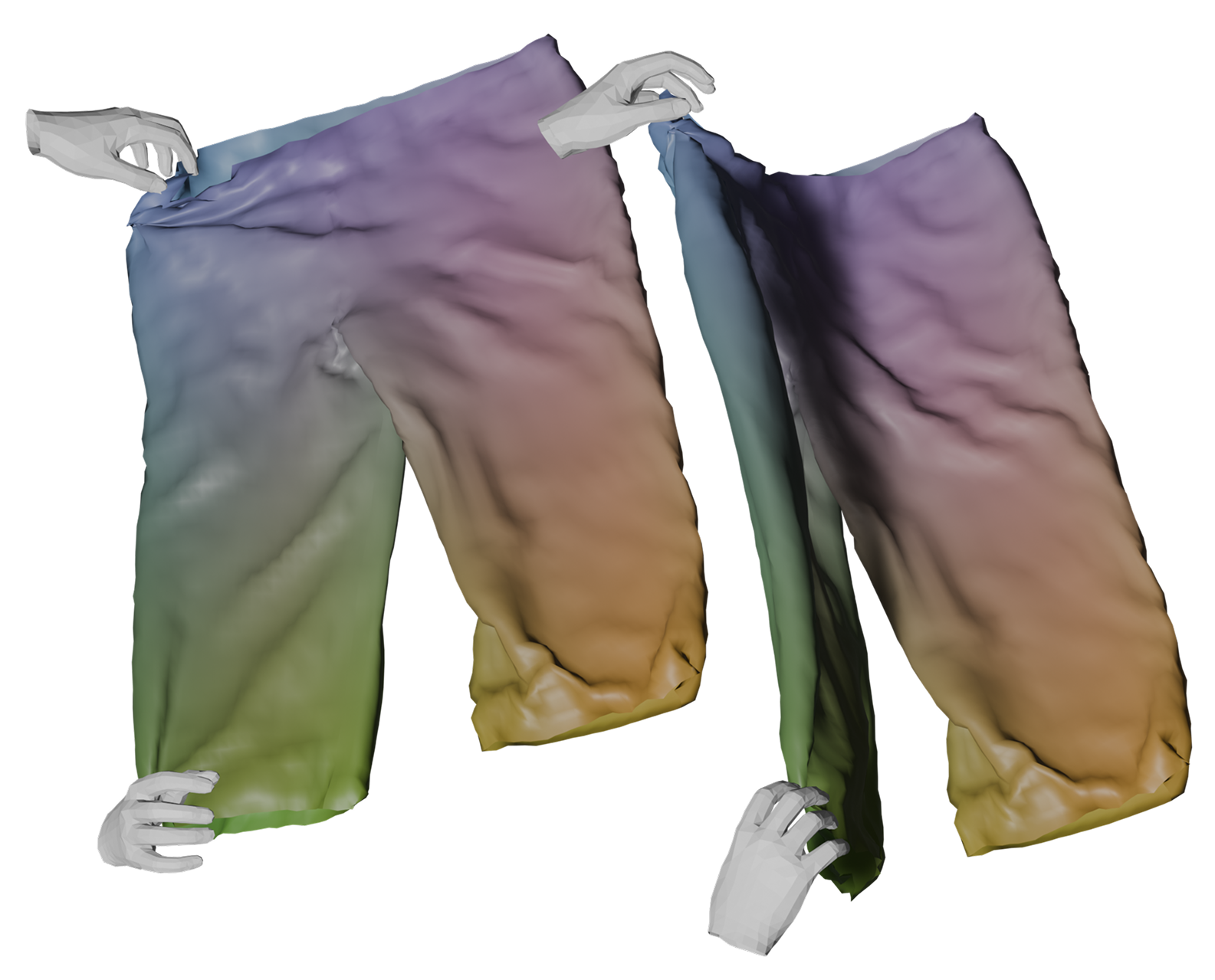}
        \caption{\texttt{Bring the right side of the trousers towards the left side and fold them in half.}}
        \label{fig:nocs_track1}
    \end{subfigure}
    \hfill
    \begin{subfigure}[t]{0.32\textwidth}
        \centering
        \includegraphics[width=\textwidth]{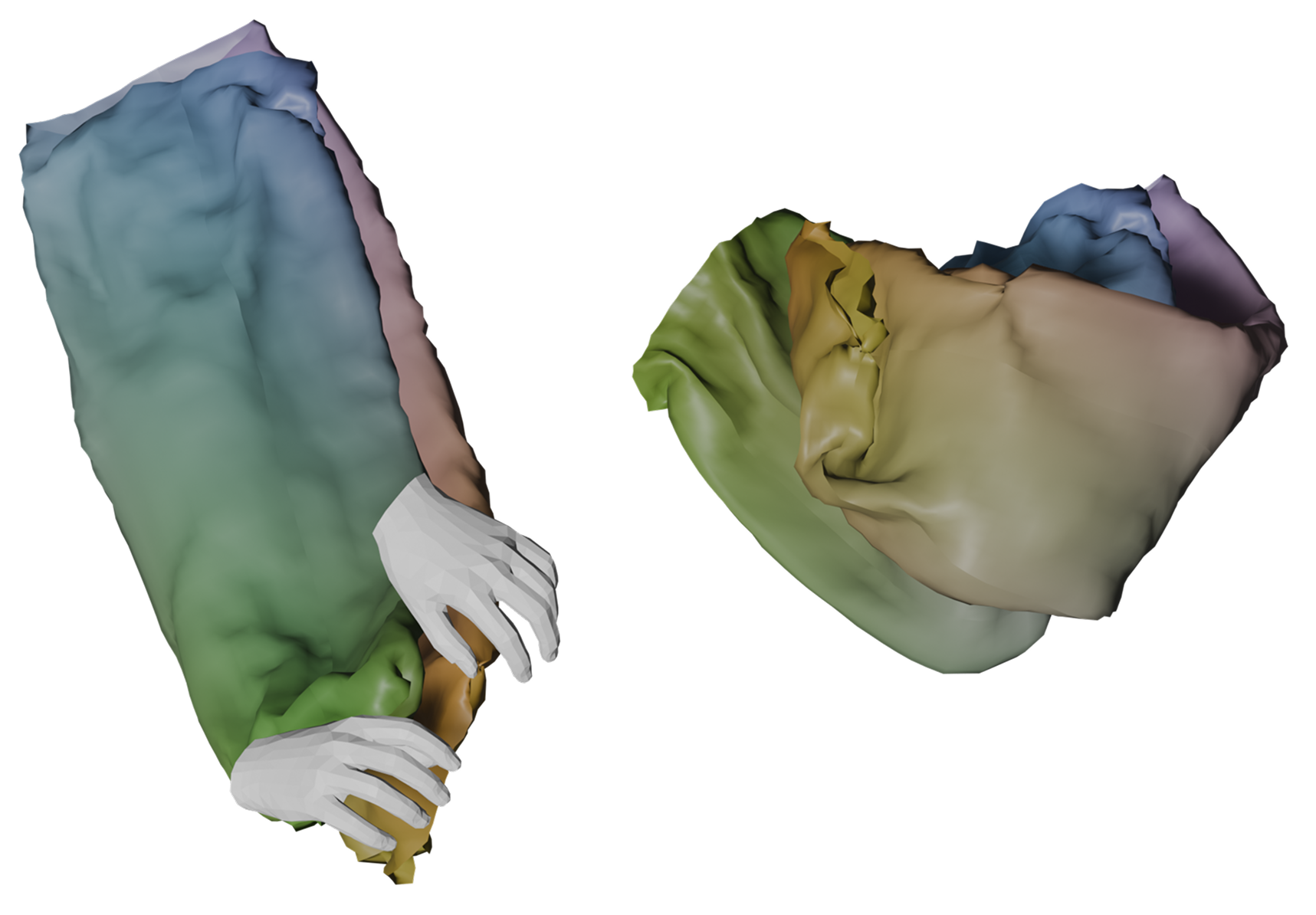}
        \caption{\texttt{Bend the trousers in half, from bottom to top.}}
        \label{fig:nocs_track2}
    \end{subfigure}
    \hfill
    \begin{subfigure}[t]{0.32\textwidth}
        \centering
        \includegraphics[width=\textwidth]{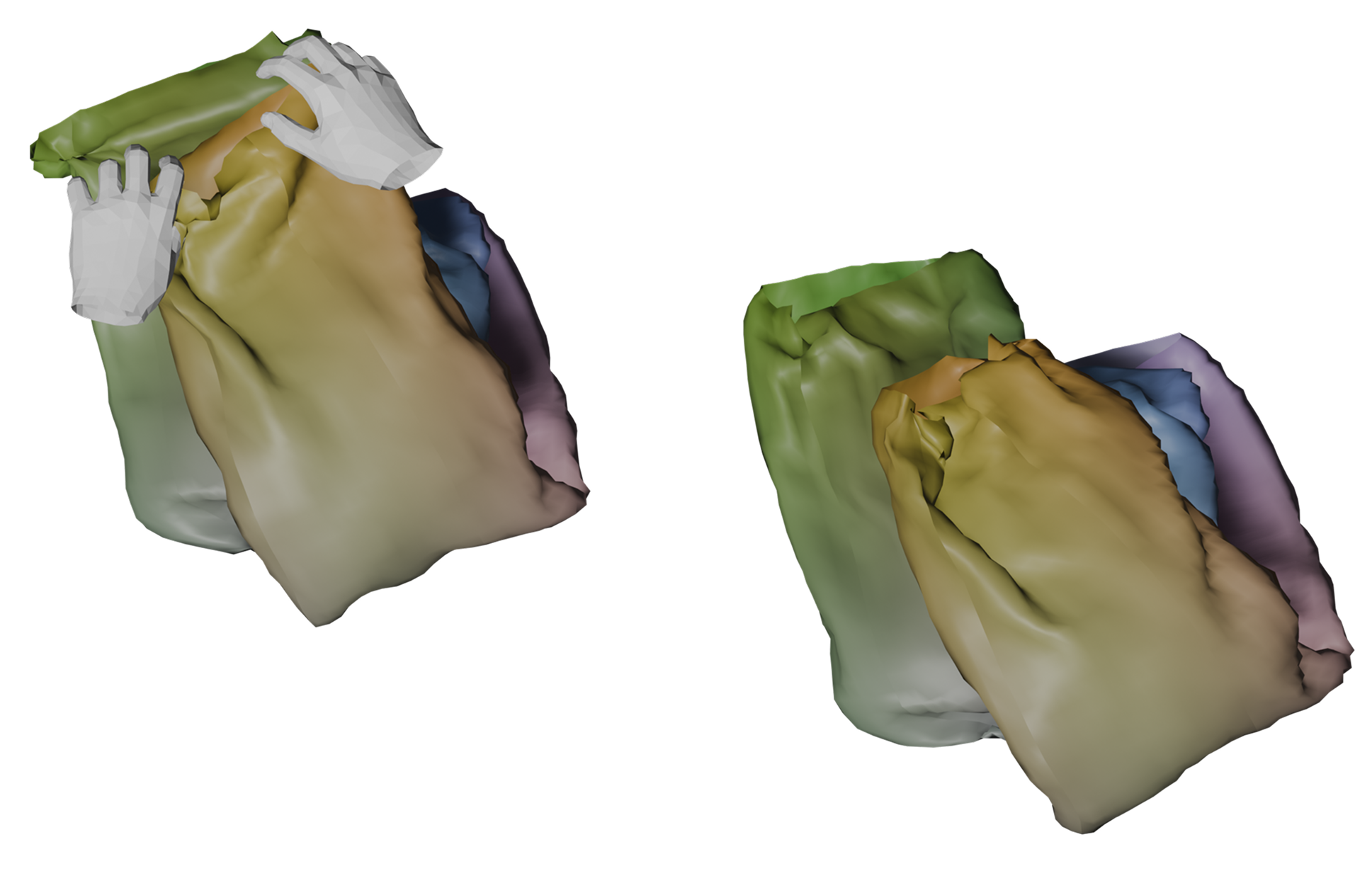}
        \caption{\texttt{Straighten out the bottom part of the trousers.}}
        \label{fig:nocs_track3}
    \end{subfigure}
    \caption{\textbf{Continuous \gls*{nocs} tracking across pick--and--place subactions.}  
Each panel shows a single pick-and-place step rendered in two columns: the left image corresponds to the pre-action state, and the right to the post-action state. We overlay the user’s VR-tracked hand positions along with the corresponding text instruction. Unlike the vertex positions, the \gls*{nocs} coordinates remain consistent throughout the manipulation, enabling stable vertex-to-semantic-region assignments as long as mesh topology is preserved.}

    \label{fig:nocs_tracking}
\end{figure}

\begin{figure}[ht]
    \centering
    \includegraphics[width=.7\textwidth]{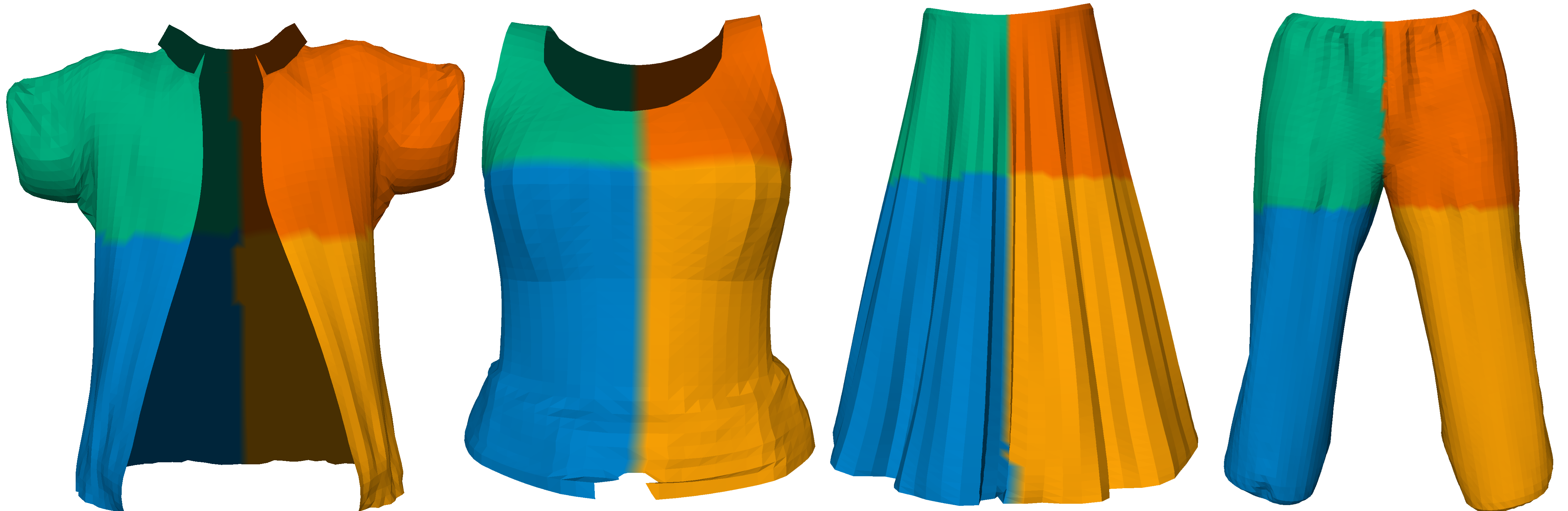}
    \caption{\textbf{Semantic pick and place positions: }We obtain the semantic location of the grip by mapping the picked vertices on the \gls*{nocs} and thresholding its coordinates. In this figure, we show an example of each category colored by thresholding the left-right and top-bottom directions. We do not threshold the front-rear direction as it is not relevant for the considered actions.}
    \label{fig:nocs}
\end{figure}

To enable consistent labeling across frames, we continuously track the \gls*{nocs} coordinates of the garment mesh throughout the manipulation sequence. This ensures that the vertex assignments remain stable despite large deformations, which is crucial for accurate semantic labeling. An illustration of this tracking process is shown in \cref{fig:nocs_tracking}, where a pair of pants is manipulated and the associated \gls*{nocs} values are maintained across frames. Once the \gls*{nocs} mapping is established, we apply axis-aligned thresholding to discretize the coordinate space and assign semantic labels to different garment regions, as shown in \cref{fig:nocs}. This approach allows us to distinguish key parts such as sleeves, legs, and waistbands across diverse garment types using a unified representation. Notably, we do not threshold the front–rear direction, as it is not relevant for the folding actions considered in this work.

When annotating bimanual actions using \gls*{nocs} coordinates, it is usual that the left and right pickers use different semantic locations, \eg, the left picker grabs the top right part, and the right picker grabs the bottom right part. In this case, we can infer that the common objective is to hold the right part of the garment, but one cannot trivially resolve many other situations. To do that, we designed a heuristics detailed in \cref{alg:semantic_position_heuristics}, which outputs a common semantic location considering the positions of the left and right pickers given the context of the action.

\begin{algorithm}[ht]
	\caption{\textsc{Determining semantic pick and place locations}}
	\label{alg:semantic_position_heuristics}
	\begin{algorithmic}[1]
		\STATE \textbf{Inputs: }Semantic location for left and right pickers along the top-bottom and left-right planes $l_v, l_h, r_v, r_h$, and action type \texttt{"pick"} or \texttt{"place"}. If action is \texttt{"place"}, semantic location of pick action $s_{\text{pick}}$
		\STATE \textbf{Output: }Semantic location and in some cases a sleeve flag
		\STATE $v \leftarrow l_v$ if $l_v=r_v$ else \textsc{null} \algorithmiccomment{Same position in top-bottom plane}
            \STATE $h \leftarrow l_h$ if $l_h=r_h$ else \textsc{null} \algorithmiccomment{Same position in left-right plane}
            \IF{$h$ is not \textsc{null}}
                \IF{$v$ is not \textsc{null}}
                    \IF{Action type is \texttt{"place"}}
                        \STATE $\rhd$ Try to avoid same pick and place position
                        \IF{$s_{\text{pick}}=h$}
                            \RETURN $v$
                        \ELSIF{$s_{\text{pick}}=v$}
                            \RETURN $h$
                        \ELSIF{$s_{\text{pick}}$ is the opposite as $h$}
                            \RETURN $h$
                        \ELSIF{$s_{\text{pick}}$ is the opposite as $v$}
                            \RETURN $v$
                        \ELSE
                            \RETURN $v$ + \texttt{" "} + $h$ \algorithmiccomment{Use both semantic locations together}
                        \ENDIF
                    \ELSE
                        \IF{The garment is a T-shirt and $v$ is top}
                            \RETURN $h$ and sleeve flag. Place action in this case is irrelevant.
                        \ELSE
                            \RETURN $v$ + \texttt{" "} + $h$ \algorithmiccomment{Use both semantic locations together}
                        \ENDIF
                    \ENDIF
                \ELSE
                    \RETURN $h$
                \ENDIF
            \ELSE
                \IF{$v$ is not \textsc{null}}
                    \RETURN $h$
                \ELSE
                    \IF{Action type is \texttt{"place"}}
                        \STATE $\rhd$ Place vertices may be wrong (\eg, sleeve over bottom)
                        \RETURN Opposite location of $s_{\text{Pick}}$
                    \ENDIF
                \ELSE
                    \RETURN \textsc{null} \algorithmiccomment{Raise error}
                \ENDIF
            \ENDIF
	\end{algorithmic}
\end{algorithm}

Once the semantic location of the pick and place positions are known, we assign a language instruction to the folding sub-action. We make use of template prompts and include the complete list in \cref{tab:language_instructions_folds,tab:language_instructions_refine,tab:language_instructions_sleeves}. We can distinguish three different kinds of actions: sleeve manipulations, refinements, and generic folds. For all of them, the prompts follow a template where the placeholders surrounded by brackets, \ie, \{which\} and \{garment\}, are replaced by the pick and place semantic locations and the garment type, respectively. If the heuristics returns a sleeve flag, we use a prompt from the set of language instructions for sleeves with the semantic pick position. If the pick and place position is the same, we assume that the volunteer is performing a refinement and uses one of the prompts from a predefined list of language templates for small actions. Otherwise, we use a generic fold prompt from a set of generic instructions.
 
\begin{table}[ht]
    \centering
    \caption{\textbf{Language instructions for sleeve manipulations.}}
    \resizebox{%
      \ifdim\width>\linewidth
        \linewidth
      \else
        \width
      \fi
    }{!}{
    \begin{tabular}{c}
        \toprule
        Fold the \{which\} sleeve towards the inside. \\
        Inwardly fold the \{which\} sleeve. \\
        Fold the \{which\} sleeve towards the body. \\
        Bend the \{which\} sleeve towards the inside. \\
        Fold the \{which\} sleeve to the center. \\
        Fold the \{which\} sleeve towards the middle. \\
        Bring the \{which\} sleeve to the center. \\
        Fold the \{which\} sleeve inward to the halfway point. \\
        Tuck the \{which\} sleeve towards the center. \\
        Meet the \{which\} sleeve at the center. \\
        Fold the \{which\} sleeve to the midpoint. \\
        Center the \{which\} sleeve. \\
        Align the \{which\} sleeve to the center. \\
        Fold the \{which\} sleeve to the axis. \\
        Bring the \{which\} sleeve to the median. \\
        Fold the \{which\} sleeve to the central point. \\
        Fold the \{which\} sleeve towards the midpoint of the shirt. \\
        Bring the \{which\} sleeve to the center seam. \\
        Fold the \{which\} sleeve to the centerline of the shirt. \\
        Fold the \{which\} sleeve to the centerline of the shirt. \\
        \bottomrule
    \end{tabular}
    }
    \label{tab:language_instructions_sleeves}
\end{table}

\begin{table}[ht]
    \centering
    \caption{\textbf{Language instructions for small refinements.}}
    \resizebox{%
      \ifdim\width>\linewidth
        \linewidth
      \else
        \width
      \fi
    }{!}{
    \begin{tabular}{c}
        \toprule
        Fold the \{which\} part of the \{garment\} neatly. \\
        Align the \{which\} part of the \{garment\} properly. \\
        Arrange the \{which\} part of the \{garment\} neatly. \\
        Straighten out the \{which\} part of the \{garment\}. \\
        Place the \{which\} part of the \{garment\} in the correct position. \\
        Ensure the \{which\} part of the \{garment\} is well-positioned. \\
        \bottomrule
    \end{tabular}
    }
    \label{tab:language_instructions_refine}
\end{table}

\begin{table}[ht]
    \centering
    \caption{\textbf{Language instructions for folds.}}
    \resizebox{%
      \ifdim\width>\linewidth
        \linewidth
      \else
        \width
      \fi
    }{!}{
    \begin{tabular}{c}
        \toprule
        Fold the \{garment\} in half, \{which1\} to \{which2\}. \\
        Fold the \{garment\} from the \{which1\} side towards the \{which2\} side. \\
        Fold the \{garment\} in half, starting from the \{which1\} and ending at the \{which2\}. \\
        Fold the \{garment\}, \{which1\} side over \{which2\} side. \\
        Bend the \{garment\} in half, from \{which1\} to \{which2\}. \\
        Fold the \{garment\}, making sure the \{which1\} side touches the \{which2\} side. \\
        Fold the \{garment\}, bringing the \{which1\} side to meet the \{which2\} side. \\
        Crease the \{garment\} down the middle, from \{which1\} to \{which2\}. \\
        Fold the \{garment\} in half horizontally, \{which1\} to \{which2\}. \\
        Make a fold in the \{garment\}, starting from the \{which1\} and ending at the \{which2\}. \\
        Fold the \{garment\} in half, aligning the \{which1\} and \{which2\} sides. \\
        Fold the \{garment\}, ensuring the \{which1\} side meets the \{which2\} side. \\
        Fold the \{garment\}, orientating from the \{which1\} towards the \{which2\}. \\
        Fold the \{garment\} cleanly, from the \{which1\} side to the \{which2\} side. \\
        Fold the \{garment\} in half, with the \{which1\} side overlapping the \{which2\}. \\
        Create a fold in the \{garment\}, going from \{which1\} to \{which2\}. \\
        Bring the \{which1\} side of the \{garment\} towards the \{which2\} side and fold them in half. \\
        Fold the waistband of the \{garment\} in half, from \{which1\} to \{which2\}. \\
        Fold the \{garment\} neatly, from the \{which1\} side to the \{which2\} side. \\
        Fold the \{garment\}, making a crease from the \{which1\} to the \{which2\}. \\
        \bottomrule
    \end{tabular}
    }
    \label{tab:language_instructions_folds}
\end{table}

\clearpage

\subsection{Filtering out divergent sequences}

\begin{figure}[ht]
    \centering
    \begin{subfigure}{0.24\linewidth}
        \centering
        \includegraphics[width=\linewidth]{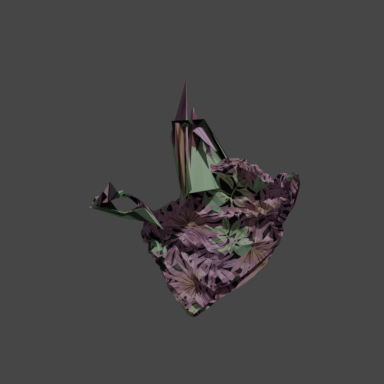}
        \caption{$t=50$}
    \end{subfigure}
    \begin{subfigure}{0.24\linewidth}
        \centering
        \includegraphics[width=\linewidth]{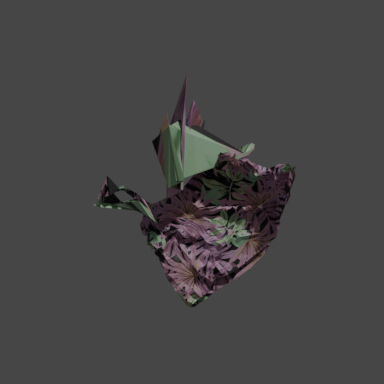}
        \caption{$t=55$}
    \end{subfigure}
    \begin{subfigure}{0.24\linewidth}
        \centering
        \includegraphics[width=\linewidth]{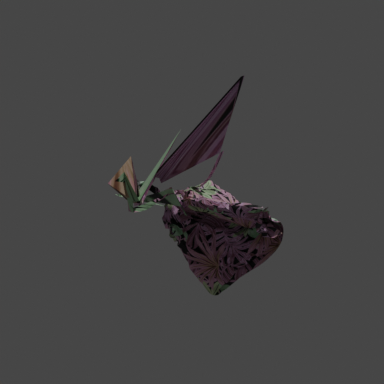}
        \caption{$t=75$}
    \end{subfigure}
    \begin{subfigure}{0.24\linewidth}
        \centering
        \includegraphics[width=\linewidth]{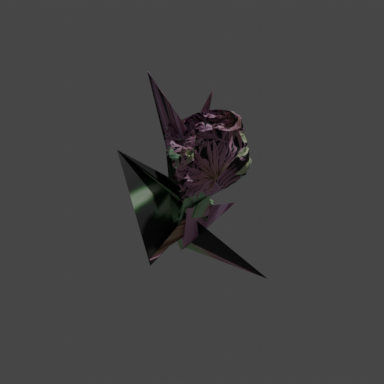}
        \caption{$t=195$}
    \end{subfigure}
    \caption{\textbf{Divergent sequence: }In this sequence, we can see that the manipulated top has become unstable in the cloth simulator, yielding unrealistic shapes. The top uses the CLOTH3D \cite{cloth3d} mesh with identifier 00156 and corresponds to the sequence 00156\_Top\_000013\_$t$ in the VR-Folding dataset \cite{garment_tracking}, where the dataset uses time notation $t$ in steps of 5, \ie $t=50$ and $t=55$ are consecutive frames.}
    \label{fig:divergent_sequence}
\end{figure}

Simulating clothes is a very challenging task for which some simulators exhibit a large gap with reality \cite{sim2real_gap}. Additionally, the constrained optimization routines used by some simulators might lead to unstable solutions where the predicted cloth vertices diverge. We found several sequences of the VR-Folding dataset \cite{garment_tracking} where the clothes underwent this phenomenon and present one of them in \cref{fig:divergent_sequence}. \cref{fig:divergent_sequence} presents an example where the simulation meshes have diverged. Wwe can observe that when the simulator becomes unstable, some of the vertex positions are excessively far, creating unusually long edges. To remove this and other occurrences of this phenomenon, we compute the set of edge lengths
 \begin{align}
     \mathcal{E}_{\text{lengths}}:=\{\norm{\vv_i = \vv_j}: \left( \vv_i,\vv_j\right)\in \mathcal{E}\}\,,
 \end{align}
where $\mathcal{E}$ is the set of vertices of the mesh. Then, we compute the quantity
 \begin{align}
     \frac{\text{max}\left[\mathcal{E}_{\text{lengths}}\right] - \mathbb{E}\left[{\mathcal{E}_{\text{lengths}}}\right]}{\sqrt{\text{VAR}\left[\mathcal{E}_{\text{lengths}}\right]}}\,,
 \end{align}
which computes how much the maximum edge length deviates from the mean normalized by the standard deviation. Finally, we filter out all meshes for which the ratio of the quantity in the equation above between the mesh of interest and the \gls*{nocs} mesh exceeds 3.5. Note that \gls*{nocs} has a different scale but the ratio of standard deviations removes this effect.

\subsection{Dataset statistics}

The VR-Folding dataset contains almost 4000 bimanual folding demonstrations from humans. The demonstrations are performed with a large variety of meshes, being almost all of them distinct. From these sequences, we are able to segment around 7000 actions and obtain aligned text instructions. By means of our proposed pipeline, we are able to obtain a diverse set of language instructions, totaling more than 1000 unique prompts, as seen in \cref{tab:dataset_stats_per_garment}.

\begin{table}[ht]
    \centering
    \begin{tabular}{ccccc}
        \toprule
        \textbf{Garment} & \textbf{\# demos} & \textbf{\# meshes} & \textbf{\# actions} &\textbf{ \# instructions} \\
        \midrule
        Skirt	&464	&462	&857	&196 \\
        Top&	893&	890	&891	&262 \\
        Trousers&	1541&	1539	&2770&	353 \\
        T-shirt&	993&	990&	2444&	254 \\
        \bottomrule
    \end{tabular}
    \caption{Caption}
    \label{tab:dataset_stats_per_garment}
\end{table}

\begin{figure}[ht]
    \centering
    \begin{subfigure}{.48\linewidth}
        \centering
        \includegraphics[width=\linewidth]{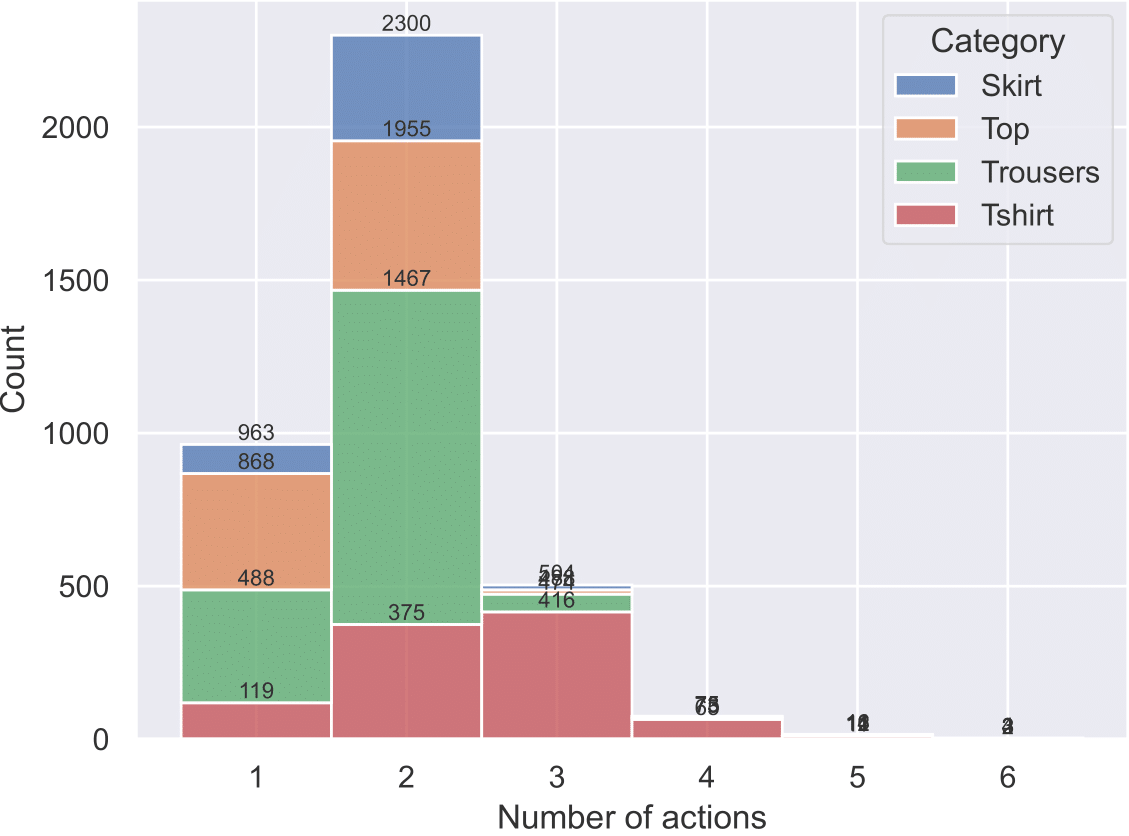}
    \end{subfigure}
    \begin{subfigure}{.48\linewidth}
        \centering
        \includegraphics[width=\linewidth]{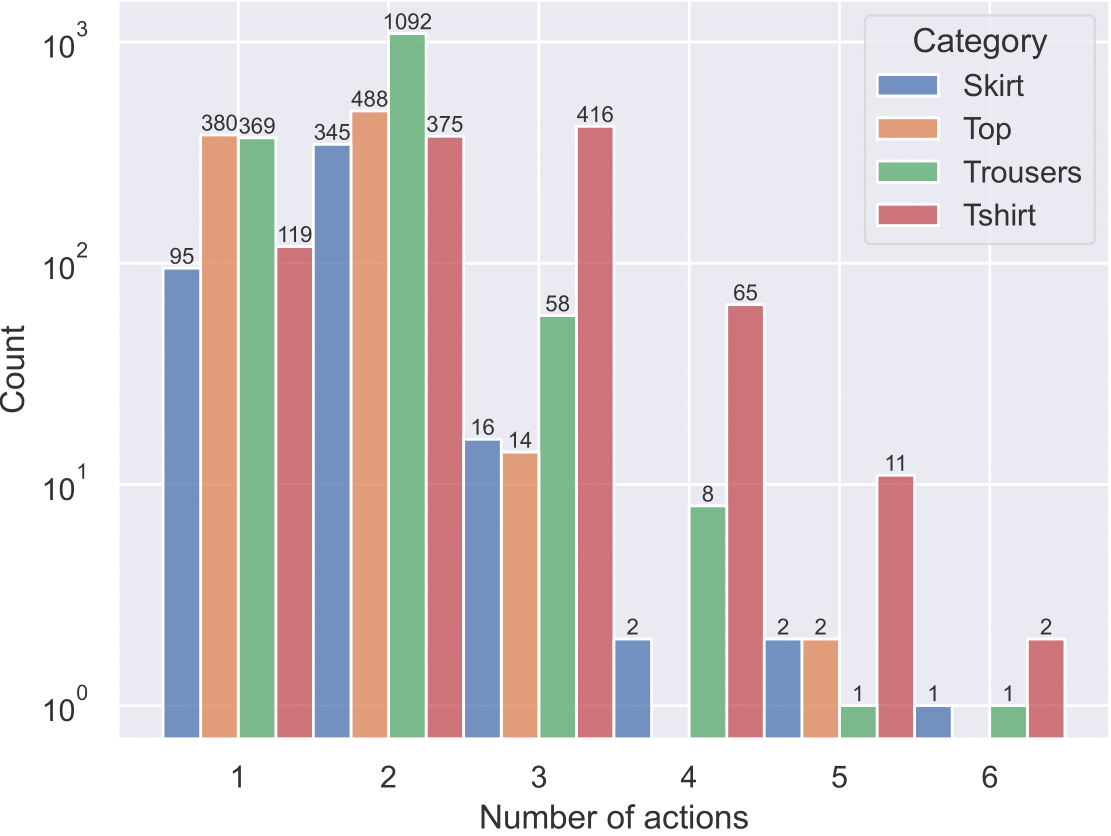}
    \end{subfigure}
    \caption{\textbf{Sequence length per category: }Here we show a histogram of the number of actions of each sequence grouped by clothing category in linear scale and stacked (left) and logarithmic scale and separated (right) for the BiFold dataset. Note that some sequences can have only one action as the sequence can be truncated by some filtering step and the valid folding sub-action can still be useful for training.}
    \label{fig:actions_per_category}
\end{figure}

In \cref{fig:actions_per_category}, we show a histogram of the number of actions of each sequence grouped by clothing category in linear scale and stacked (left) and logarithmic scale and separated (right) for the BiFold dataset. Note that some sequences can have only one action as the sequence can be truncated by some filtering step and the valid folding sub-action can still be useful for training. We can see that there are sequences with up to six parsed actions, showing the clear need for our novel pipeline for action parsing and annotation as the demonstrators do not follow the predefined instructions.

\begin{figure}[ht]
    \begin{subfigure}{.5\linewidth}
        \centering
        \includegraphics[width=\linewidth]{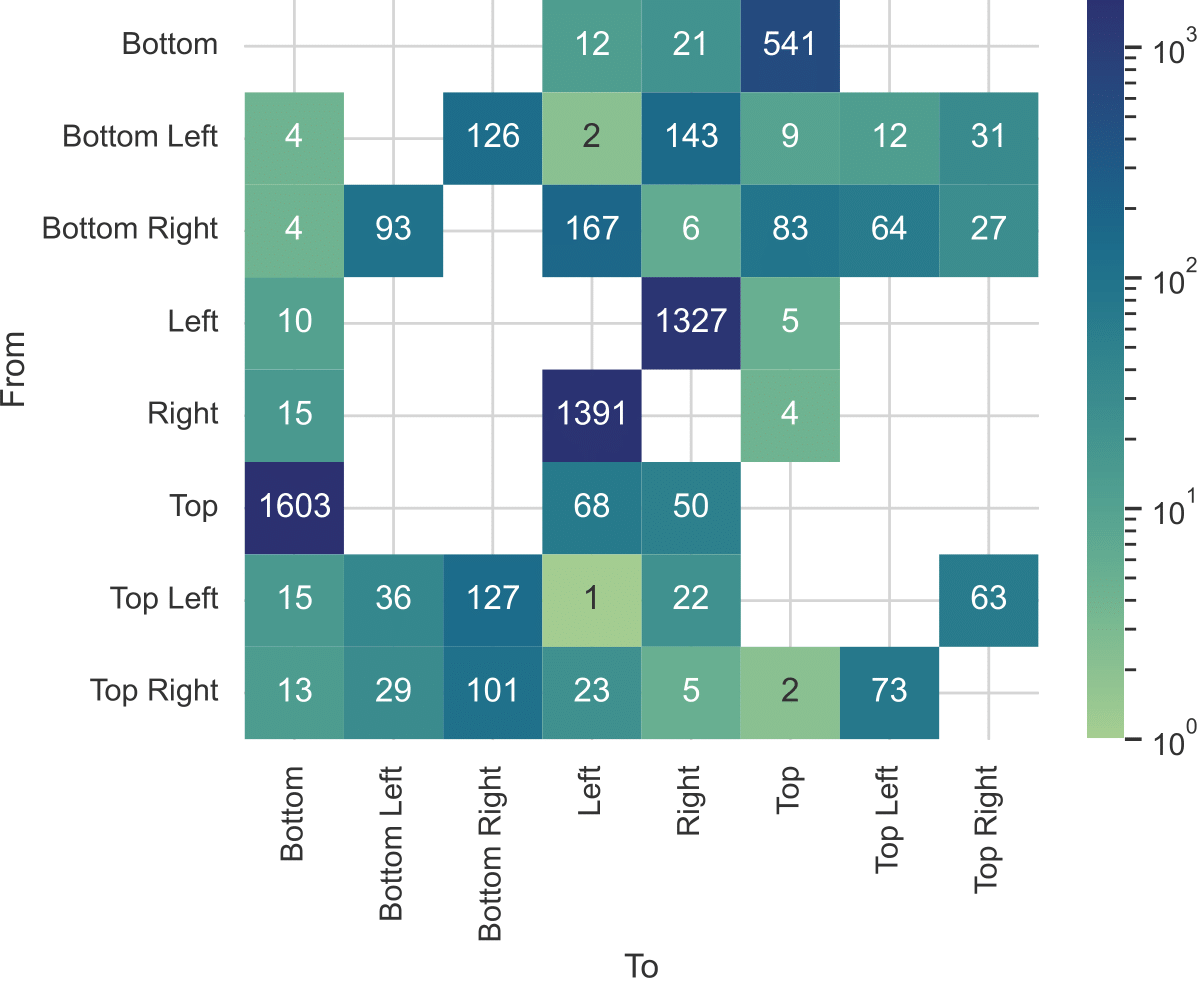}
        \caption{Distribution of the from$\to$to folds.}
        \label{fig:which_folds}
    \end{subfigure}
    \begin{minipage}[t]{.3\textwidth}
        \vspace{-8cm}
        \begin{subfigure}{\linewidth}
            \centering
            \includegraphics[width=.7\linewidth]{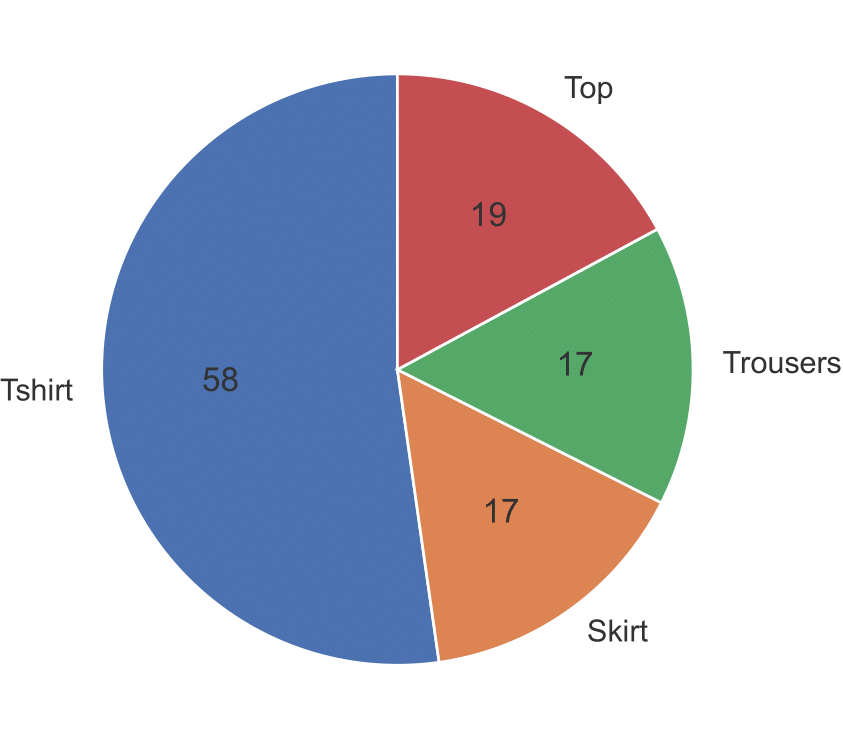}
            \caption{Use of refinement actions}
            \label{fig:refinement_actions}
        \end{subfigure}
        \begin{subfigure}{\linewidth}
            \centering
            \includegraphics[width=.7\linewidth]{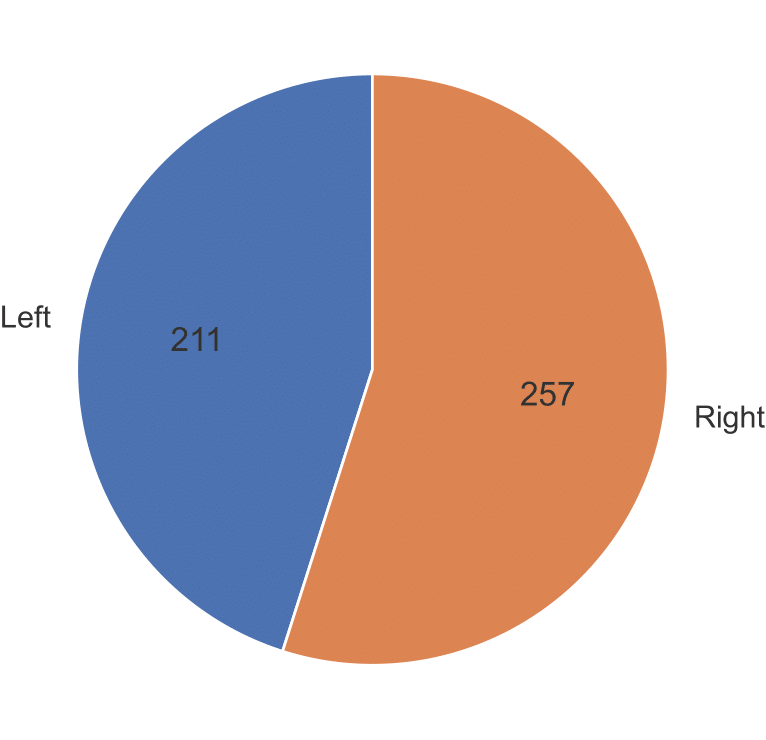}
            \caption{Which sleeve is folded first?}
            \label{fig:which_sleeve_first}
        \end{subfigure}
    \end{minipage}
    \caption{\textbf{Additional BiFold dataset statistics}}
    \label{fig:additional_stats}
\end{figure}

In \cref{fig:which_folds}, we show the distribution of the semantic locations of the origin and end of the manipulation actions. While in the up-down folds, most volunteers prefer to fold top to down, the folds along the left-right direction present an almost equal number of samples from left to right than from right to left.

When referring to folding sleeves, \cref{fig:which_sleeve_first} shows a similar distribution of equal preference for the left and right arm with the latter being slightly more used for the first fold.

Finally, in \cref{fig:refinement_actions} we can see that there is a non-negligible number of refinement actions, which the volunteers use to correct suboptimal folds. The existence of the suboptimal folds is the main motivation for the context that BiFold incorporates, which allows us to keep track of the previous actions.

\clearpage

\section{Experimental setup}

\subsection{Modeling}

For each dataset, we train a single model for all the cloth categories using the resolution of the images in the training set. That means unimanual models ingest a square image with a resolution of 224 pixels, while bimanual models use a higher resolution of 384. All models use a Vision Transformer image backbone \cite{dosovitskiy2021an} that tokenizes the input image by taking $16\times16$ patches. We use the SigLIP base model \cite{siglip} available through Huggingface\footnote{\url{https://huggingface.co/docs/transformers/en/model_doc/siglip}}. We process the RGB images and the input text using the default SigLIP processor. During inference of the bimanual models, we apply the mask to the input image and fill the rest with the background color of the training images.

The SigLIP model is trained with a contrastive objective similar to CLIP \cite{clip} and hence learns to extract a single image embedding and a text embedding that lie in the same space and are close if they have semantic similarities. Instead, we are interested in retrieving tokens for the image and language information to be fused using a transformer \cite{transformers}. Doing so allows incorporating additional conditioning signals as we do in the BiFold version with context and has the potential of adding new modalities. Moreover, this formulation allows the processing of the tokens corresponding to the input image and transforming them back to the image domain to obtain value maps. The tokens we use are the last hidden states of SigLIP, which have dimension 768.

The pretraining of SigLIP on large-scale datasets enables us to learn the invariances of images from data and create a high-level representation of images and text. However, the pretraining objective may focus on distinctive parts of the text that can make a discriminative embedding on the shared space, which does not necessarily align with the representations needed for our problem. With this in mind, we use \gls*{lora} \cite{hu2022lora}, allowing us to modify the inner activations while retaining the knowledge that would not be possible to learn from small datasets such as that from Deng \etal \cite{language_deformable_manipulation}. 

Once we obtain the output tokens of the SigLIP model with the \gls*{lora} modifications, we append a class token to indicate the modality of each input and concatenate the result, \ie, one for RGB images and one for text. For the version with context, we process each RGB image separately, add a single RGB image class token, and add a learned positional embedding element-wise to be able to distinguish time steps and pixel positions.

The resulting tokens are processed using a transformer encoder with 8 blocks having 16 heads each and dimensions 4 times that of the input. The convolutional decoder heads all have the same architecture, which consists of 5 2D convolutional layers with kernel size 1 that halve the number of channels every two layers until reaching a single channel. Each of the layers but the last one is followed by upsampling layers with a scale factor of 2, yielding an image with the same resolution as the input.

\subsection{Additional ablations}

Besides the ablations provided in the paper, we experimented with other design choices that proved unsuccessful during the early development of this project:

\begin{itemize}
    \item \textbf{Decoder models: }Instead of relying on convolutional decoders to obtain the heatmaps, we tried replacing them with transformer decoders similar to those in MAE \cite{mae_2022}. Nevertheless, the performance decreased in this case, and the predicted heatmaps had patch artifacts. We show an example of such artifacts in \cref{fig:artifact}.

    \begin{figure}[ht]
        \centering
        \includegraphics[width=0.5\linewidth]{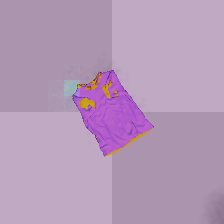}
        \caption{\textbf{Patch artifact: }When using transformer decoders, we observe patch artifacts in the heatmap predictions.}
        \label{fig:artifact}
    \end{figure}
    
    \item \textbf{Predicting segmentation masks: }Since the pick positions have to fall into the cloth region, we use a segmentation mask to enforce this. Instead of assuming that the mask is known or relying on an out-of-the-box segmentation model as we do for real-world samples, we experimented with how to guide a constrained pick prediction without the cloth region as input. To do that, we added an extra decoder head and supervised its output with either cross-entropy loss or a combination of dice and focal losses \cite{focal_loss} as done to train state-of-the-art segmentation models like SAM \cite{sam_Kirillov_2023_ICCV}. The output of this model was then used to restrict the domain of the predicted pick heatmaps. Despite promising and useful in waiving the requirement for input masks, this method offered worse performance.
    
    \item \textbf{Conditioning place position on the pick position: }Given the interplay between the prediction of pick and place positions, we experimented with an ancestral sampling approach. To do that, we first predicted pick positions and then used the output to condition the place prediction. This method offered no notable benefits, showing that the tokens at the output of the decoder contain enough information that the decoders know how to multiplex to obtain the correct pick and place positions.
\end{itemize}

\subsection{Simulation}

The pick and place manipulation primitive uses the following steps:
\begin{enumerate}
    \item Set pick and place heights using the radius of the picker, regardless of the world coordinate of the vertex.
    \item The picker is moved to the picking position but at a predefined height.
    \item The picker moves to the pick position and closes the gripper.
    \item The picker moves to the position in 2).
    \item The picker is moved to the placing position but at a predefined height.
    \item The picker goes to the placing position and opens the gripper.
    \item The picker moves to the place position at the same predefined height as in 2).
\end{enumerate}
All the movements are performed at a speed of 5 mm/action except steps (2) and (7), which we perform 100 times faster as they are supposed to not interact with the cloth. The bimanual primitive uses the unimanual primitive with the actions executed at the same time for both pickers.

\subsection{Real}

\begin{figure}[ht]
    \centering
    \begin{subfigure}{0.24\linewidth}
        \centering
        \includegraphics[width=\linewidth]{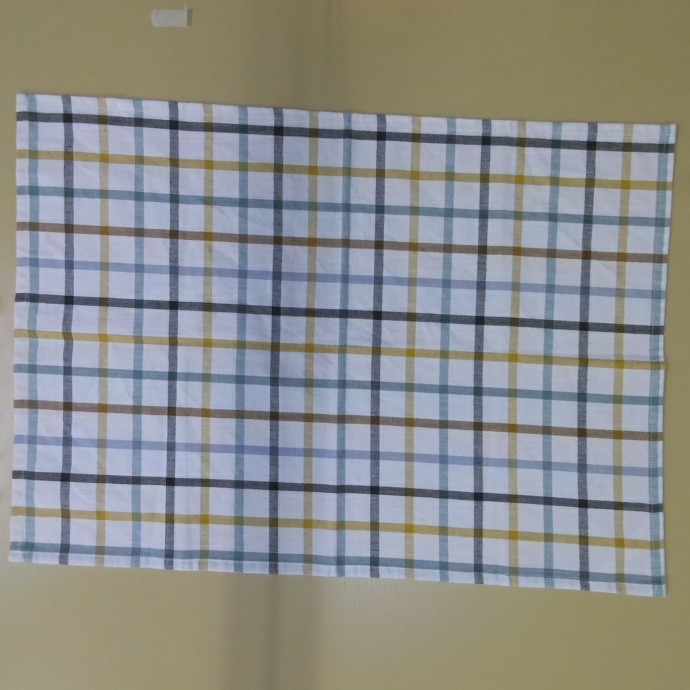}
        \caption{Checkered rag \cite{household_dataset}}
    \end{subfigure}
    \begin{subfigure}{0.24\linewidth}
        \centering
        \includegraphics[width=\linewidth]{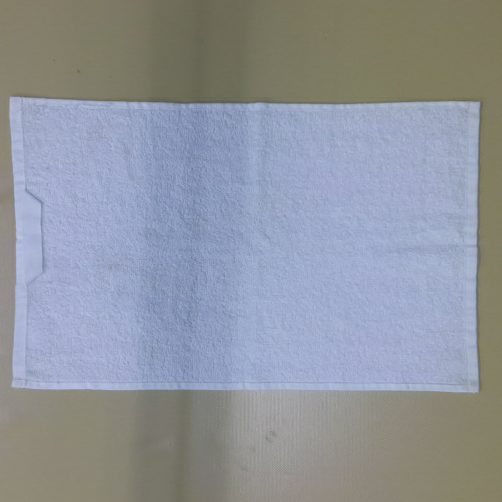}
        \caption{Small towel \cite{household_dataset}}
    \end{subfigure}
    \begin{subfigure}{0.24\linewidth}
        \centering
        \includegraphics[width=\linewidth]{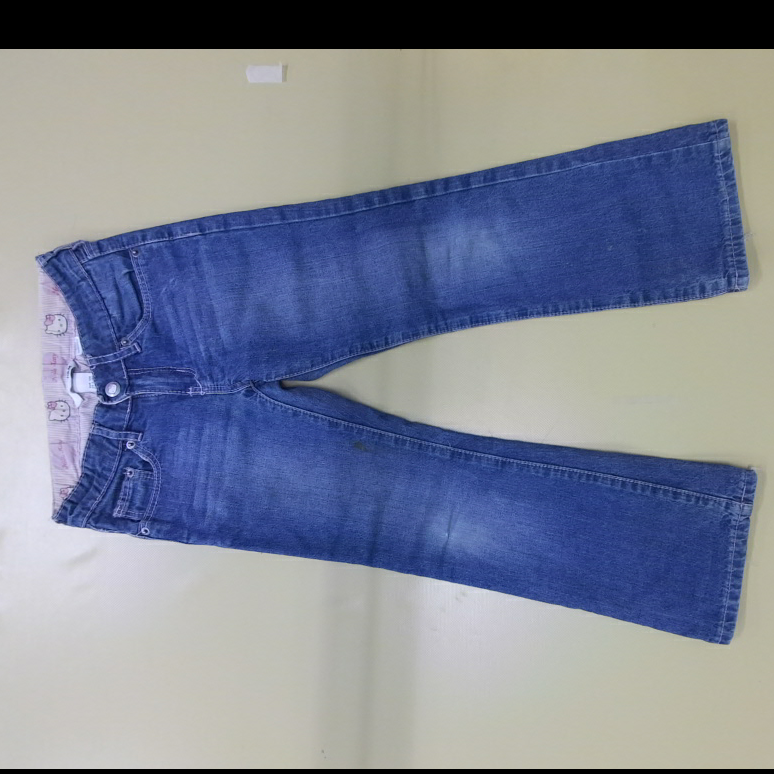}
        \caption{Jeans 1}
    \end{subfigure}
    \begin{subfigure}{0.24\linewidth}
        \centering
        \includegraphics[width=\linewidth]{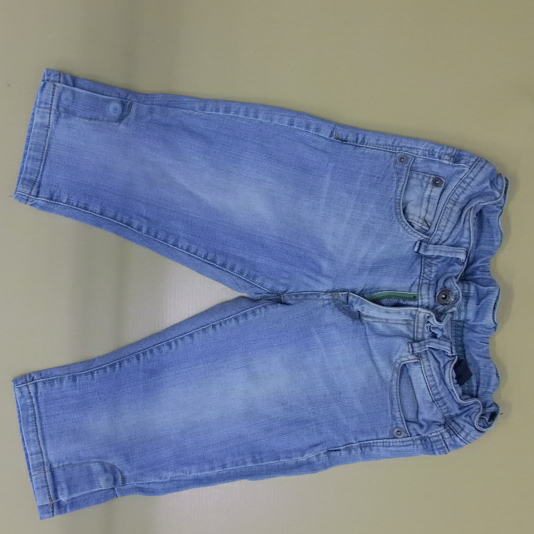}
        \caption{Jeans 2}
    \end{subfigure}
    \begin{subfigure}{0.24\linewidth}
        \centering
        \includegraphics[width=\linewidth]{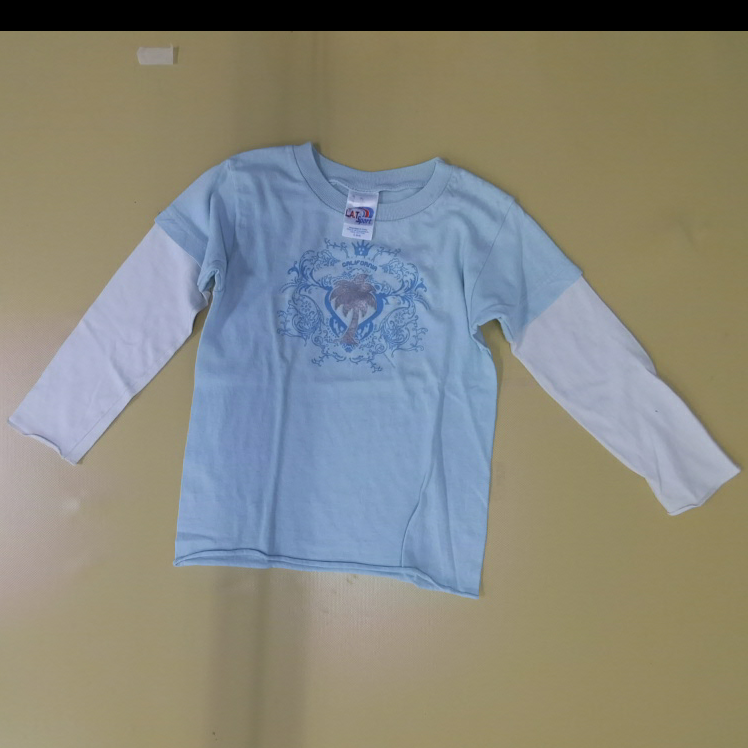}
        \caption{Long-sleeved T-shirt}
    \end{subfigure}
    \begin{subfigure}{0.24\linewidth}
        \centering
        \includegraphics[width=\linewidth]{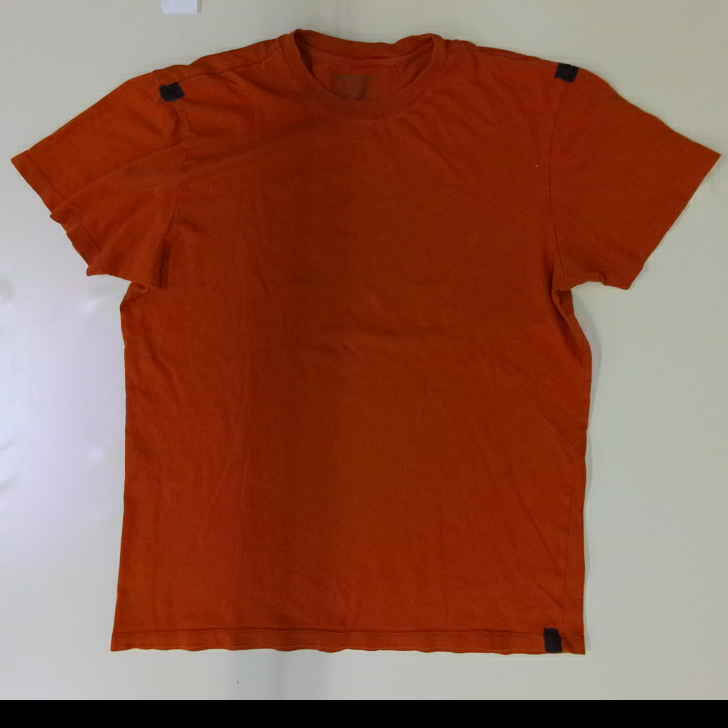}
        \caption{Short-sleeved T-shirt 1}
    \end{subfigure}
    \begin{subfigure}{0.24\linewidth}
        \centering
        \includegraphics[width=\linewidth]{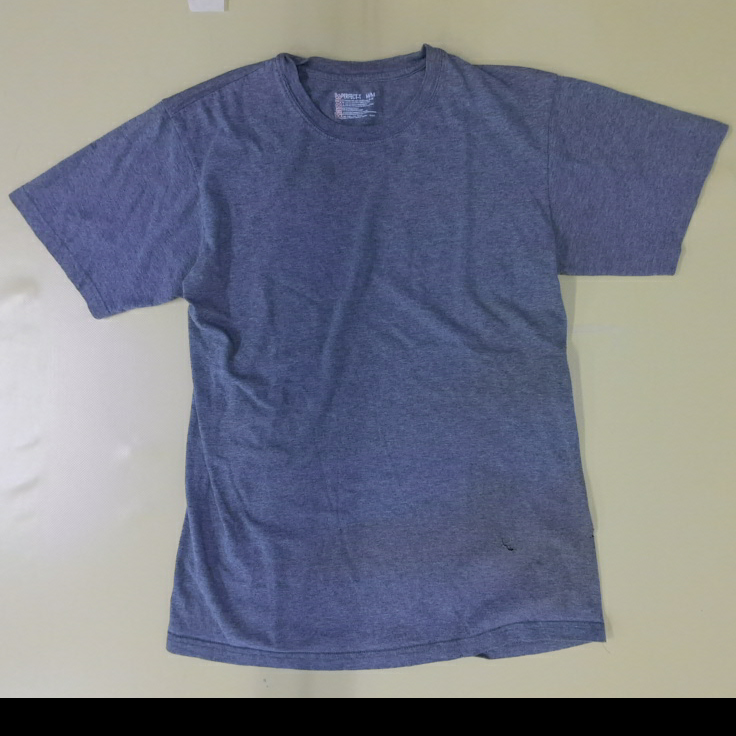}
        \caption{Short-sleeved T-shirt 2}
    \end{subfigure}
    \begin{subfigure}{0.24\linewidth}
        \centering
        \includegraphics[width=\linewidth]{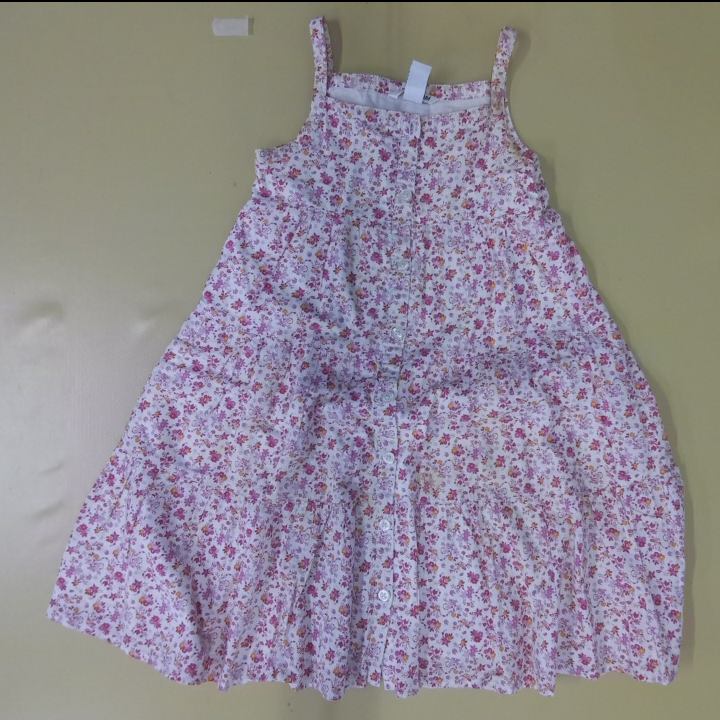}
        \caption{Dress}
    \end{subfigure}
    \caption{\textbf{Real dataset: }The real dataset is composed of eight garments with different materials, topologies, and textures. In this figure we show the initial configuration, where the clothes are flattened on a table.}
    \label{fig:real_dataset}
\end{figure}

To obtain segmentation masks, we use the ViT-h model variant of SAM \cite{sam_Kirillov_2023_ICCV} using pixel coordinates as input prompts. The masks are then used to determine square crops of the images provided by the Azure Kinect camera, which provides 1280x720 pixel images. To reduce the influence of the noise of the depth sensor, we take 10 images and use the median value for each pixel. In particular, we compose axis aligned bounding boxes taking all the images of a given folding sequence. Then, we crop the images to a size determined as the maximum
side length of the bounding box of all the images and a margin of 10 pixels added to prevent pick and place positions from falling on the border of the image. If needed, we pad the images using constant padding with value 0. In \cref{fig:real_dataset} we present the resulting cropped images for the clothes used in the real setup.

We evaluate our model on eight pieces of cloth: The checkered rag and small towel from the public household dataset \cite{household_dataset}, two jeans, a long-sleeved and two short-sleeved T-shirts, and a dress. The rag, towel, and dress are unseen clothing categories. The jeans have a new material different from the smooth fabrics obtained in rendering. The long-sleeved T-shirt is a double-layer tee that does not appear in any training example. Besides, we record the dataset in a real setup with new lighting conditions, obtain shadows and reflexes, and the garments have different textures. Overall, the real dataset has a significant shift in the distribution of the inputs.

\section{Additional qualitative evaluation}

\begin{figure*}[ht]
    \centering
    \captionsetup[subfigure]{labelformat=empty}
    \begin{subfigure}[t]{0.19\linewidth}%
        \centering
        \includegraphics[width=\linewidth]{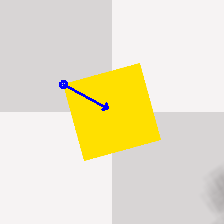}
        \caption{\scriptsize [USI] \texttt{Make a crease at the leftmost top corner of the cloth and fold it towards the center.}}
    \end{subfigure}
    \begin{subfigure}[t]{0.19\linewidth}
        \centering
        \includegraphics[width=\linewidth]{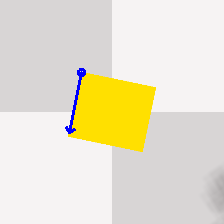}
        \caption{\scriptsize [USI] \texttt{Make a fold in the cloth by halving it from top to lowermost.}}
    \end{subfigure}
    \begin{subfigure}[t]{0.19\linewidth}
        \centering
        \includegraphics[width=\linewidth]{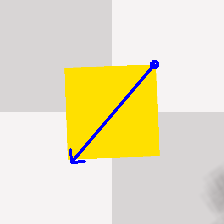}
        \caption{\scriptsize [UT] \texttt{Fold the rightmost top corner of the fabric to its diagonal corner.}}
    \end{subfigure}
    \begin{subfigure}[t]{0.19\linewidth}
        \centering
        \includegraphics[width=\linewidth]{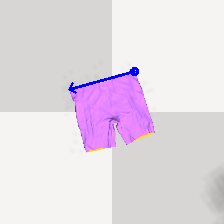}
        \caption{\scriptsize [UT] \texttt{Fold the Trousers in half, with the rightmost side overlapping the leftmost.}}
    \end{subfigure}
    \begin{subfigure}[t]{0.19\linewidth}    
        \centering
        \includegraphics[width=\linewidth]{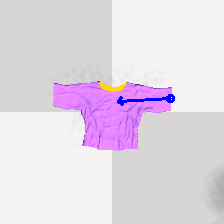}
        \caption{\scriptsize [USI] \texttt{Fold the right-hand sleeve to the centerline of the shirt.}}
    \end{subfigure}
    \begin{subfigure}[t]{0.19\linewidth}%
        \centering
        \includegraphics[width=\linewidth]{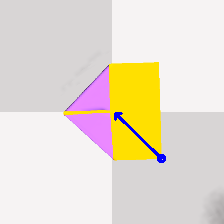}
        \caption{\scriptsize [UT] \texttt{Create a fold from the lower right corner of the cloth towards the center.}}
    \end{subfigure}
    \begin{subfigure}[t]{0.19\linewidth}
        \centering
        \includegraphics[width=\linewidth]{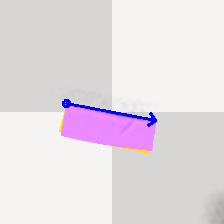}
        \caption{\scriptsize [SI] \texttt{Fold the fabric in half, starting from the left side.}}
    \end{subfigure}
    \begin{subfigure}[t]{0.19\linewidth}
        \centering
        \includegraphics[width=\linewidth]{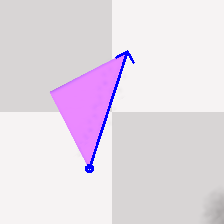}
        \caption{\scriptsize [USI] \texttt{Bring the bottommost left corner of the cloth down to the right upper corner, folding it in half diagonally.}}
    \end{subfigure}
    \begin{subfigure}[t]{0.19\linewidth}    
        \centering
        \includegraphics[width=\linewidth]{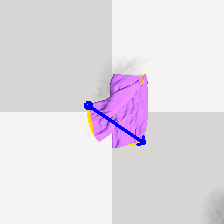}
        \caption{\scriptsize [USI] \texttt{Fold the Trousers, making a crease from the leftmost to the rightmost.}}
    \end{subfigure}
    \begin{subfigure}[t]{0.19\linewidth}
        \centering
        \includegraphics[width=\linewidth]{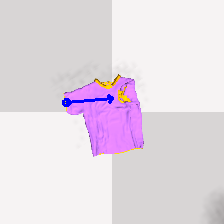}
        \caption{\scriptsize [UT] \texttt{Fold the left-hand sleeve towards the inside.}}
    \end{subfigure}
    \begin{subfigure}[t]{0.19\linewidth}%
        \centering
        \includegraphics[width=\linewidth]{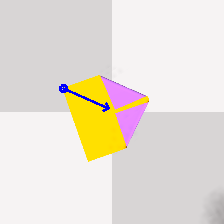}
        \caption{\scriptsize [USI] \texttt{Make a crease at the leftmost top corner of the cloth and fold it towards the center.}}
    \end{subfigure}
    \begin{subfigure}[t]{0.19\linewidth}
        \centering
        \includegraphics[width=\linewidth]{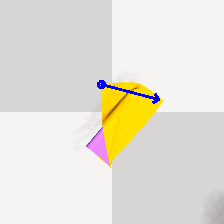}
        \caption{\scriptsize [USI] \texttt{Fold the cloth in half, starting from the leftmost side and meeting the right.}}
    \end{subfigure}
    \begin{subfigure}[t]{0.19\linewidth}
        \centering
        \includegraphics[width=\linewidth]{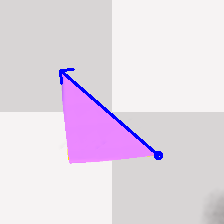}
        \caption{\scriptsize [UT] \texttt{Make a fold from the bottommost right corner of the fabric to the top left-hand.}}
    \end{subfigure}
    \begin{subfigure}[t]{0.19\linewidth}    
        \centering
        \includegraphics[width=\linewidth]{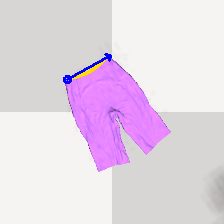}
        \caption{\scriptsize [SI] \texttt{Make a fold in the Trousers, starting from the left-hand and ending at the rightmost.}}
    \end{subfigure}
    \begin{subfigure}[t]{0.19\linewidth}
        \centering
        \includegraphics[width=\linewidth]{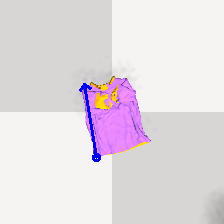}
        \caption{\scriptsize [SI] \texttt{Flip the bottom of the T-shirt towards the top.}}
    \end{subfigure}
    \begin{subfigure}[t]{0.19\linewidth}%
        \centering
        \includegraphics[width=\linewidth]{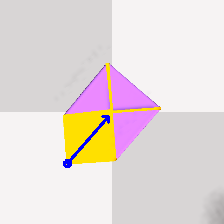}
        \caption{\scriptsize [SI] \texttt{Bring the bottom left-hand corner of the cloth to the middle with a fold.}}
    \end{subfigure}
    \begin{subfigure}[t]{0.19\linewidth}
        \centering
        \includegraphics[width=\linewidth]{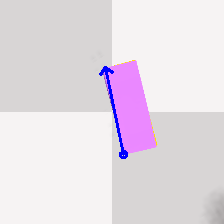}
        \caption{\scriptsize [UT] \texttt{Fold the textile symmetrically, starting on the bottom.}}
    \end{subfigure}
    \begin{subfigure}[t]{0.19\linewidth}
        \centering
        \includegraphics[width=\linewidth]{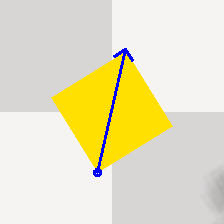}
        \caption{\scriptsize [SI] \texttt{Fold the trousers in half horizontally, top left to right.}}
    \end{subfigure}
    \begin{subfigure}[t]{0.19\linewidth}    
        \centering
        \includegraphics[width=\linewidth]{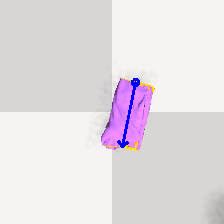}
        \caption{\scriptsize [SI] \texttt{Fold the Trousers by bringing the waistband down to meet the bottom.}}
    \end{subfigure}
    \begin{subfigure}[t]{0.19\linewidth}
        \centering
        \includegraphics[width=\linewidth]{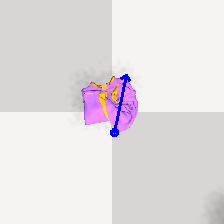}
        \caption{\scriptsize [SI] \texttt{Tuck the bottom of the T-shirt upwards.}}
    \end{subfigure}
    \caption{\textbf{Qualitative unimanual SoftGym: }For the unimanual version, training data comes from SoftGym and is generated through scripted movements, leading to high success rates. In each example, we indicate the type of instruction: Seen Instruction (SI), UnSeen Instruction (USI), or Unseen Task (UT).}
    \label{fig:qualitative_unimanual}
\end{figure*}

\begin{figure*}[ht]
    \centering
    \captionsetup[subfigure]{labelformat=empty}
    \begin{subfigure}[t]{0.19\linewidth}    
        \centering
        \includegraphics[width=\linewidth]{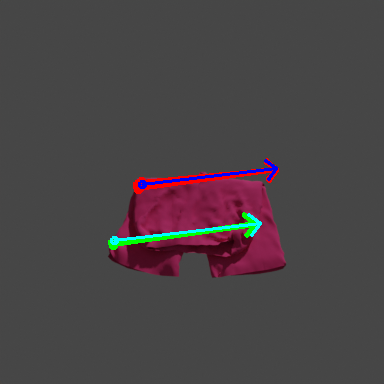}
        \caption{\scriptsize \texttt{Fold the trousers, orientating from the right towards the left.}}
    \end{subfigure}
    \begin{subfigure}[t]{0.19\linewidth}
        \centering
        \includegraphics[width=\linewidth]{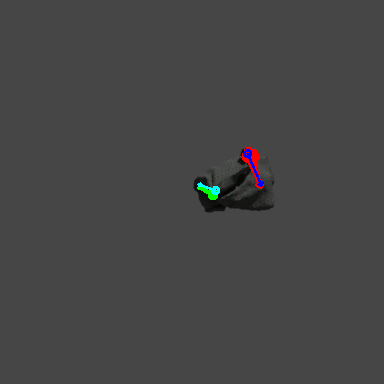}
        \caption{\scriptsize \texttt{Fold the top neatly, from the right side to the bottom side.}}
    \end{subfigure}
    \begin{subfigure}[t]{0.19\linewidth}
        \centering
        \includegraphics[width=\linewidth]{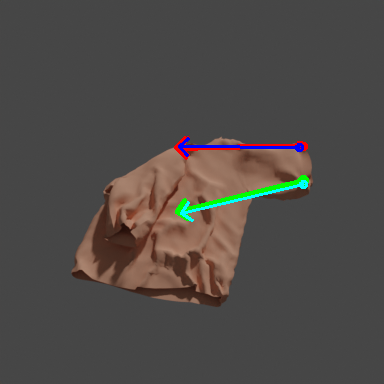}
        \caption{\scriptsize \texttt{Fold the left sleeve to the centerline of the shirt.}}
    \end{subfigure}
    \begin{subfigure}[t]{0.19\linewidth}
        \centering
        \includegraphics[width=\linewidth]{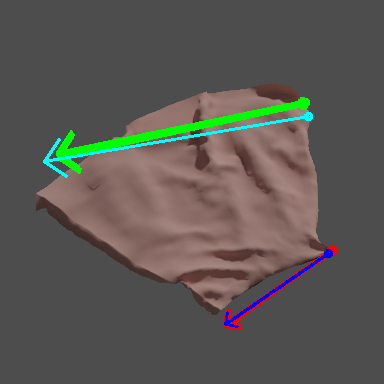}
        \caption{\scriptsize \texttt{Fold the skirt cleanly, from the right side to the left side.}}
    \end{subfigure}
    \begin{subfigure}[t]{0.19\linewidth}    
        \centering
        \includegraphics[width=\linewidth]{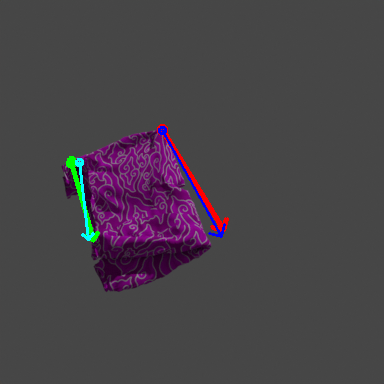}
        \caption{\scriptsize \texttt{Fold the trousers in half, bottom right to left.}}
    \end{subfigure}
    \begin{subfigure}[t]{0.19\linewidth}
        \centering
        \includegraphics[width=\linewidth]{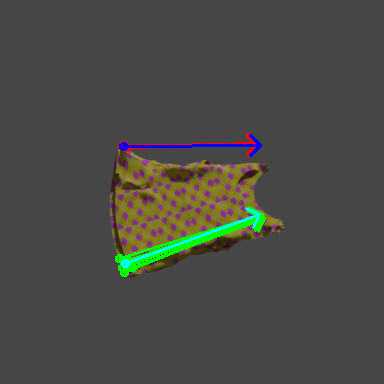}
        \caption{\scriptsize \texttt{Fold the top, making sure the bottom side touches the top side.}}
    \end{subfigure}
    \begin{subfigure}[t]{0.19\linewidth}
        \centering
        \includegraphics[width=\linewidth]{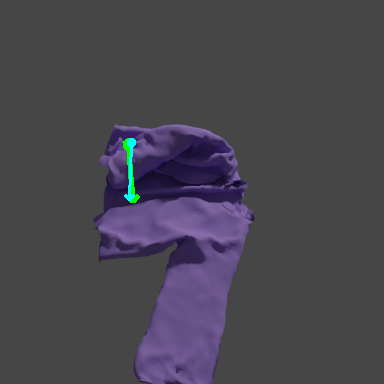}
        \caption{\scriptsize \texttt{Fold the right sleeve towards the body only using the right arm.}}
    \end{subfigure}
    \begin{subfigure}[t]{0.19\linewidth}
        \centering
        \includegraphics[width=\linewidth]{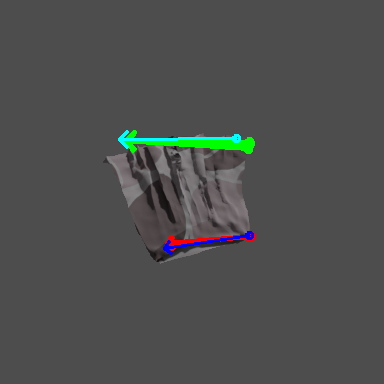}
        \caption{\scriptsize \texttt{Bend the skirt in half, from right to left.}}
    \end{subfigure}
    \begin{subfigure}[t]{0.19\linewidth}    
        \centering
        \includegraphics[width=\linewidth]{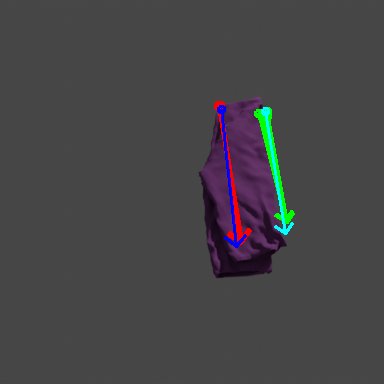}
        \caption{\scriptsize \texttt{Fold the trousers in half, with the top side overlapping the bottom.}}
    \end{subfigure}
    \begin{subfigure}[t]{0.19\linewidth}
        \centering
        \includegraphics[width=\linewidth]{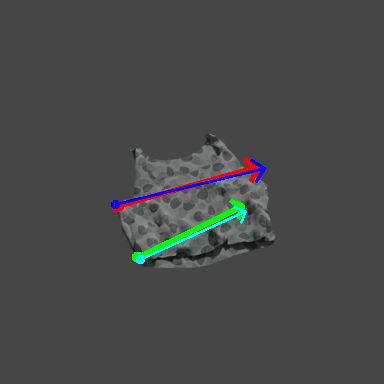}
        \caption{\scriptsize \texttt{Fold the top, making a crease from the right to the left.}}
    \end{subfigure}
    \begin{subfigure}[t]{0.19\linewidth}
        \centering
        \includegraphics[width=\linewidth]{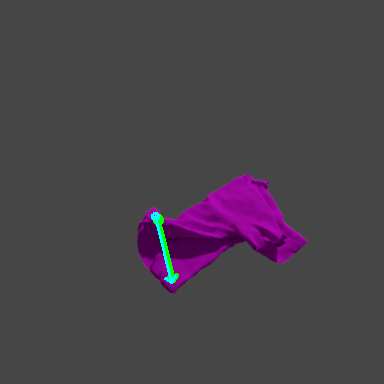}
        \caption{\scriptsize \texttt{Fold the tshirt in half, with the bottom right side overlapping the bottom left only using the right arm.}}
    \end{subfigure}
    \begin{subfigure}[t]{0.19\linewidth}
        \centering
        \includegraphics[width=\linewidth]{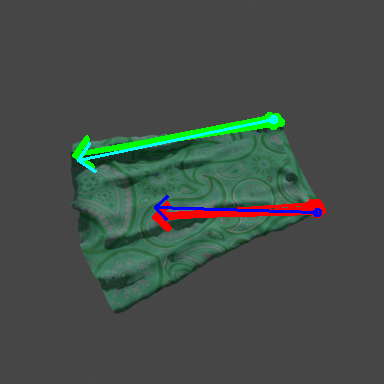}
        \caption{\scriptsize \texttt{Fold the skirt, bringing the top side to meet the bottom side.}}
    \end{subfigure}
    \begin{subfigure}[t]{0.19\linewidth}    
        \centering
        \includegraphics[width=\linewidth]{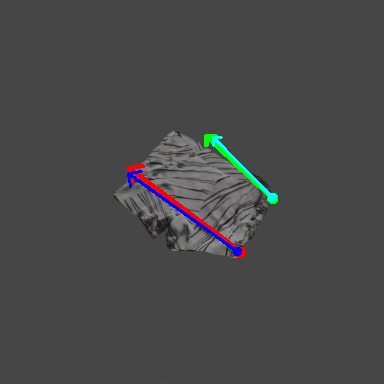}
        \caption{\scriptsize \texttt{Fold the trousers in half, starting from the left and ending at the right.}}
    \end{subfigure}
    \begin{subfigure}[t]{0.19\linewidth}
        \centering
        \includegraphics[width=\linewidth]{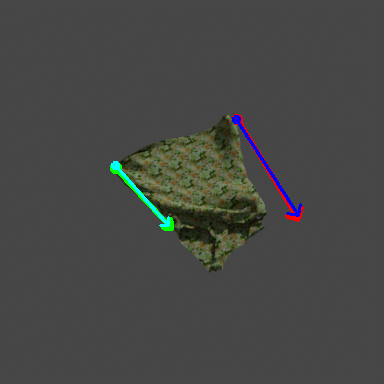}
        \caption{\scriptsize \texttt{Bring the bottom side of the top towards the top side and fold them in half.}}
    \end{subfigure}
    \begin{subfigure}[t]{0.19\linewidth}
        \centering
        \includegraphics[width=\linewidth]{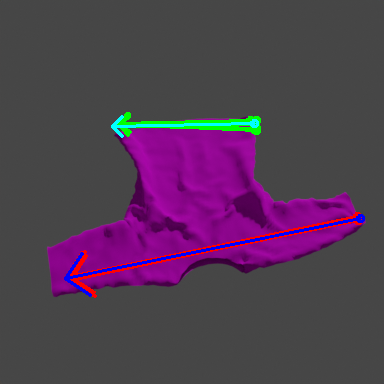}
        \caption{\scriptsize \texttt{Fold the tshirt, making a crease from the right to the left.}}
    \end{subfigure}
    \begin{subfigure}[t]{0.19\linewidth}
        \centering
        \includegraphics[width=\linewidth]{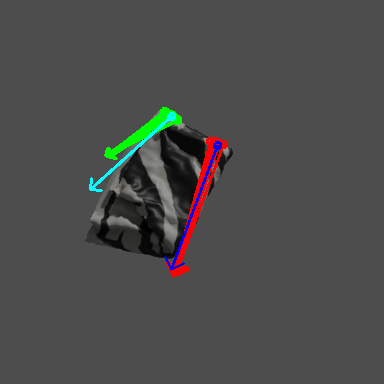}
        \caption{\scriptsize \texttt{Fold the skirt in half, aligning the right and left sides.}}
    \end{subfigure}
    \begin{subfigure}[t]{0.19\linewidth}    
        \centering
        \includegraphics[width=\linewidth]{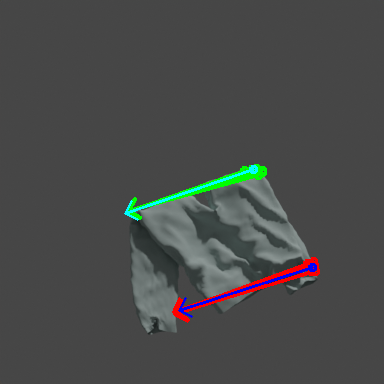}
        \caption{\scriptsize \texttt{Fold the trousers, bottom left side over bottom right side.}}
    \end{subfigure}
    \begin{subfigure}[t]{0.19\linewidth}
        \centering
        \includegraphics[width=\linewidth]{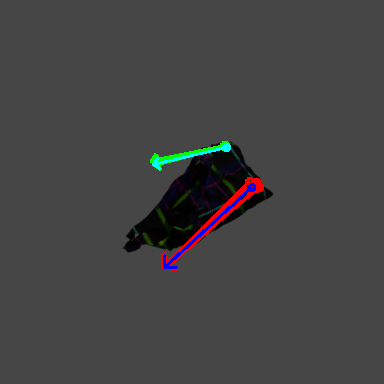}
        \caption{\scriptsize \texttt{Make a fold in the top, starting from the bottom and ending at the top.}}
    \end{subfigure}
    \begin{subfigure}[t]{0.19\linewidth}
        \centering
        \includegraphics[width=\linewidth]{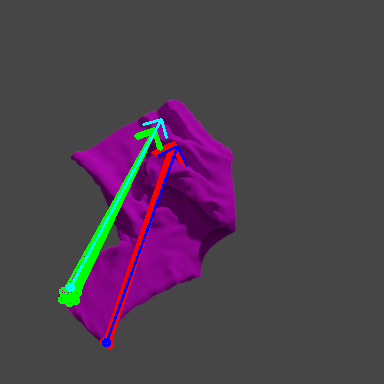}
        \caption{\scriptsize \texttt{Fold the left sleeve inward to the halfway point.}}
    \end{subfigure}
    \begin{subfigure}[t]{0.19\linewidth}
        \centering
        \includegraphics[width=\linewidth]{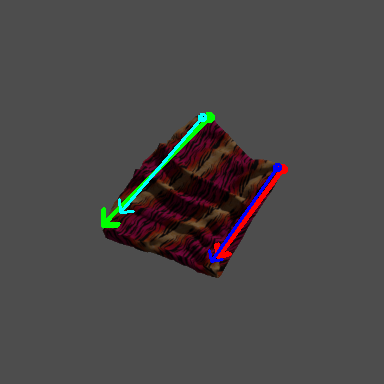}
        \caption{\scriptsize \texttt{Fold the skirt neatly, from the right side to the left side.}}
    \end{subfigure}
    \caption{\textbf{Additional qualitative examples on our dataset: }BiFold can learn different folding patterns conditioned on language and a starting folding position. The obtained actions successfully replicate the human demonstrations with different sizes, cloth textures, and camera positions.}
    \label{fig:qualitative_sim_extra}
\end{figure*}

\begin{figure*}[ht]
    \centering
    \captionsetup[subfigure]{labelformat=empty}
    \begin{subfigure}[t]{0.19\linewidth}    
        \centering
        \includegraphics[width=\linewidth]{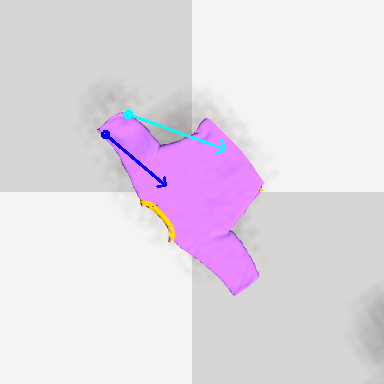}
        \caption{\scriptsize \texttt{Fold the left sleeve towards the midpoint of the shirt.}}
    \end{subfigure}
    \begin{subfigure}[t]{0.19\linewidth}
        \centering
        \includegraphics[width=\linewidth]{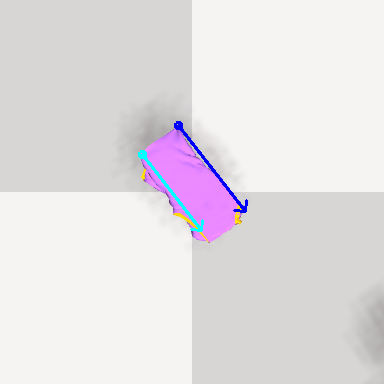}
        \caption{\scriptsize \texttt{Make a fold in the tshirt, starting from the bottom left and ending at the top.}}
    \end{subfigure}
    \begin{subfigure}[t]{0.19\linewidth}
        \centering
        \includegraphics[width=\linewidth]{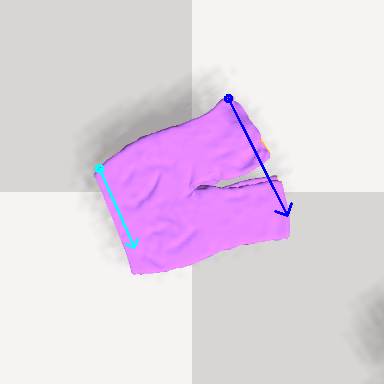}
        \caption{\scriptsize \texttt{Fold the trousers neatly, from the left side to the right side.}}
    \end{subfigure}
    \begin{subfigure}[t]{0.19\linewidth}
        \centering
        \includegraphics[width=\linewidth]{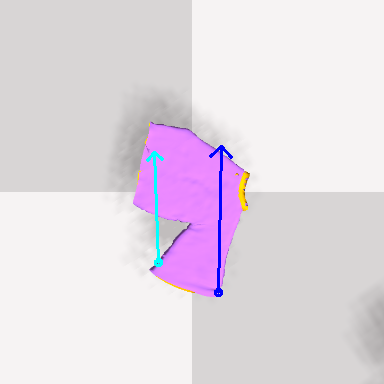}
        \caption{\scriptsize \texttt{Fold the tshirt, left side over right side.}}
    \end{subfigure}
    \begin{subfigure}[t]{0.19\linewidth}    
        \centering
        \includegraphics[width=\linewidth]{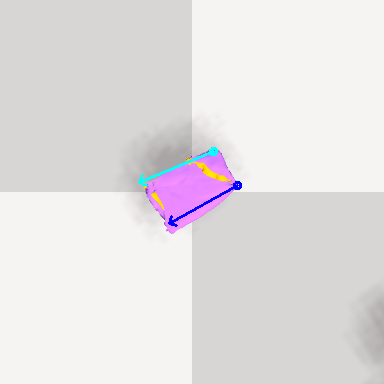}
        \caption{\scriptsize \texttt{Fold the top neatly, from the bottom left side to the bottom right side.}}
    \end{subfigure}
    \begin{subfigure}[t]{0.19\linewidth}
        \centering
        \includegraphics[width=\linewidth]{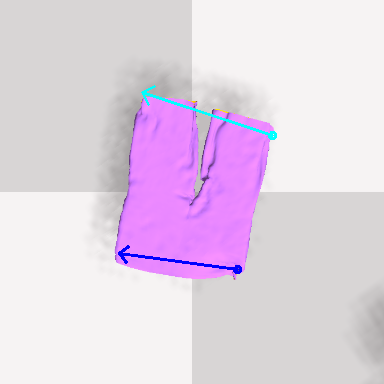}
        \caption{\scriptsize \texttt{Make a fold in the trousers, starting from the right and ending at the left.}}
    \end{subfigure}
    \begin{subfigure}[t]{0.19\linewidth}
        \centering
        \includegraphics[width=\linewidth]{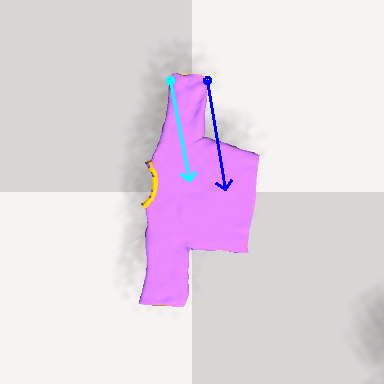}
        \caption{\scriptsize \texttt{Bring the left sleeve to the center.}}
    \end{subfigure}
    \begin{subfigure}[t]{0.19\linewidth}
        \centering
        \includegraphics[width=\linewidth]{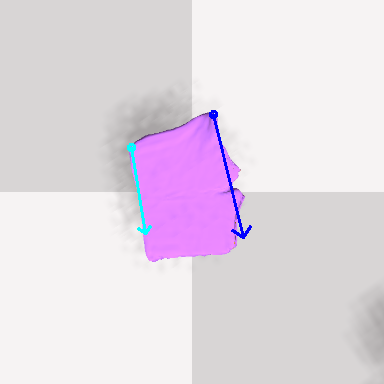}
        \caption{\scriptsize \texttt{Fold the trousers in half horizontally, top left to right.}}
    \end{subfigure}
    \begin{subfigure}[t]{0.19\linewidth}    
        \centering
        \includegraphics[width=\linewidth]{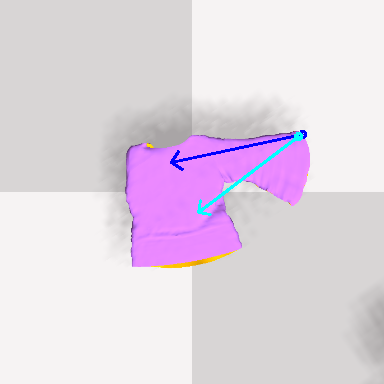}
        \caption{\scriptsize \texttt{Fold the tshirt from the left side towards the right side.}}
    \end{subfigure}
    \begin{subfigure}[t]{0.19\linewidth}
        \centering
        \includegraphics[width=\linewidth]{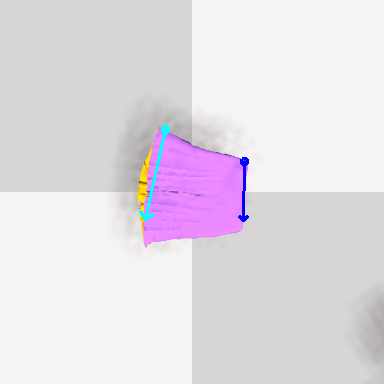}
        \caption{\scriptsize \texttt{Bend the skirt in half, from right to left.}}
    \end{subfigure}
    \begin{subfigure}[t]{0.19\linewidth}
        \centering
        \includegraphics[width=\linewidth]{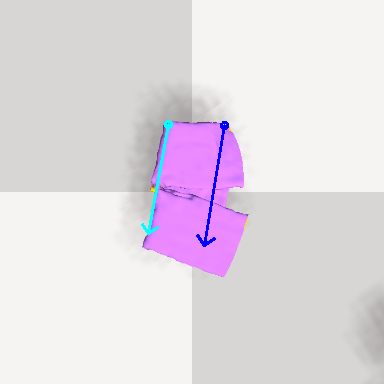}
        \caption{\scriptsize \texttt{Fold the trousers, left side over right side.}}
    \end{subfigure}
    \begin{subfigure}[t]{0.19\linewidth}
        \centering
        \includegraphics[width=\linewidth]{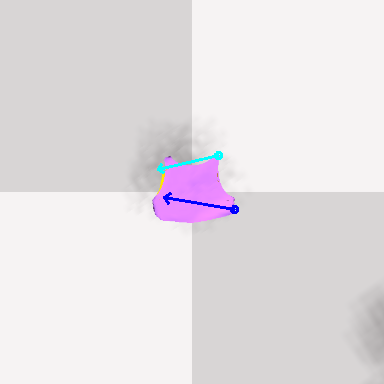}
        \caption{\scriptsize \texttt{Fold the top in half, starting from the left and ending at the right.}}
    \end{subfigure}
    \begin{subfigure}[t]{0.19\linewidth}    
        \centering
        \includegraphics[width=\linewidth]{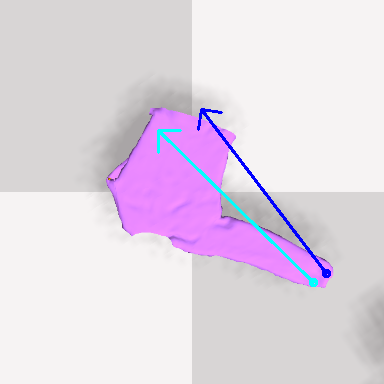}
        \caption{\scriptsize \texttt{Fold the right sleeve to the centerline of the shirt.}}
    \end{subfigure}
    \begin{subfigure}[t]{0.19\linewidth}
        \centering
        \includegraphics[width=\linewidth]{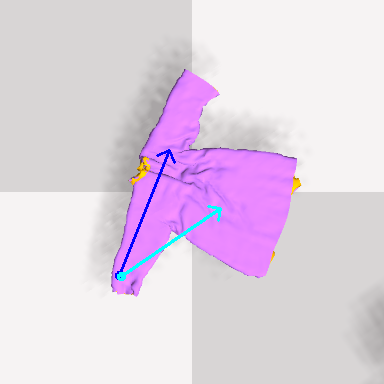}
        \caption{\scriptsize \texttt{Fold the right sleeve inward to the halfway point.}}
    \end{subfigure}
    \begin{subfigure}[t]{0.19\linewidth}
        \centering
        \includegraphics[width=\linewidth]{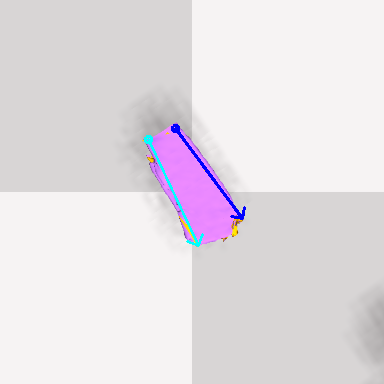}
        \caption{\scriptsize \texttt{Fold the tshirt cleanly, from the bottom side to the top side.}}
    \end{subfigure}
    \begin{subfigure}[t]{0.19\linewidth}
        \centering
        \includegraphics[width=\linewidth]{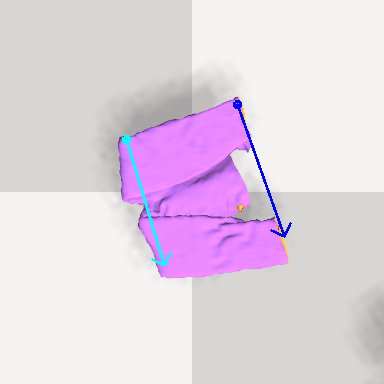}
        \caption{\scriptsize \texttt{Bend the trousers in half, from bottom left to bottom right.}}
    \end{subfigure}
    \begin{subfigure}[t]{0.19\linewidth}    
        \centering
        \includegraphics[width=\linewidth]{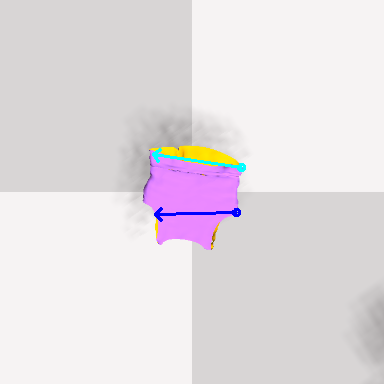}
        \caption{\scriptsize \texttt{Fold the top, orientating from the right towards the left.}}
    \end{subfigure}
    \begin{subfigure}[t]{0.19\linewidth}
        \centering
        \includegraphics[width=\linewidth]{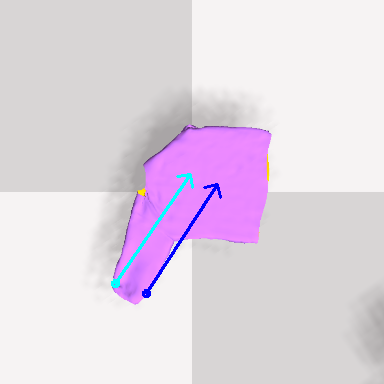}
        \caption{\scriptsize \texttt{Fold the right sleeve to the center.}}
    \end{subfigure}
    \begin{subfigure}[t]{0.19\linewidth}
        \centering
        \includegraphics[width=\linewidth]{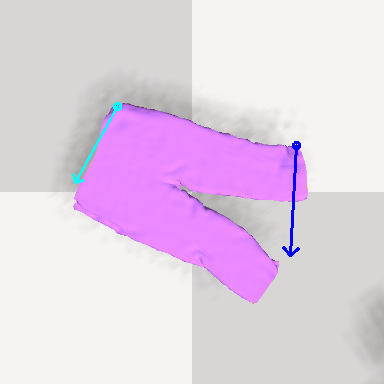}
        \caption{\scriptsize \texttt{Fold the trousers, making a crease from the left to the right.}}
    \end{subfigure}
    \begin{subfigure}[t]{0.19\linewidth}
        \centering
        \includegraphics[width=\linewidth]{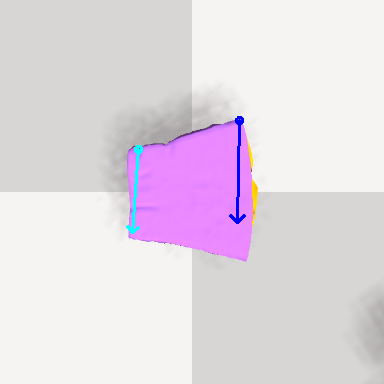}
        \caption{\scriptsize \texttt{Fold the skirt cleanly, from the left side to the right side.}}
    \end{subfigure}
    \caption{\textbf{Qualitative bimanual SoftGym: }Despite the apparent distribution shift between the training dataset and the SoftGym environment, BiFold is able to predict coherent actions.}
    \label{fig:qualitative_softgym}
\end{figure*}

\begin{figure}[ht]
    \centering
    \captionsetup[subfigure]{labelformat=empty}
    \begin{subfigure}[t]{0.24\linewidth}
        \centering
        \includegraphics[width=\linewidth]{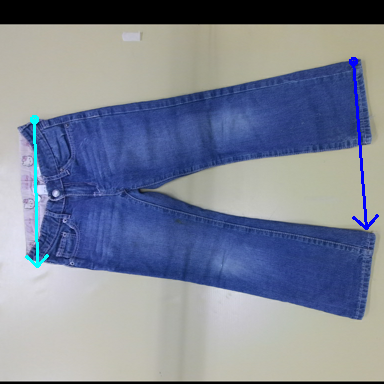}
        \caption{\scriptsize \texttt{Fold the trousers in half, with the left side overlapping the right.}}
    \end{subfigure}
    \begin{subfigure}[t]{0.24\linewidth}
        \centering
        \includegraphics[width=\linewidth]{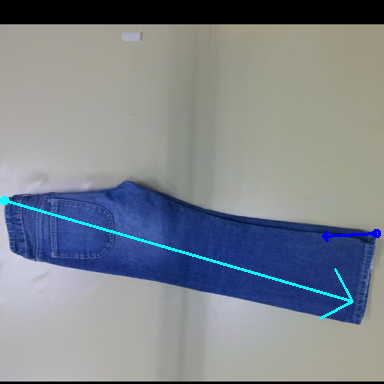}
        \caption{\scriptsize \texttt{Fold the trousers, top side over bottom side.}}
    \end{subfigure}
    \begin{subfigure}[t]{0.24\linewidth}
        \centering
        \includegraphics[width=\linewidth]{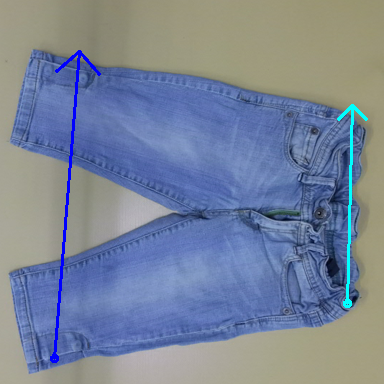}
        \caption{\scriptsize \texttt{Fold the trousers, left side over right side.}}
    \end{subfigure}
    \begin{subfigure}[t]{0.24\linewidth}
        \centering
        \includegraphics[width=\linewidth]{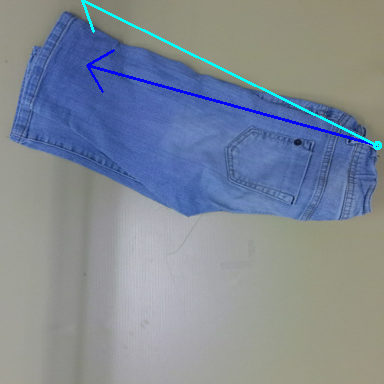}
        \caption{\scriptsize \texttt{Fold the trousers, making a crease from the top to the bottom.}}
    \end{subfigure}
    \begin{subfigure}[t]{0.24\linewidth}    
        \centering
        \includegraphics[width=\linewidth]{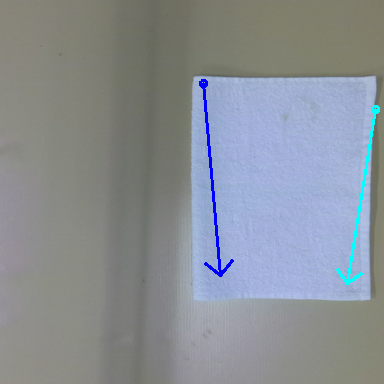}
        \caption{\scriptsize \texttt{Create a fold in the towel, going from top to bottom.}}
    \end{subfigure}
    \begin{subfigure}[t]{0.24\linewidth}
        \centering
        \includegraphics[width=\linewidth]{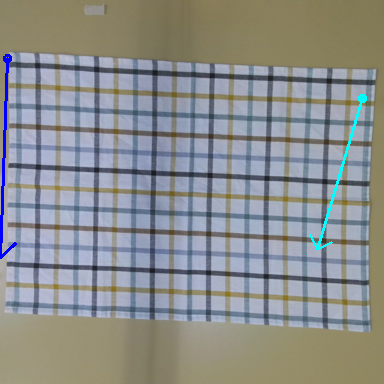}
        \caption{\scriptsize \texttt{Fold the towel in half, aligning the left and right sides.}}
    \end{subfigure}
    \begin{subfigure}[t]{0.24\linewidth}
        \centering
        \includegraphics[width=\linewidth]{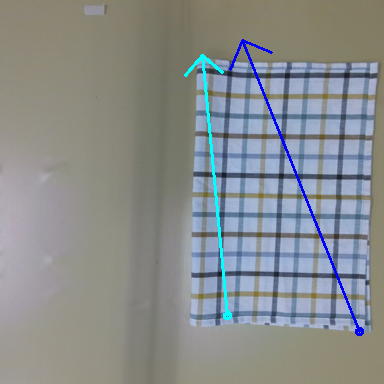}
        \caption{\scriptsize \texttt{Fold the cloth in half, aligning the top and bottom sides.}}
    \end{subfigure}
    \begin{subfigure}[t]{0.24\linewidth}
        \centering
        \includegraphics[width=\linewidth]{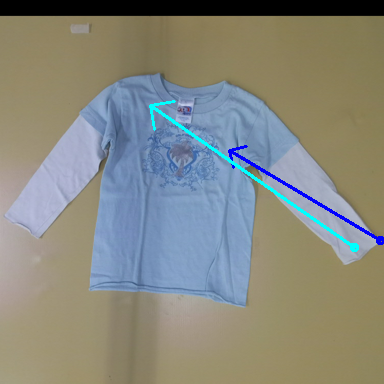}
        \caption{\scriptsize \texttt{Fold the left sleeve to the center.}}
    \end{subfigure}
    \begin{subfigure}[t]{0.24\linewidth}
        \centering
        \includegraphics[width=\linewidth]{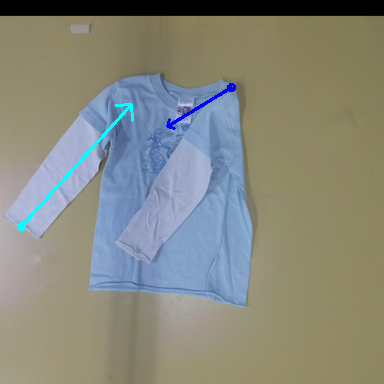}
        \caption{\scriptsize \texttt{Fold the right sleeve towards the body.}}
    \end{subfigure}
    \begin{subfigure}[t]{0.24\linewidth}
        \centering
        \includegraphics[width=\linewidth]{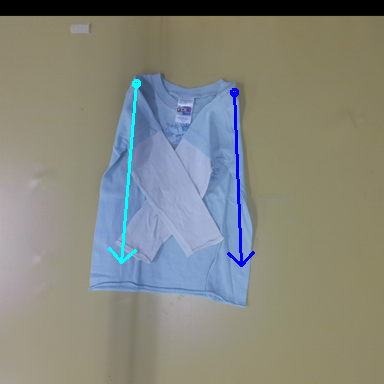}
        \caption{\scriptsize \texttt{Fold the tshirt, top side over bottom side.}}
    \end{subfigure}
    \begin{subfigure}[t]{0.24\linewidth}
        \centering
        \includegraphics[width=\linewidth]{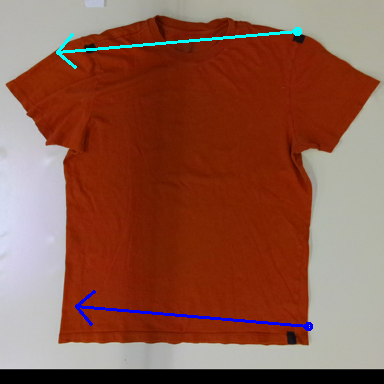}
        \caption{\scriptsize \texttt{Bend the tshirt in half, from left to right.}}
    \end{subfigure}
    \begin{subfigure}[t]{0.24\linewidth}
        \centering
        \includegraphics[width=\linewidth]{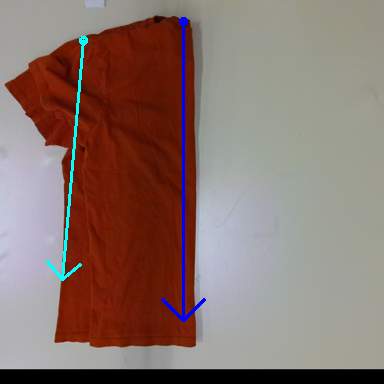}
        \caption{\scriptsize \texttt{Fold the tshirt, making sure the top side touches the bottom side.}}
    \end{subfigure}
    \begin{subfigure}[t]{0.24\linewidth}
        \centering
        \includegraphics[width=\linewidth]{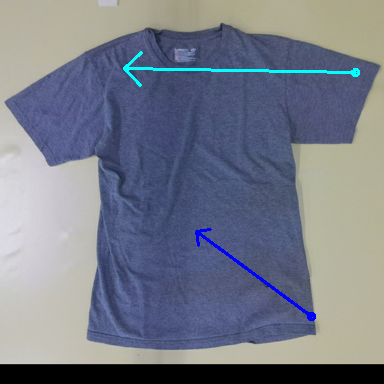}
        \caption{\scriptsize \texttt{Fold the tshirt in half, aligning the left and right sides.}}
    \end{subfigure}
    \begin{subfigure}[t]{0.24\linewidth}
        \centering
        \includegraphics[width=\linewidth]{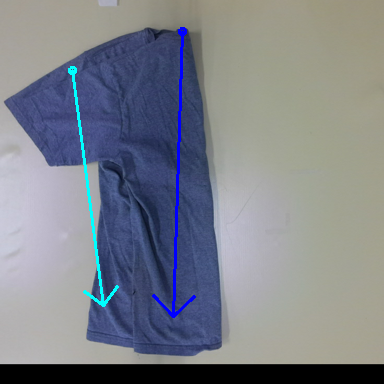}
        \caption{\scriptsize \texttt{Fold the tshirt in half, with the top side overlapping the bottom.}}
    \end{subfigure}
    \begin{subfigure}[t]{0.24\linewidth}
        \centering
        \includegraphics[width=\linewidth]{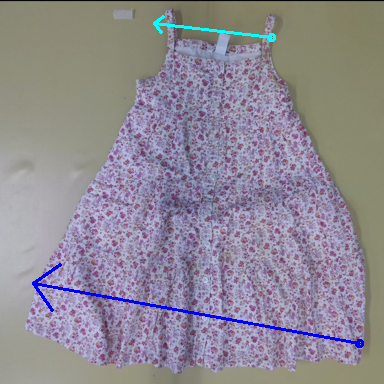}
        \caption{\scriptsize \texttt{Fold the waistband of the dress in half, from left to right.}}
    \end{subfigure}
    \begin{subfigure}[t]{0.24\linewidth}
        \centering
        \includegraphics[width=\linewidth]{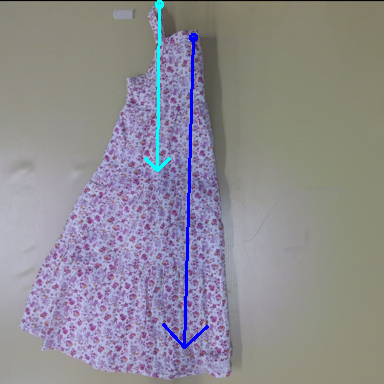}
        \caption{\scriptsize \texttt{Bring the top side of the skirt towards the bottom side and fold them in half.}}
    \end{subfigure}
    \caption{\textbf{Additional qualitative examples on real data: }We present the results obtained with the different configurations of the real dataset and trying a diverse set of prompts taken from the templates of our dataset.}
    \label{fig:qualitative_real_extra}
\end{figure}

We present some qualitative evaluations on the unimanual folding dataset in \cref{fig:qualitative_unimanual}. We also include some bimanual action predictions on the images obtained from our dataset in \cref{fig:qualitative_sim_extra}, on Softgym in \cref{fig:qualitative_softgym}, and on real clothes on \cref{fig:qualitative_real_extra}. 

\clearpage

\subsection{End-to-end unimanual folding rollout}

\begin{figure}[ht]
    \centering
    \begin{subfigure}{.20\linewidth}
        \centering
        \includegraphics[width=\linewidth]{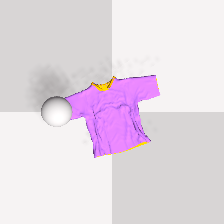}
    \end{subfigure}
    \begin{subfigure}{.20\linewidth}
        \centering  
        \includegraphics[width=\linewidth]{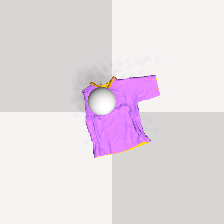}
    \end{subfigure}
    \begin{subfigure}{.20\linewidth}
        \centering  
        \includegraphics[width=\linewidth]{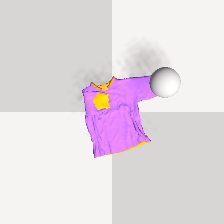}
    \end{subfigure}
    \begin{subfigure}{.20\linewidth}
        \centering  
        \includegraphics[width=\linewidth]{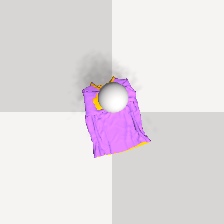}
    \end{subfigure}
    \begin{subfigure}{.20\linewidth}
        \centering  
        \includegraphics[width=\linewidth]{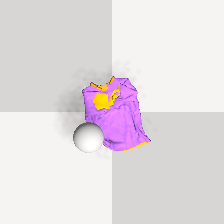}
    \end{subfigure}
    \begin{subfigure}{.20\linewidth}
        \centering  
        \includegraphics[width=\linewidth]{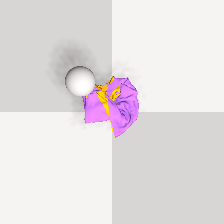}
    \end{subfigure}
    \begin{subfigure}{.20\linewidth}
        \centering  
        \includegraphics[width=\linewidth]{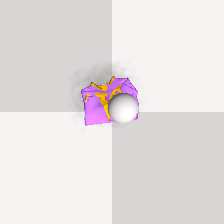}
    \end{subfigure}
    \begin{subfigure}{.20\linewidth}
        \centering  
        \includegraphics[width=\linewidth]{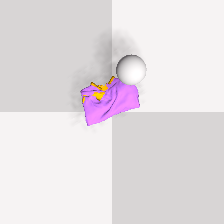}
    \end{subfigure}
    \caption{\textbf{End-to-end unimanual folding: }Example of a whole folding rollout using the BiFold with context trained with 1000 demonstrations on the unimanual folding dataset \cite{language_deformable_manipulation}.}
    \label{fig:unimanual_rollout}
\end{figure}

Similarly to the end-to-end bimanual cloth folding frames included in \cref{fig:qualitative_bimanual_rollout}, we include an example of a full rollout for the unimanual model in \cref{fig:unimanual_rollout}.

\subsection{Out of distribution prompts}

\begin{figure*}[ht]
    \centering
    \captionsetup[subfigure]{labelformat=empty}
    \begin{subfigure}[t]{0.2\linewidth}%
        \centering
        \includegraphics[width=\linewidth]{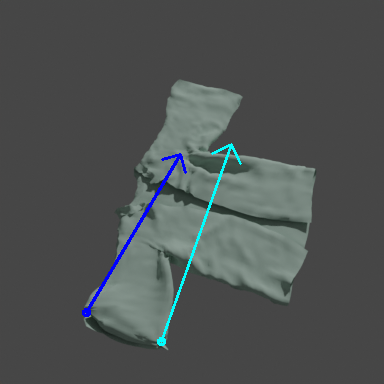}
    \end{subfigure}
    \begin{subfigure}[t]{0.2\linewidth}
        \centering
        \includegraphics[width=\linewidth]{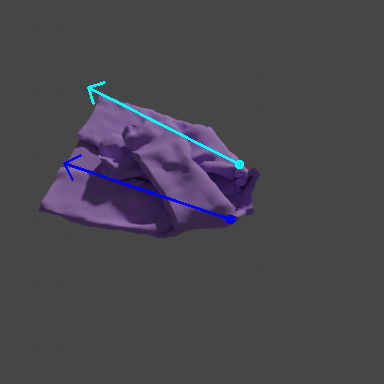}
    \end{subfigure}
    \begin{subfigure}[t]{0.2\linewidth}
        \centering
        \includegraphics[width=\linewidth]{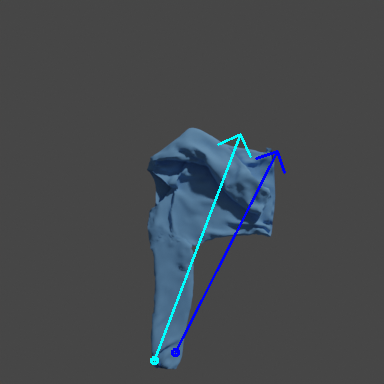}
    \end{subfigure}
    \begin{subfigure}[t]{0.2\linewidth}
        \centering
        \includegraphics[width=\linewidth]{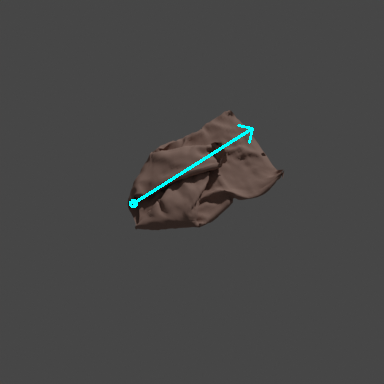}
    \end{subfigure}
    \begin{subfigure}[t]{0.2\linewidth}    
        \centering
        \includegraphics[width=\linewidth]{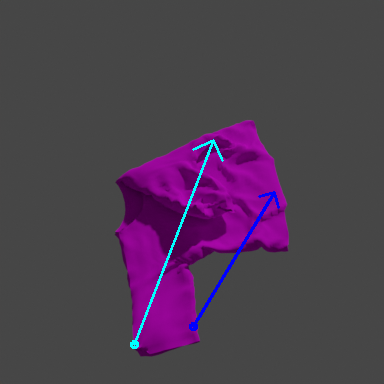}
    \end{subfigure}
    \begin{subfigure}[t]{0.2\linewidth}%
        \centering
        \includegraphics[width=\linewidth]{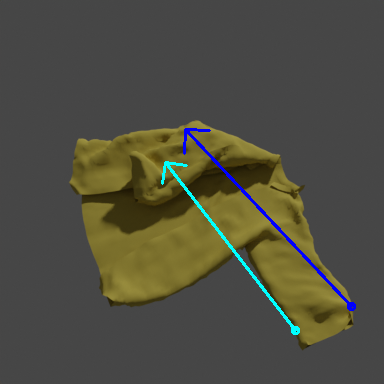}
    \end{subfigure}
    \begin{subfigure}[t]{0.2\linewidth}
        \centering
        \includegraphics[width=\linewidth]{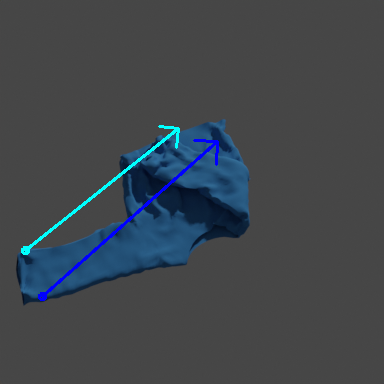}
    \end{subfigure}
    \begin{subfigure}[t]{0.2\linewidth}
        \centering
        \includegraphics[width=\linewidth]{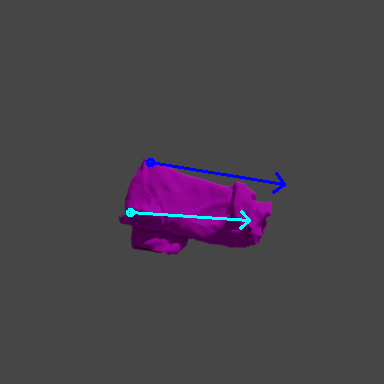}
    \end{subfigure}
    \caption{\texttt{"Fold a T-shirt into a square"}.}
    \label{fig:square_ood}
\end{figure*}

BiFold predicts a single pick-and-place action at a time. Hence, we do not expect it to perform a whole rollout with a single instruction (when performing an end-to-end folding, we provide a sequence of instructions). However, we found it interesting to probe the model with some out-of-distribution instructions. When prompted with \texttt{"Fold a T-shirt into a square"}, the model predicts an action that folds a part of the shirt inwards to make it more similar to a square, as seen in \cref{fig:square_ood}. While the language instruction is new, we can expect this behavior as the dataset contains similar folding steps.

\begin{figure*}[ht]
    \centering
    \captionsetup[subfigure]{labelformat=empty}
    \begin{subfigure}[t]{0.2\linewidth}%
        \centering
        \includegraphics[width=\linewidth]{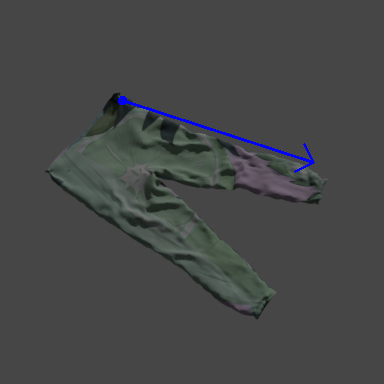}
    \end{subfigure}
    \begin{subfigure}[t]{0.2\linewidth}
        \centering
        \includegraphics[width=\linewidth]{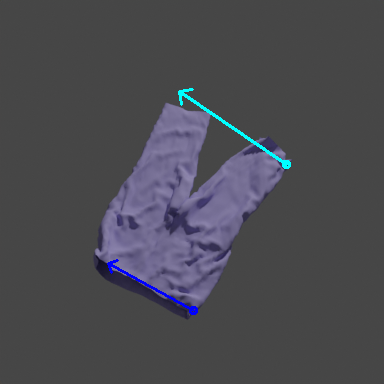}
    \end{subfigure}
    \begin{subfigure}[t]{0.2\linewidth}
        \centering
        \includegraphics[width=\linewidth]{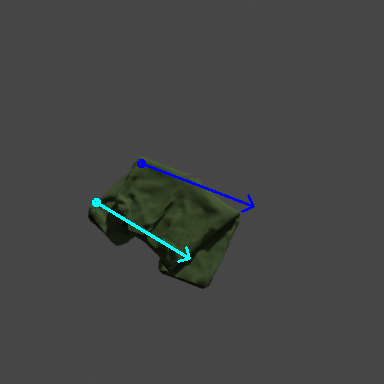}
    \end{subfigure}
    \begin{subfigure}[t]{0.2\linewidth}
        \centering
        \includegraphics[width=\linewidth]{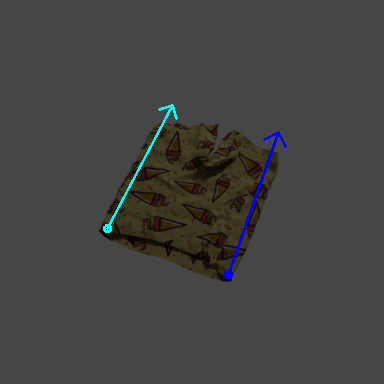}
    \end{subfigure}
    \begin{subfigure}[t]{0.2\linewidth}    
        \centering
        \includegraphics[width=\linewidth]{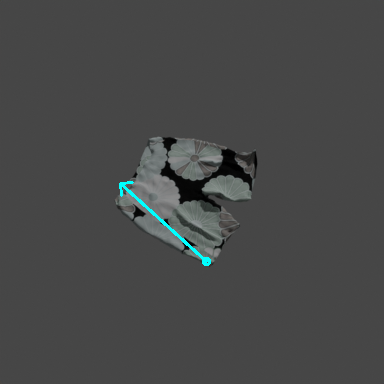}
    \end{subfigure}
    \begin{subfigure}[t]{0.2\linewidth}%
        \centering
        \includegraphics[width=\linewidth]{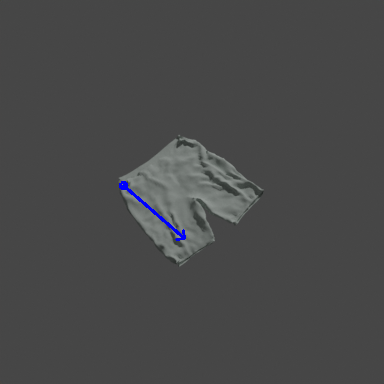}
    \end{subfigure}
    \begin{subfigure}[t]{0.2\linewidth}
        \centering
        \includegraphics[width=\linewidth]{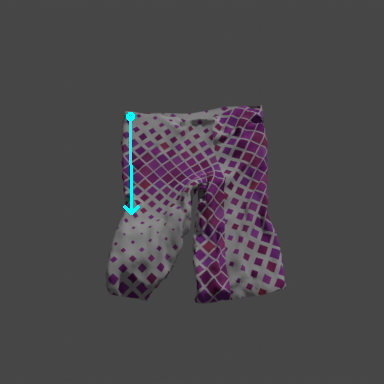}
    \end{subfigure}
    \begin{subfigure}[t]{0.2\linewidth}
        \centering
        \includegraphics[width=\linewidth]{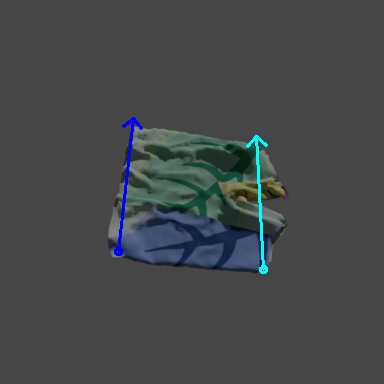}
    \end{subfigure}
    \caption{\texttt{"Fold the trousers in L shape"}.}
    \label{fig:lshape_ood}
\end{figure*}

When using the instruction \texttt{"Fold the trousers in L shape"} we obtain outputs like the ones in \cref{fig:lshape_ood}. In the first row, we see that BiFold ignores the instruction and performs the fold for the current image observation that is statistically more common in the dataset. In the second row, we can see that it tries different actions with a single hand that can. Notably, there were many more actions using only one hand for this instruction than on average, but the model does not replicate a way to achieve an L-fold that would grasp the end of one of the legs with one or two arms and move it diagonally up and to the other side, which we can expect due to the absence of similar folds in the dataset.

\clearpage

\subsection{Failures}

\begin{figure*}[ht]
    \centering
    \captionsetup[subfigure]{labelformat=empty}
    \begin{subfigure}[t]{0.24\linewidth}    
        \centering
        \includegraphics[width=\linewidth]{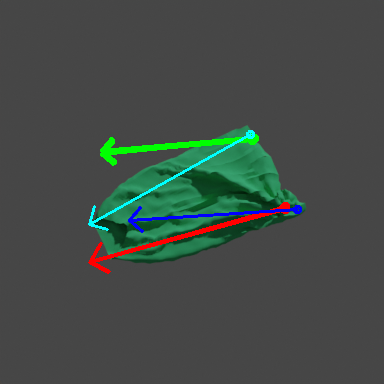}
        \caption{\scriptsize \texttt{Fold the skirt, making sure the bottom side touches the top side.}}
    \end{subfigure}
    \begin{subfigure}[t]{0.24\linewidth}
        \centering
        \includegraphics[width=\linewidth]{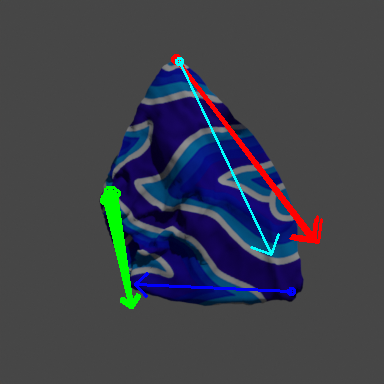}
        \caption{\scriptsize \texttt{Fold the skirt, making a crease from the left to the right.}}
    \end{subfigure}
    \begin{subfigure}[t]{0.24\linewidth}
        \centering
        \includegraphics[width=\linewidth]{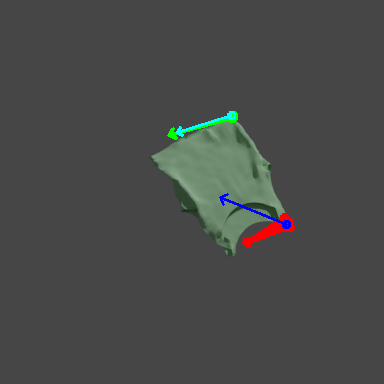}
        \caption{\scriptsize \texttt{Fold the top, right side over left side.}}
    \end{subfigure}
    \begin{subfigure}[t]{0.24\linewidth}
        \centering
        \includegraphics[width=\linewidth]{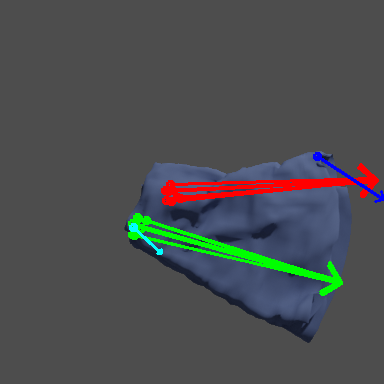}
        \caption{\scriptsize \texttt{Fold the skirt, top side over bottom side.}}
    \end{subfigure}
    \begin{subfigure}[t]{0.24\linewidth}    
        \centering
        \includegraphics[width=\linewidth]{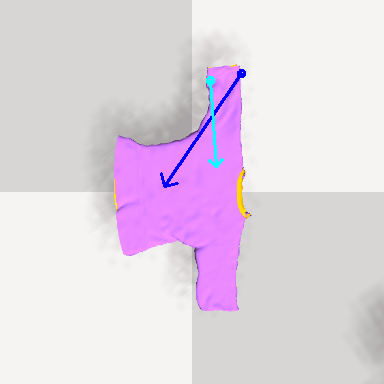}
        \caption{\scriptsize \texttt{Center the right sleeve.}}
    \end{subfigure}
    \begin{subfigure}[t]{0.24\linewidth}
        \centering
        \includegraphics[width=\linewidth]{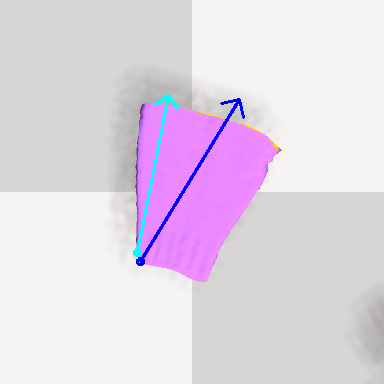}
        \caption{\scriptsize \texttt{Create a fold in the skirt, going from top to bottom.}}
    \end{subfigure}
    \begin{subfigure}[t]{0.24\linewidth}
        \centering
        \includegraphics[width=\linewidth]{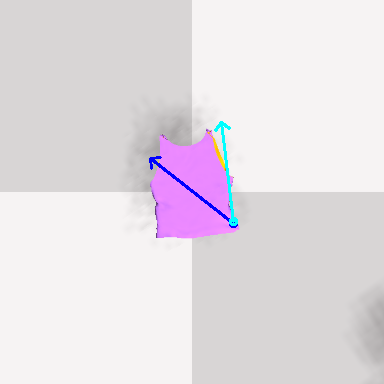}
        \caption{\scriptsize \texttt{Fold the top neatly, from the bottom side to the top side.}}
    \end{subfigure}
    \begin{subfigure}[t]{0.24\linewidth}
        \centering
        \includegraphics[width=\linewidth]{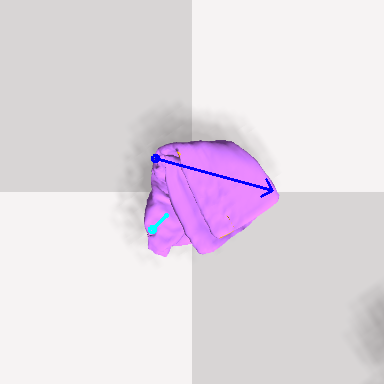}
        \caption{\scriptsize \texttt{Crease the trousers down the middle, from top to bottom.}}
    \end{subfigure}
    \begin{subfigure}[t]{0.24\linewidth}    
        \centering
        \includegraphics[width=\linewidth]{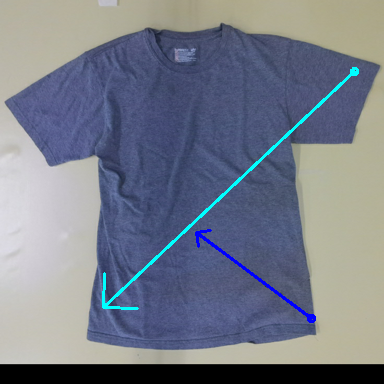}
        \caption{\scriptsize \texttt{Fold the tshirt in half, left to right.}}
    \end{subfigure}
    \begin{subfigure}[t]{0.24\linewidth}
        \centering
        \includegraphics[width=\linewidth]{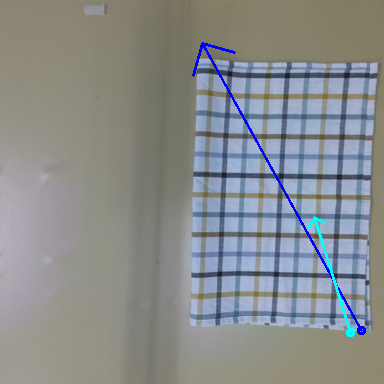}
        \caption{\scriptsize \texttt{Bend the cloth in half, from top to bottom.}}
    \end{subfigure}
    \begin{subfigure}[t]{0.24\linewidth}
        \centering
        \includegraphics[width=\linewidth]{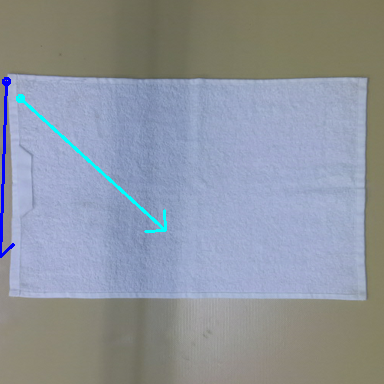}
        \caption{\scriptsize \texttt{Fold the towel neatly, from the left side to the right side.}}
    \end{subfigure}
    \begin{subfigure}[t]{0.24\linewidth}
        \centering
        \includegraphics[width=\linewidth]{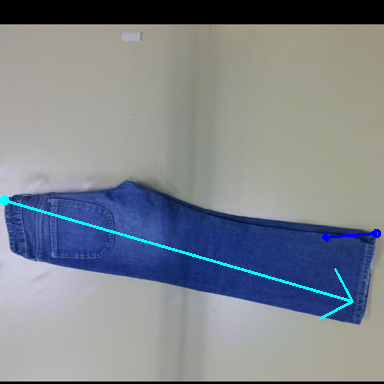}
        \caption{\scriptsize \texttt{Fold the trousers neatly, from the top side to the bottom side.}}
    \end{subfigure}
    \caption{\textbf{Failures:} Collection of the worse predictions obtained for each dataset and cloth category.}
    \label{fig:failures}
\end{figure*}

We present some failures of BiFold with context when predicting bimanual actions in \cref{fig:failures}.

\clearpage

\section{Limitations}

\subsection{Previous approaches}

Most of the previous approaches work with depth maps. While this design decision makes the model naturally invariant to texture and luminance changes, it also discards the color information, which is included in most sensing devices and provides useful cues to understand the cloth state. While the invariance to appearance changes theoretically reduces the sim-to-real gap, real depth sensors are noisy and yield imperfect point clouds. The majority of prior works only train on perfect depth maps and segmentation masks, which makes the reality gap larger in practice. Our solution is based on incorporating color information through a foundational model fine-tuned with LoRA \cite{hu2022lora} to maintain the generalization capabilities gained with large-scale pretraining.

\begin{figure}[ht]
    \centering
    \begin{subfigure}{.45\linewidth}
        \centering
        \includegraphics[trim={25cm 0 25cm 0}, clip, width=\linewidth]{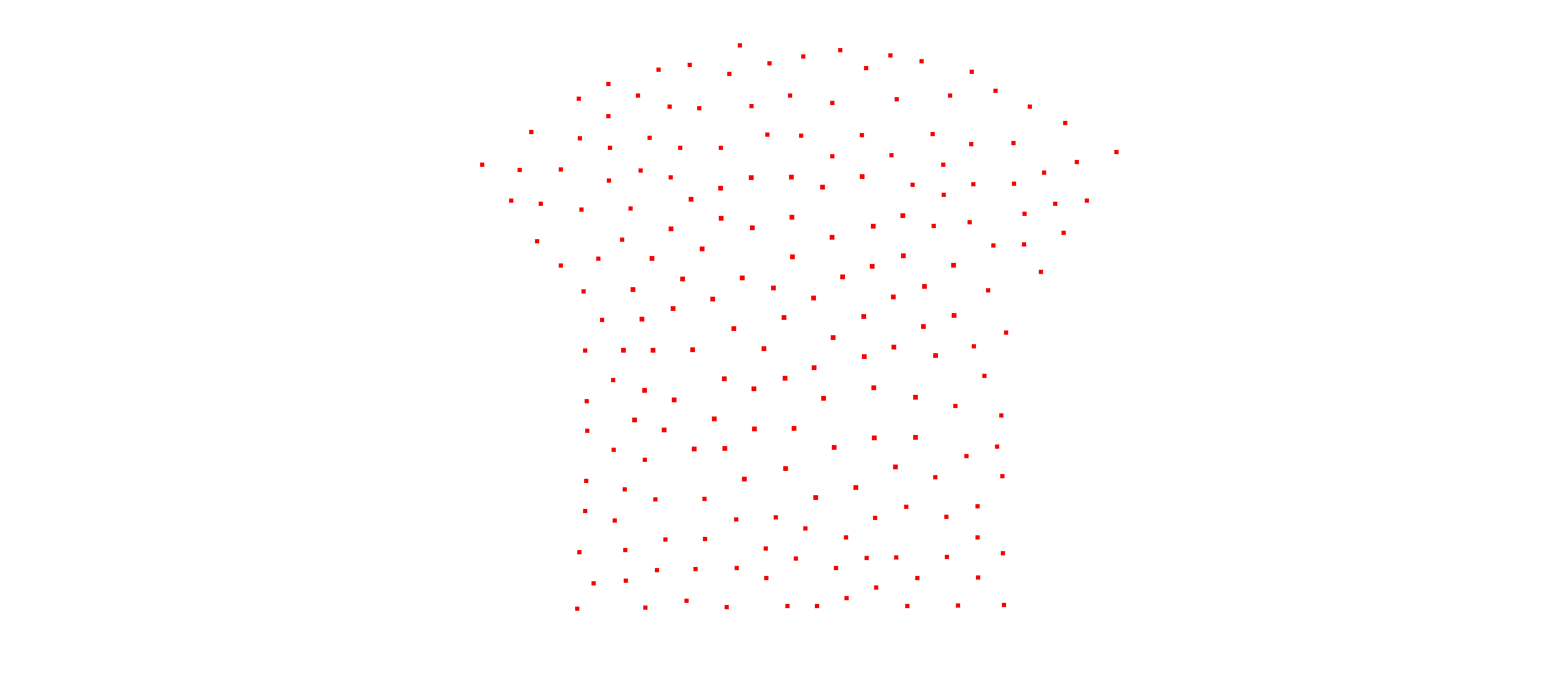}
    \end{subfigure}
    \begin{subfigure}{.45\linewidth}
        \centering
        \includegraphics[trim={25cm 0 25cm 0}, clip, width=\linewidth]{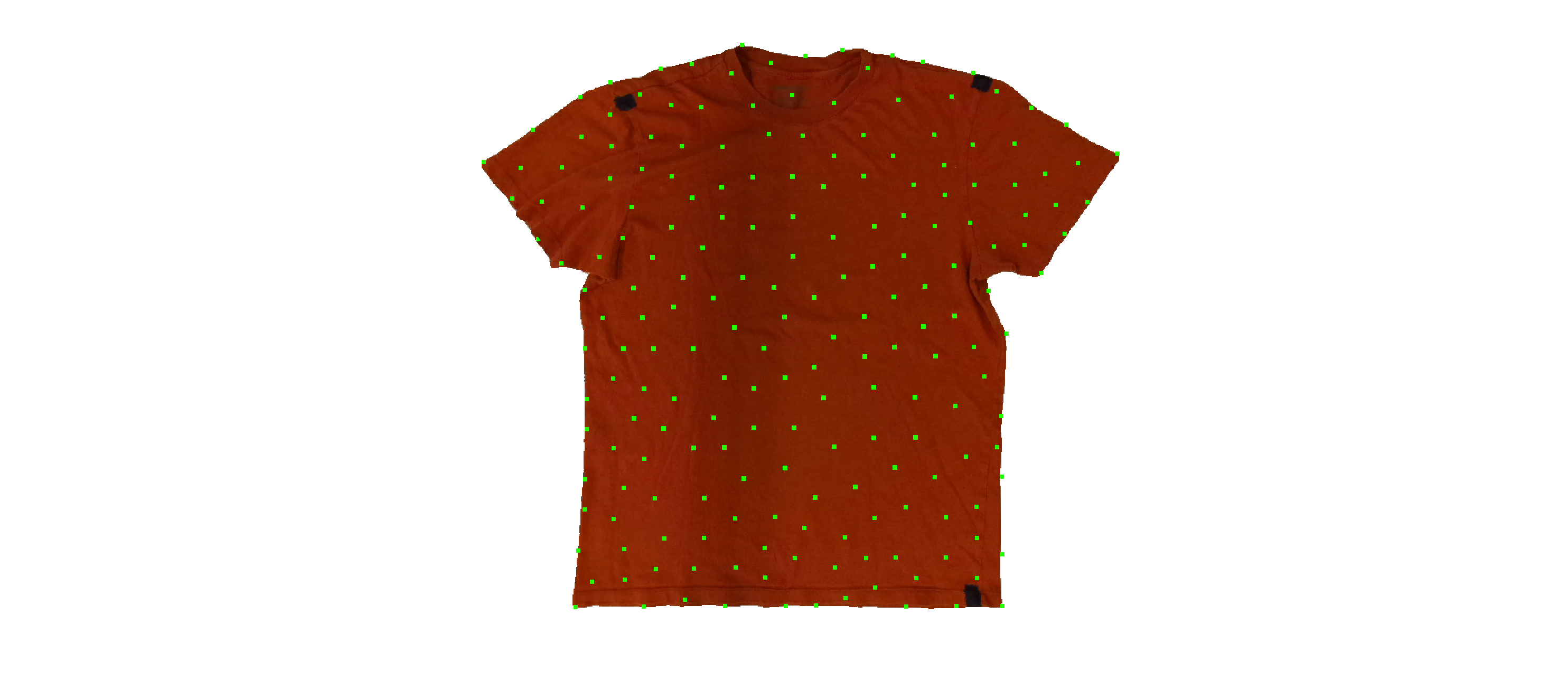}
    \end{subfigure}
    \caption{\textbf{Sub sampled real-world point cloud: }In this figure we show the point cloud with 200 points obtained by following the approach by Deng \etal \cite{language_deformable_manipulation}. On the right, we show the sampled points in green on top of the real-world point cloud. For the sake of visualization, we show the same point cloud in red without a background, where we can see that many details of the input are lost including important cues for manipulation such as the contour and the armpits.}
    \label{fig:subsampled_real_world}
    \end{figure}

While the approach to predict pick positions on segmented point clouds that some prior works adopt has some benefits, it also has many drawbacks that motivated us to operate on the pixel space. The benefit is that, instead of predicting pixel coordinates, these methods predict a probability distribution over the points and selects the best of them for the pick position. Naturally, the point is in the 3D world where the robot operates and does not need to be back-projected, which could ease computing relations between points as Euclidean geometry in 3D is easier than the projective geometry that images have. One drawback of these class of methods is that, in practice, the point cloud needs to be downsampled and some methods can only work with a reduced number of vertices. As an example, Deng \etal \cite{language_deformable_manipulation} only keep 200 points. As shown in \cref{fig:subsampled_real_world}, for real-world point clouds this implies a huge reduction of the input size that hinders learning useful quantities.

In particular, the subsampling fails to cover relevant parts of the manipulated garment such as the edges and corners, where there is less density of points than in the center of the cloth. This limitation is also shared to some extent by other methods relying on processing downsampled point clouds. For example, UniFolding \cite{xue2023unifolding} processes an input point cloud by first downsampling it to 8,000 points, quantizing their coordinates, and finally selecting 4,000 of such quantized points (we get the number of sampled points from the configuration\footnote{\url{https://github.com/xiaoxiaoxh/UniFolding/blob/main/config/virtual_experiment_stage1/train_virtual_base.yaml/\#L8}} and the sampling strategy in the data processing part\footnote{\url{https://github.com/xiaoxiaoxh/UniFolding/blob/fa481bc235b67a9cec2a16811ce4ed1f598c7c13/learning/datasets/vr_dataset.py/\#L234}} of the official implementation). The sampling strategy is to randomly choose from the input set of points without replacement. This strategy may lead to poor coverage of the input point cloud as it is possible that the points of a given zone are not selected. This could be improved by performing FPS instead.

\begin{figure}[ht]
    \centering
    \includegraphics[width=0.5\linewidth]{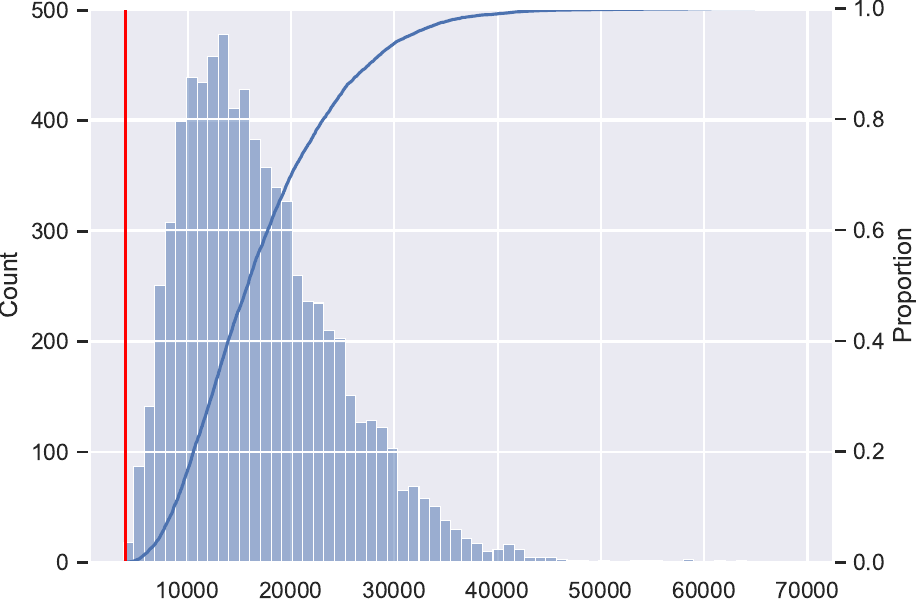}
    \caption{\textbf{Number of pixels in the segmentation mask: }We show a histogram of the number of pixels in the cloth region on the BiFold dataset, with the count indicated in the left axis. We also provide the cumulative distribution of these values, where the cumulative probability is shown in the right axis. Additionally, we add a vertical {\color{red}\textbf{red}} line indicating the number of possible grasp points with the UniFolding approach \cite{xue2023unifolding}.}
    \label{fig:histogram_mask_points}
\end{figure}

Even if using smarter sampling strategies such as FPS, the average number of pixels inside the cloth region, i.e., the effective number of possible pick positions, is on average 17,171. We present the histogram of values in \cref{fig:histogram_mask_points}.

We can see that, while the minimum number of pixels in the cloth region is 3,760, which is slightly under the number of points of UniFolding, the majority of samples have a larger value, being 69,221 points the maximum. If considering the number of total pixels in the image, there are 147,456 possible place points.

\begin{figure}[ht]
    \centering
    \includegraphics[width=0.5\linewidth]{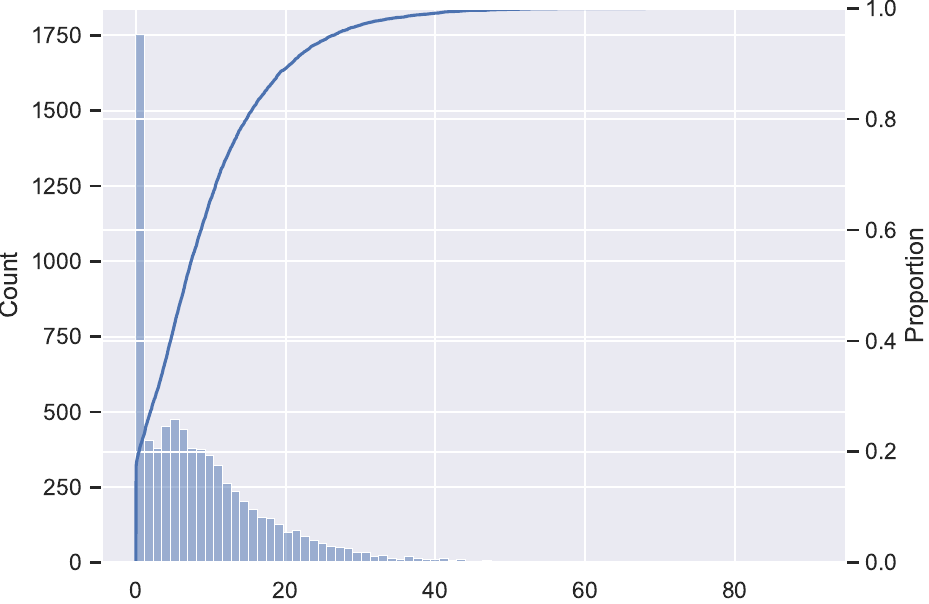}
    \caption{\textbf{Distance in pixels to the segmentation mask}: This plot shows the histogram of distances from place positions to the cloth segmentation mask across the train and test partition of the BiFold dataset, with the count in the left axis. We also provide the cumulative distribution of these values on top using the right axis as scale.}
    \label{fig:histogram_out_of_face}
\end{figure}

\begin{figure}[ht]
    \centering
    \begin{subfigure}{.24\linewidth}
        \includegraphics[width=.95\linewidth]{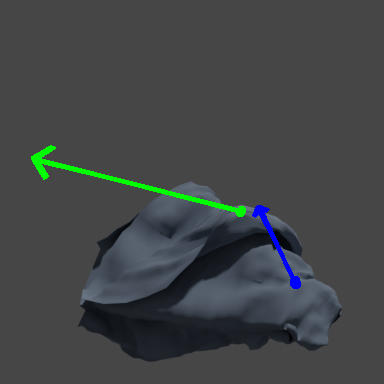}
        \caption{Distance = 105.1 px}
    \end{subfigure}
    \begin{subfigure}{.24\linewidth}
        \includegraphics[width=.95\linewidth]{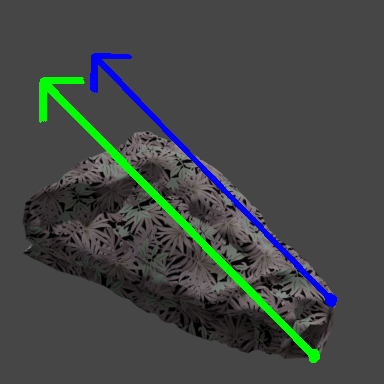}
        \caption{Distance = 86.1 px}
    \end{subfigure}
    \begin{subfigure}{.24\linewidth}
        \includegraphics[width=.95\linewidth]{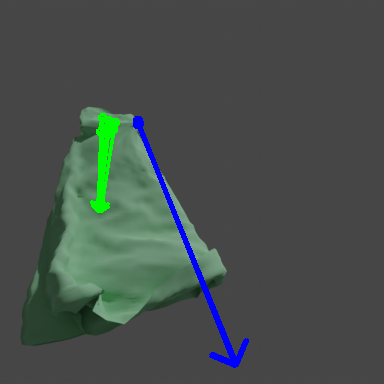}
        \caption{Distance = 78.2 px}
    \end{subfigure}
    \begin{subfigure}{.24\linewidth}
        \includegraphics[width=.95\linewidth]{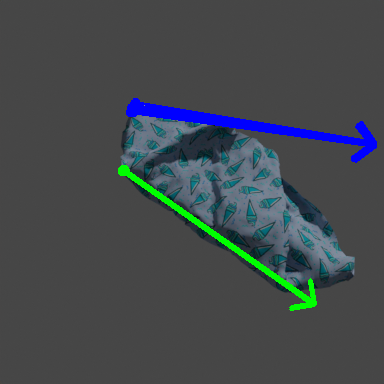}
        \caption{Distance = 78.2 px}
    \end{subfigure}
    \caption{\textbf{Extreme examples of place position out of segmentation mask: }Pick and place positions for {\color{green}\textbf{right}} and {\color{blue}\textbf{left}} actions are represented as the origin and endpoints of an arrow. Each action uses eight vertices, which might be distinct and fall into different pixels.}
    \label{fig:extreme_examples}
\end{figure}

Some of the previous approaches predicting both the picking and placing position using a segmented point cloud of the manipulated garment. While all pick positions fall in the cloth region, as they must specify a part of the cloth for the grasp, this is not satisfied for the place positions of the bimanual cloth manipulation dataset. To put numbers to it, we compute the average distance in pixels between the ground truth place positions of a given arm (recall that our dataset provides eight contact points) and the segmentation mask of the cloth. Then, we take the maximum value between the previous distances for the right and left arm, yielding the maximum distance to a mask. Doing so, we obtain the histogram of \cref{fig:histogram_out_of_face}, which shows that roughly 80\% of the samples of the dataset have a place position with a non-zero distance to the segmentation mask. The large number of place positions out of the cloth region shows the importance of using the pixel space and not the cloth point cloud. We include samples with large distances in \cref{fig:extreme_examples}.

\begin{figure}[ht]
    \centering
    \begin{subfigure}{.45\linewidth}
        \centering
        \includegraphics[trim={25cm 0 40cm 0}, clip, width=\linewidth]{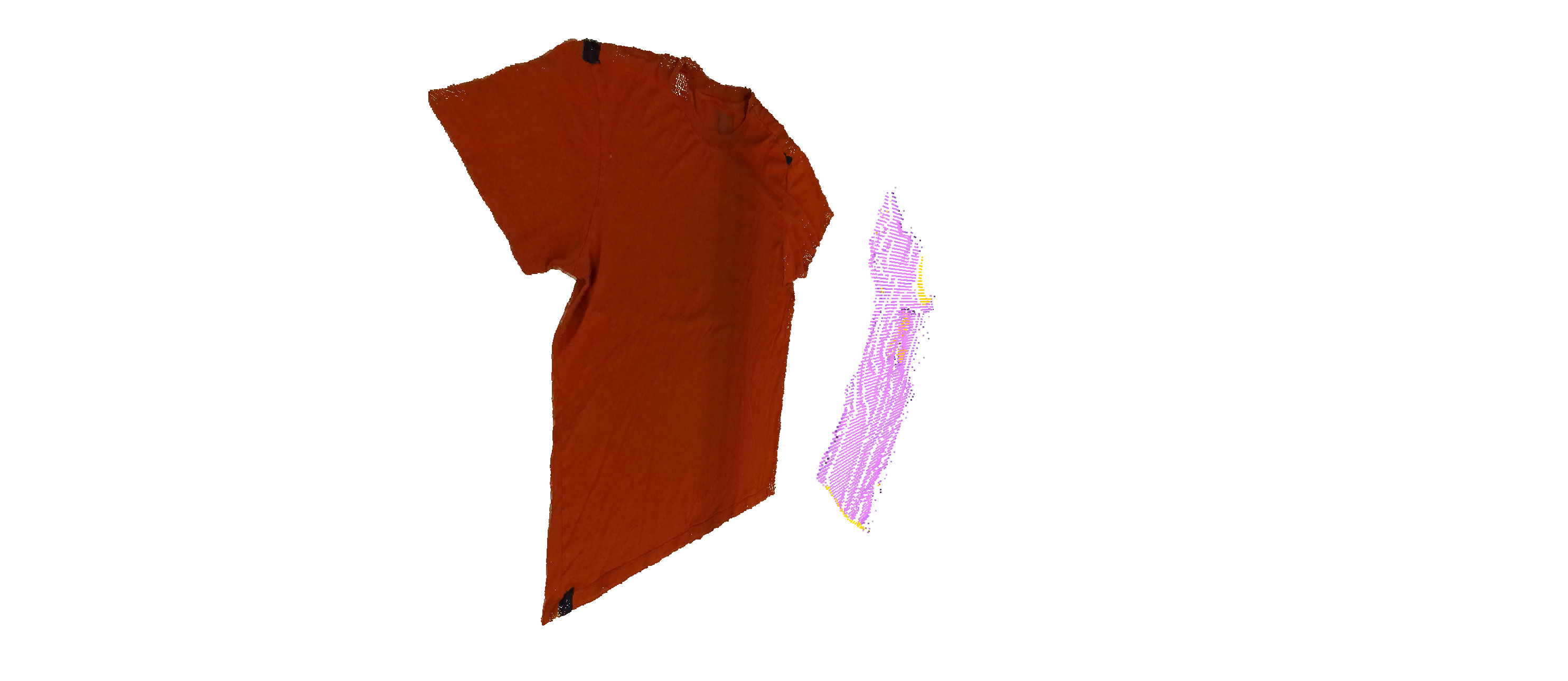}
        \caption{Side view with separated point clouds.}
    \end{subfigure}
    \begin{subfigure}{.45\linewidth}
        \centering
        \includegraphics[trim={25cm 0 30cm 0}, clip, width=\linewidth]{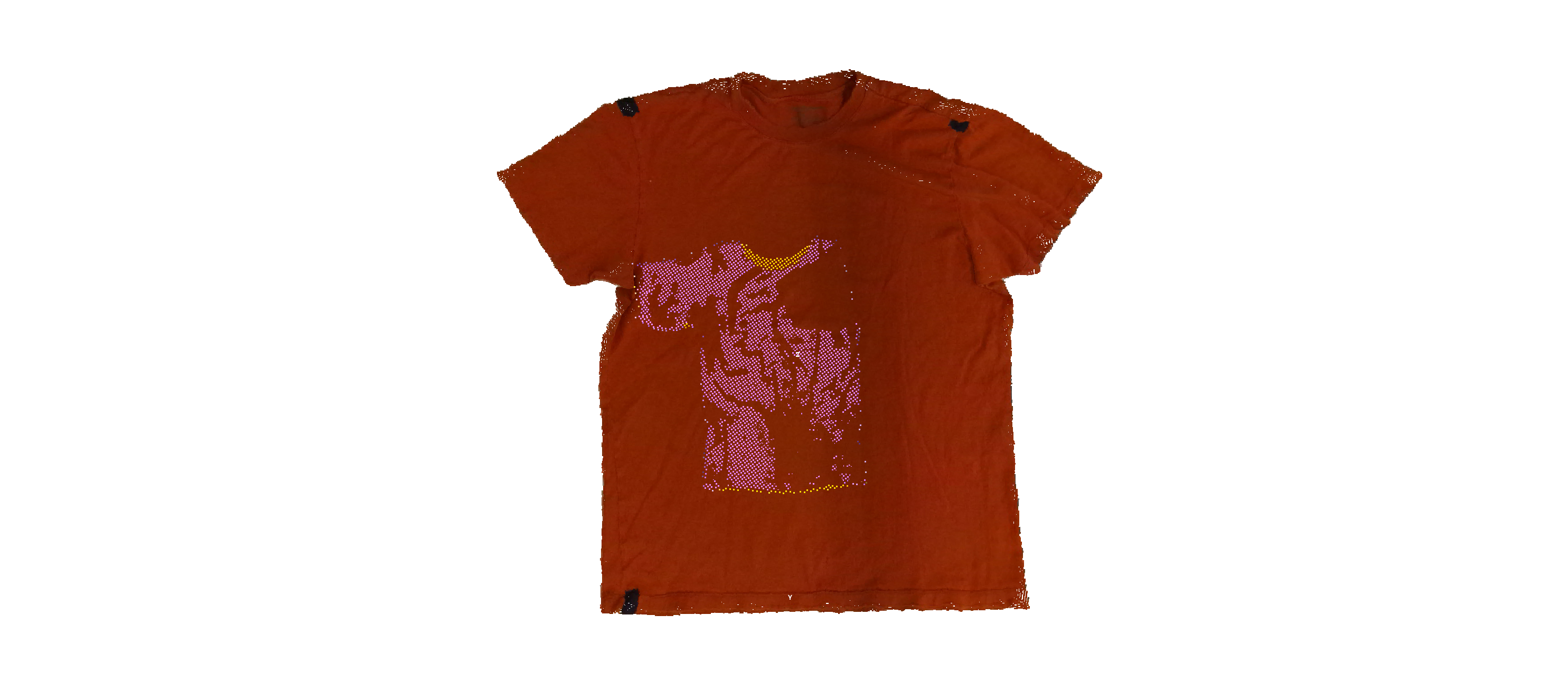}
        \caption{Top view with superimposed point clouds.}
    \end{subfigure}
    \caption{\textbf{Input difference between real garments and those in the unimanual dataset: }We present a comparison between two T-shirts from the real-world dataset (in red) and the unimanual dataset (in pink and yellow). As we can see, the captured real-world point clouds are denser than those in the unimanual dataset ($\sim$307,000 vs.~$\sim$5,000 points). Moreover, we can see a clear difference in scale, with the simulated garment having a rough length from top to bottom of only 30 cm.}
    \label{fig:point_cloud_comparison}
\end{figure}

Another limitation more specific of Deng \etal \cite{language_deformable_manipulation} is that the processing of point clouds uses fixed thresholds. For example, the subsampling is performed after a voxelization step with a fixed grid size of 0.0125 m. Then, the authors create a graph from the resulting point cloud that serves as input to the visual connectivity graph model. To create the graph, the authors find vertex neighbors using a $\ell_2$ ball with a radius of 0.045 m. Overall, this makes the approach highly dependent on the scale, meaning that scaling up the same garment would result in different input graphs. As a reference, we include an illustration with the point clouds of T-shirts from the real-world evaluation and the unimanual dataset in \cref{fig:point_cloud_comparison}.

Finally, another limitation common in the previous methods is that they cannot keep a running memory and take previous observations into account. By doing that, BiFold can resolve ambiguities using past observations (e.g., parts of the cloth that were visible but are not currently visible or symmetries that appear with some cloth states such as a garment folded in a square shape and perceptually similar up-down and left-right directions) and apply refinement actions to correct suboptimal actions.

\subsection{BiFold dataset}

Our dataset does not contain a variety of backgrounds, lighting, and materials. We address the lack of diversity of background by using image masking, and the large set of textures used in our dataset accounts for variability regarding the visual appearance. We obtain semantic locations based on \gls*{nocs} thresholding, which is a reasonable approach but leverages the fact that the garments of the same category have a similar shape. Considering clothes with more variability such as those in the ClothesNet dataset \cite{zhou2023clothesnet} may require other approaches. One possibility is to use a contrastive vision-language model and find the nearest neighbor in the embedding space from a set of category-specific names, \eg, hem, collar, sleeve, cuff. This is similar to approaches that distill language features to 3D, a technique applied to robotics by Shen \etal \cite{shen2023distilled}.

\subsection{BiFold model and evaluation}

BiFold and all other approaches predicting value maps to infer pick and place positions are trained to minimize image-based metrics. As we can observe in the qualitative results that evaluate pixel accuracy in \cref{tab:experiments_bimanual_vr_folding}, the BiFold variant with context can obtain outstanding results. However, we are ultimately interested in improving the success metrics on folding actions instead. One of the problems is that using start and end positions close in the pixel space to the optimal positions may result in radically different configurations after performing the pick-and-place action. This is due to the nearly infinite degrees of freedom of clothes.

When assessing the pick-and-place success, using a dataset of human demonstrations calls into question the evaluation procedure to compare to an oracle. As we have seen, demonstrations are sometimes bad, which can lead to confusing performance reports. While there exist metrics to evaluate simple garment manipulation actions such as folding in half \cite{benchmarking_bimanual}, some of the actions we include in our dataset, \eg, refinement actions, cannot be directly evaluated with such metrics.

\begin{figure}[ht]
    \centering
    \begin{subfigure}[t]{0.24\linewidth}
        \centering
        \includegraphics[width=\linewidth]{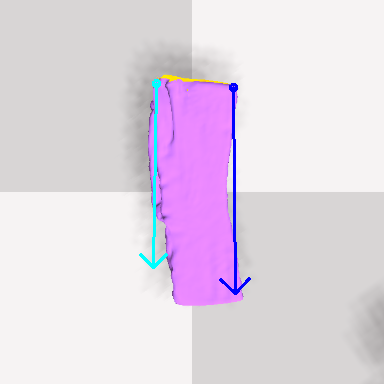}
    \end{subfigure}
    \hfill
    \begin{subfigure}[t]{0.24\linewidth}    
        \centering
        \includegraphics[width=\linewidth]{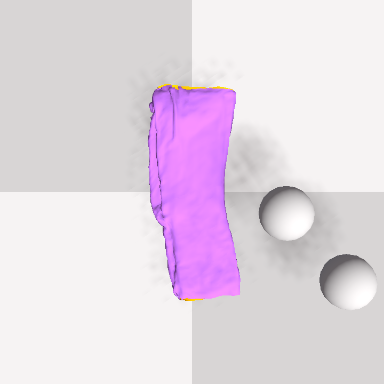}
    \end{subfigure}
    \begin{subfigure}[t]{0.24\linewidth}
        \centering
        \includegraphics[width=\linewidth]{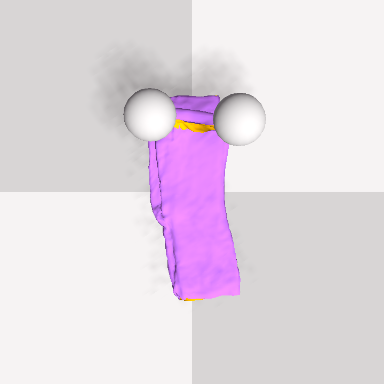}
    \end{subfigure}
    \begin{subfigure}[t]{0.24\linewidth}
        \centering
        \includegraphics[width=\linewidth]{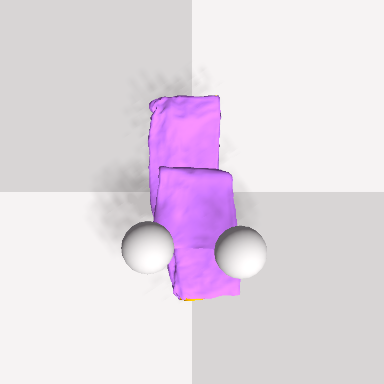}
    \end{subfigure}
    \caption{\textbf{Grasping failure: }Example of an instance in which the root of the failure is on the motion primitive. In this case, the language instruction is \texttt{"Fold the trousers, orientating from the bottom right towards the top left."}. BiFold produces satisfactory pick and place locations (left), but the actions of the primitive (from second image onwards) fail to grasp the lower layer due to hard-coded action to approach the cloth and pick it up.}
    \label{fig:example_bad_grasp}
\end{figure}

Among the baselines considered in this work, Foldsformer \cite{mo2022foldsformer} and Deng \etal \cite{language_deformable_manipulation} use SoftGym as a simulator \cite{lin_2021_softgym}, while CLIPort \cite{shridhar2021cliport} relies on PyBullet \cite{coumans_2021_bullet}. However, according to the benchmark performed by Blanco-Mulero \etal \cite{sim2real_gap}, the cloth physics of these simulators is inaccurate. Among all the simulators tested in the benchmark, MuJoCo \cite{todorov_2012_mujoco} is chosen as the one achieving closer dynamics. Even if the pick and place positions are perfect, the success metric also considers errors from the manipulator. While simulators allow the grasp mesh vertices and hence achieve an ideal grasp, deploying on real robots introduces additional errors due to incorrect grasps or workspace limitations. This can be seen in some examples in simulation, and we include one of them in \cref{fig:example_bad_grasp}.

A natural next step could be to not rely on primitives and directly predict robot actions, which could help improve the grasp and evaluation of real robots. To do that we could use a diffusion policy \cite{chi2023diffusionpolicy} and condition it with the predicted pick and place actions \cite{ma2024hierarchical}.

In this work, we also focus on folding subactions instead of an end-to-end fold. However, performing a fold would amount to either following a predefined list of ordered instructions or using a LLM-based planner to break down the task into steps \cite{huang2023voxposer}.

\end{document}